\pdfoutput=1
\documentclass[10pt,twocolumn,letterpaper]{article}
\newif\ifOnCueArxiv
\OnCueArxivtrue
\PassOptionsToPackage{table}{xcolor}
\ifOnCueArxiv\usepackage[pagenumbers]{cvpr}\else\usepackage[review]{cvpr}\fi
\def\confName{CVPR}
\def\confYear{2027}

\usepackage[T1]{fontenc}
\usepackage{times}
\usepackage{microtype}
\usepackage{array,multirow,colortbl,amsmath,amssymb}
\setlist{nosep,leftmargin=*,label=$\bullet$}
\usepackage{balance}
\usepackage{placeins}
\usepackage{algorithm}
\usepackage{algpseudocode}
\floatname{algorithm}{Algorithm}
\algrenewcommand\algorithmicrequire{\textbf{Input:}}
\algrenewcommand\algorithmicensure{\textbf{Output:}}
\definecolor{cvprblue}{rgb}{0.21,0.49,0.74}
\usepackage[pagebackref,breaklinks,colorlinks,allcolors=cvprblue]{hyperref}
\definecolor{CoreBlue}{HTML}{155797}
\definecolor{IntentBlue}{HTML}{617A8F}
\definecolor{OnCueRuleShade}{gray}{0.96}
\definecolor{OnCueModelShade}{gray}{0.92}
\definecolor{OnCueKeyShade}{HTML}{BDDDF5}
\definecolor{OnCueRankOne}{HTML}{BDDDF5}
\definecolor{OnCueRankTwo}{HTML}{DCECF9}
\definecolor{OnCueRankThree}{HTML}{F0F6FC}
\newcommand{\oncuerank}[2]{\cellcolor{OnCueRank#1}#2}
\newlength{\oncuefracone}
\newlength{\oncuefractwo}
\newlength{\oncuefracthree}
\newcommand{\oncuetablealign}{\settowidth{\oncuefracone}{\normalfont .0}\settowidth{\oncuefractwo}{\normalfont .00}\settowidth{\oncuefracthree}{\normalfont .000}}
\newcommand{\oncuefraction}[2]{\makebox[#1][l]{\if\relax\detokenize{#2}\relax\else .#2\fi}}
\newcommand{\oncuenum}[3]{#2\ifcase#1\relax\or\oncuefraction{\oncuefracone}{#3}\or\oncuefraction{\oncuefractwo}{#3}\or\oncuefraction{\oncuefracthree}{#3}\fi}
\newcommand{\oncuekey}[1]{{\setlength{\fboxsep}{1.2pt}\hspace{-\fboxsep}\colorbox{OnCueKeyShade}{\strut#1}\hspace{-\fboxsep}}}
\newcommand{\oncuechange}[1]{\nobreak\hspace{1.5pt}\makebox[27pt][r]{{\normalfont\fontsize{6.7}{7.5}\selectfont\color{black!50}(#1)}}}

\newcommand{\oncuevalueheading}[1]{#1\kern28.5pt}
\newcommand{\oncuenochange}{\kern28.5pt}
\newcommand{\oncuetail}[1]{\mbox{#1}}
\newcommand{\oncuestd}[1]{\nobreak\hspace{1pt}{\normalfont\fontsize{6.6}{7.3}\selectfont\color{black!65}\ensuremath{\pm}}\hspace{1pt}\makebox[10pt][r]{{\normalfont\fontsize{6.6}{7.3}\selectfont\color{black!65}#1}}}
\newcommand{\oncuenostd}{\hphantom{\oncuestd{0.0}}}
\makeatletter
\newcommand{\oncueci}[1]{\unskip\nobreak\hspace{1.4pt}{\normalfont\fontsize{6.5}{7.2}\selectfont\color{black!65}\oncue@ci#1\@nil}}
\def\oncue@ci#1,#2\@nil{\makebox[20pt][r]{[#1,}\makebox[20pt][r]{#2]}}
\makeatother
\makeatletter
\providecommand{\theHALG@line}{}
\renewcommand{\theHALG@line}{\thealgorithm.\arabic{ALG@line}}
\makeatother

\newcommand{\oncuetitlerest}{Generating Safety-Critical Driving Scenarios\\under Occlusion with Temporal Control}
\ifOnCueArxiv
\title{\raisebox{-0.18\height}{\includegraphics[height=19pt]{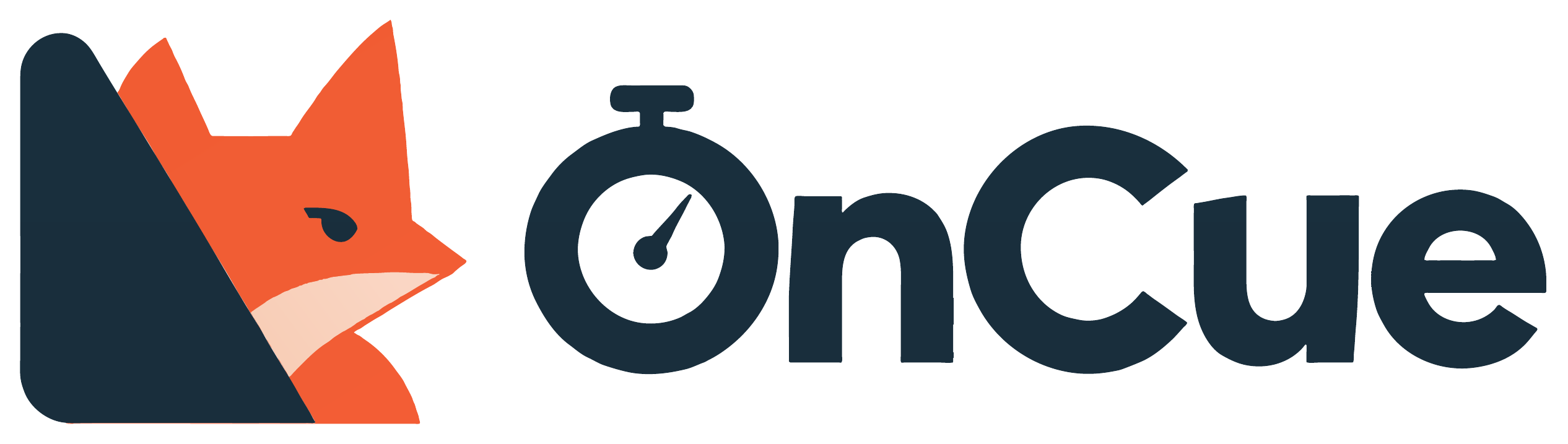}}: \oncuetitlerest}
\else
\title{OnCue: \oncuetitlerest}
\fi

\ifOnCueArxiv
\author{Kailang Ma$^{1}$\quad Li Song$^{2}$\quad Weining Ren$^{3}$\quad Zirui Li$^{4}$\quad
Xianjie Dai$^{1}$\\
Jingguang Li$^{1}$\quad Zuyi Guo$^{1}$\quad Benwang Chen$^{1}$\quad Heye Huang$^{1,*}$\\[4pt]
$^{1}$Korea Advanced Institute of Science \& Technology\quad $^{2}$Zhijia Technology\\
$^{3}$The Chinese University of Hong Kong\quad $^{4}$Nanyang Technological University\\[2pt]
{\small $^{*}$Corresponding author: \texttt{heye.huang@kaist.ac.kr}}}
\else
\author{Anonymous submission}
\fi
\date{}
\begin{document}
\maketitle
\begin{abstract}
Occluded hazards challenge autonomous driving because visibility timing determines available avoidance actions. Existing scenario generators control hazard motion without jointly specifying exposure and reference-contact times. We propose OnCue, a framework that generates safety-critical encounters in real driving logs with these temporal conditions. Scene geometry provides hidden starts and timed waypoints for pedestrian, cyclist, and vehicle trajectory generation. An offline response oracle labels encounters as Reactive (avoidable after exposure) or Anticipatory (avoidable only with earlier visibility) and stores safe maneuvers as training demonstrations. On the Waymo Open Motion Dataset, 90.1--96.6\% of valid trajectories meet both requested times within $\pm 0.2$~s. OnCue matches the requested reaction window between them in 92.8\% of shared vehicle requests versus at most 42.1\% for adapted adversarial generators. Across matched cyclist encounters, longer windows lower PDM-Closed collision from 59.1\% to 5.5\%, while the learned planner LatentDriver shows the opposite trend. OnCue demonstrations reduce LatentDriver target collision by 54.8 points on held-out Reactive tests, but reduce overall no-target progress by 12.7~m. These results highlight the value of temporal control for evaluating and improving planners under occlusion.\ifOnCueArxiv{} Code is available at \url{https://github.com/mankasto/oncue}.\fi

\looseness=-1
\par

\end{abstract}
\begin{figure}[!t]
\centering
\includegraphics[width=\columnwidth]{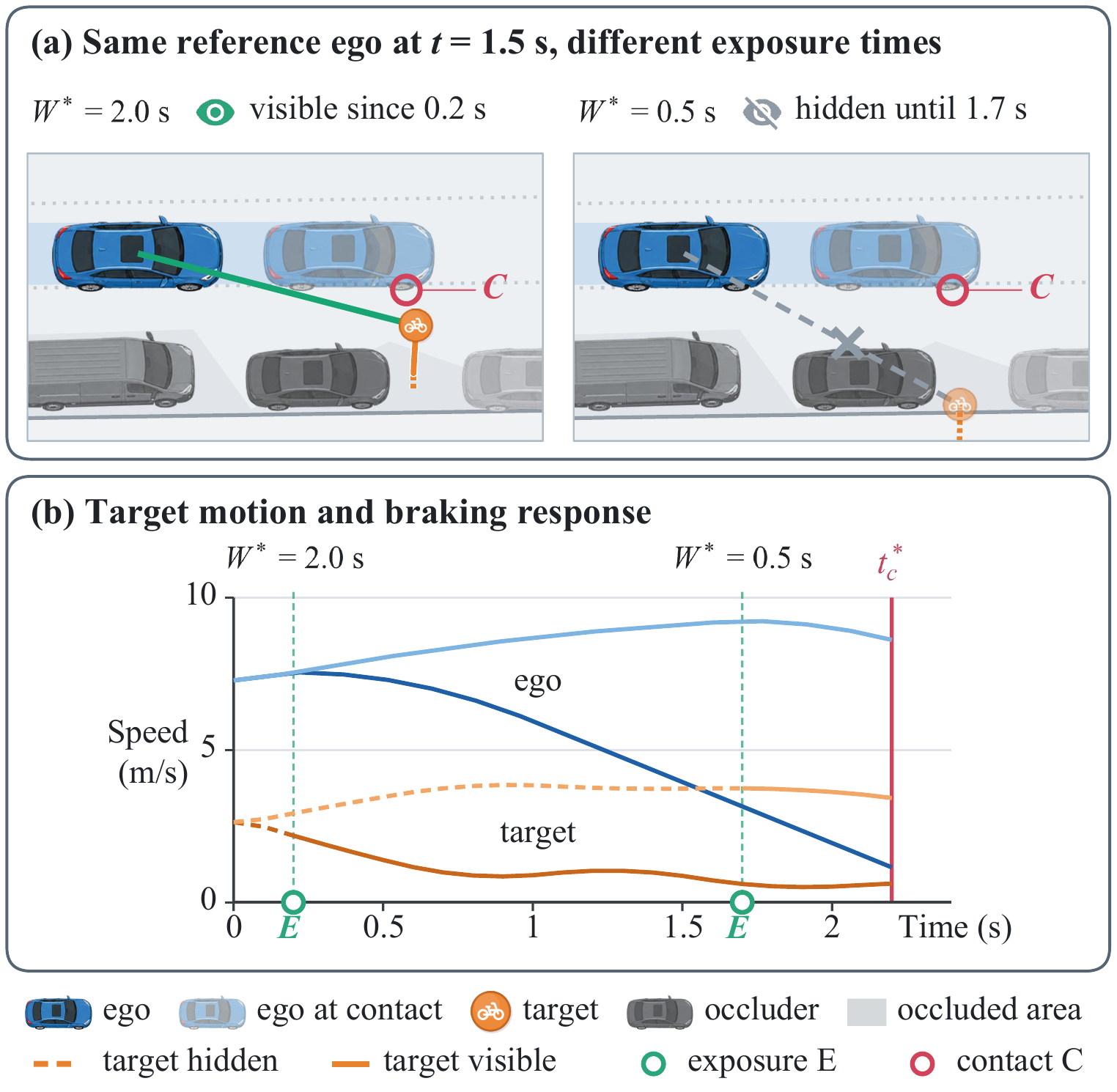}
\caption{\textbf{OnCue controls when a hidden hazard becomes visible.} (a) Two requests with the same reference contact and requested reaction windows $W^* = 2.0$\,s and 0.5\,s. (b) Target and ego speeds under the same braking from exposure $E$. Earlier exposure leaves time to slow before contact $t^*_c$ (red line).}\label{fig1}
\end{figure}

\section{Introduction}

Safe autonomous driving requires timely responses to hazards hidden behind surrounding traffic. A pedestrian emerging from behind a parked vehicle or a cyclist crossing through a queue can leave little time to react. In such encounters, the interval between first visibility and potential conflict determines which avoidance actions remain feasible. These encounters thus test both reaction to visible hazards and anticipation of hidden ones~\cite{fisher2006}. Evaluating planners under occlusion therefore requires controlling when a hidden hazard first becomes visible while keeping the conflict location and time fixed.
\par

Scenario generation makes such rare encounters available for controlled evaluation~\cite{nade,d2rl,safebench}. Driving logs such as the Waymo Open Motion Dataset (WOMD)~\cite{womd} provide realistic traffic context, but replay fixes when road users become visible, so each encounter can only be tested with its recorded timing. Existing generators manipulate target motion to create safety-critical interactions~\cite{cat,seal,sage}, while controllable motion models guide trajectories with rules or costs~\cite{ctg,motiondiffuser}. Occlusion-focused methods further construct hidden encounters and evaluate ego responses~\cite{dykhne2024,hcmts,kahn2022,urm}. However, these approaches mainly control how a hazard moves and generally lack joint control over when it becomes visible and when it reaches the ego. This makes it difficult to test the same conflict under earlier or later exposure. Moreover, collision outcomes alone cannot reveal whether safe avoidance remained possible after exposure or required earlier information.
\par

This raises a key question: can we specify both event times while preserving the surrounding traffic? The challenge is that target placement, motion, and visibility are coupled. Changing target speed can shift exposure, while moving the hidden start can alter arrival at the conflict. Matching the reaction window alone is also insufficient, as equal errors in both event times can shift the entire encounter within recorded traffic. Temporal control therefore requires jointly constraining exposure and reference-contact times.
\par

To address this challenge, we propose OnCue, a framework for generating safety-critical driving scenarios with joint temporal control. The key idea is to make exposure and reference contact explicit conditions of target motion. A fixed reference ego defines the conflict independently of the tested planner. The target's first confirmed visibility defines its \emph{exposure time}~\cite{rss2017}, while first contact with the reference ego defines its \emph{reference-contact time}~\cite{nhtsa2019paeb}. Their interval specifies the \emph{reaction window} $W$ (Figure~\ref{fig1}). OnCue converts these temporal conditions into geometry-derived hidden starts and timed waypoints to guide learned trajectory generation within recorded traffic. An offline, planner-independent response oracle then searches for safe maneuvers, providing response labels and executable demonstrations.
\par

As a general framework, OnCue turns hazard timing from a fixed property of replayed logs into a controllable test variable. By varying exposure while fixing the reference contact, it reveals how planners respond to different reaction windows. Its response oracle distinguishes encounters with available post-exposure maneuvers (\emph{Reactive}) from those where safe actions are found only under earlier visibility (\emph{Anticipatory}), within the evaluated action library. Moreover, the resulting safe maneuvers serve as training demonstrations, connecting failure diagnosis with planner adaptation.
\par

Our main contributions are as follows:
\looseness=-1
\begin{itemize}\setlength{\itemsep}{2pt}\setlength{\parskip}{0pt}\interlinepenalty=10000
\item We propose OnCue, a framework for generating safety-critical driving scenarios under occlusion with temporal control. It enables systematic evaluation of planner responses as hidden hazards become visible earlier or later.
\item We develop geometry-guided conditional trajectory generation, combining constructed hidden starts and timed waypoints with learned motion generation. A planner-independent response oracle labels available avoidance and \oncuetail{stores executable demonstrations.}\looseness=-1
\item We demonstrate how temporally controlled hazards support planner evaluation and adaptation. Closed-loop experiments across target classes and planners reveal how available reaction time affects driving behavior, and training with oracle demonstrations improves target avoidance.
\end{itemize}

\section{Related Work}

\par\noindent\textbf{Safety-critical scenario generation.}
Safety-critical encounters are rare in driving logs, so prior work either synthesizes them for testing~\cite{dingsurvey} or retrieves them at scale~\cite{refav}. Some methods reshape traffic distributions via interventions~\cite{nade,d2rl}, learned statistics~\cite{neuralnde}, encounter generation~\cite{mtg}, or policy-adaptive sampling~\cite{learningcollide}, whereas others search target motion via trajectory and LiDAR perturbations~\cite{advsim}, learned priors~\cite{strive}, or kinematic controls~\cite{king}. Recent methods further select adversarial candidates (CAT~\cite{cat}), learn adversarial policies (GOOSE~\cite{goose}, SEAL~\cite{seal}), guide diffusion toward collisions (SAFE-SIM~\cite{safesim}, DiffScene~\cite{diffscene}), or tune adversariality and realism (SAGE~\cite{sage}). However, these methods target collision or realism and do not model visibility, so they lack control over when a hazard becomes visible.
\looseness=-1
\par

\par\noindent\textbf{Controllable motion generation.}
Controllable motion models let users request behavior through temporal rules~\cite{ctg}, trajectory costs~\cite{motiondiffuser}, or collision specifications~\cite{collide}. Other models extend control to initial placement~\cite{unigen} or add occlusion and signals to scene generation~\cite{scenediffuserpp}, while simulator construction~\cite{chatscene} and log augmentation~\cite{interplan} supply test settings. Yet these requests are rarely tied to when the ego can see the target. OnCue instead derives motion conditions from visibility and contact geometry, linking both requested event times to the generated trajectory.
\looseness=-1
\par

\par\noindent\textbf{Occlusion-aware planning and testing.}
Hidden-agent models~\cite{bivo} and risk evaluation~\cite{kahn2022} characterize unseen hazards. For planning, risk-aware methods use generated encounters~\cite{urm} or reason about unseen obstacles~\cite{foresee,mpcocclusion}, while event-based vision addresses sensing latency~\cite{eventvision}. For testing, prior work generates hidden pedestrian motion~\cite{priisalu2022}, OCC-Gen~\cite{dykhne2024} constructs occlusion-caused collisions, and HC-MTS~\cite{hcmts} searches worst-case hidden trajectories in occluded regions. Neither OCC-Gen nor HC-MTS explicitly controls both exposure and reference-contact times. In contrast, OnCue jointly schedules these events to compare reaction windows within the same traffic situation.
\looseness=-1
\par

\par\noindent\textbf{Planner safety evaluation.}
Standardized protocols~\cite{euroncap2024} and a simulation-based testing framework~\cite{safebench} organize safety evaluation, while driver studies connect hazard anticipation to timely response~\cite{fisher2006}. Yet interpreting a collision requires identifying whether and when avoidance was possible. Ego solution search~\cite{strive,king} and feasibility-aware generation~\cite{frea,safer} check whether avoidance is available, which also depends on when the action starts~\cite{ghodsi2021}. OnCue thus evaluates safe responses relative to exposure, separating collisions avoidable after exposure from those avoidable only earlier, and retains the safe maneuvers as demonstrations.
\looseness=-1
\par

\begin{figure*}[!t]
\centering
\includegraphics[width=\textwidth,height=.76\textheight,keepaspectratio]{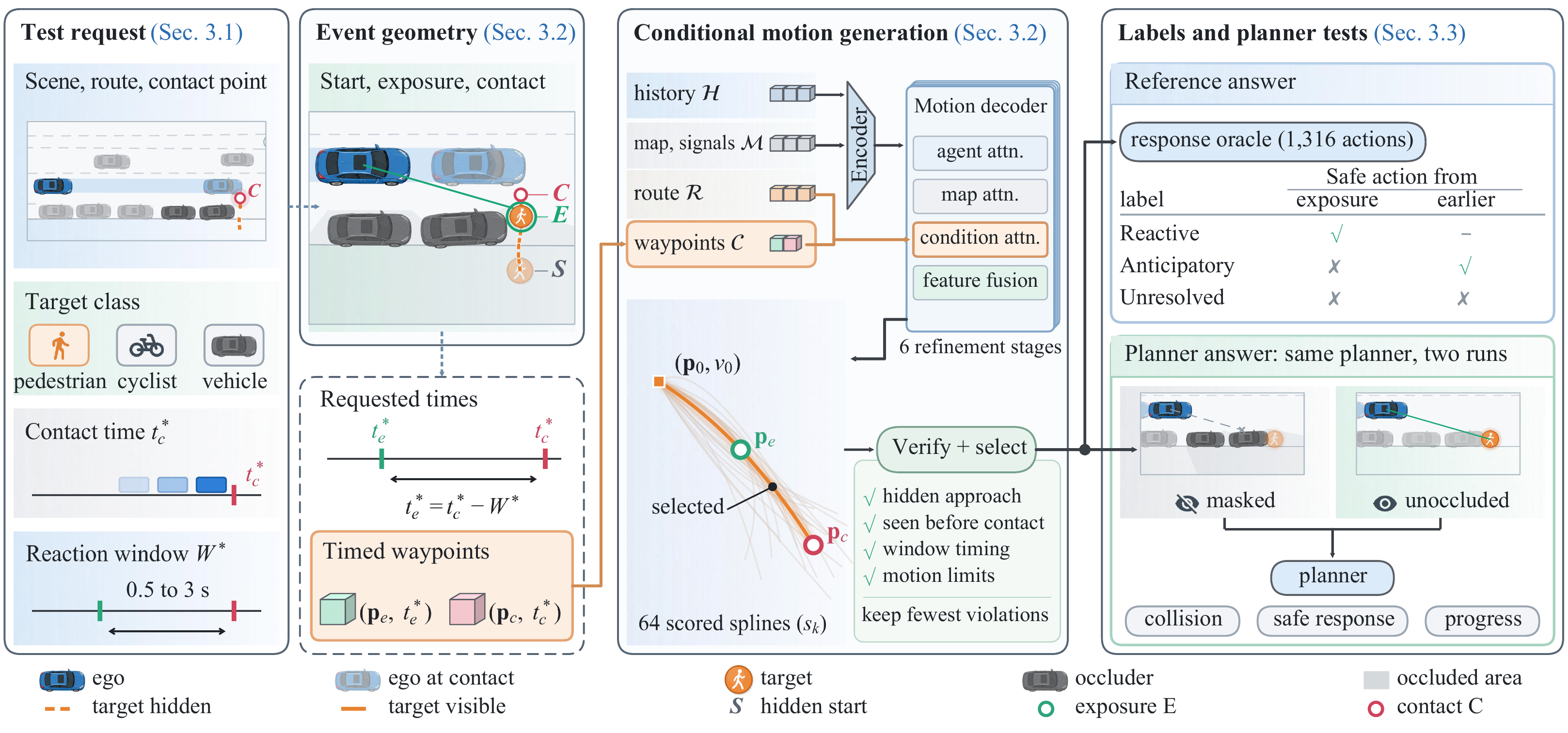}
\caption{\textbf{OnCue framework.} Timed conditions guide generation, followed by response labeling and planner evaluation.}\label{fig2}
\end{figure*}

\section{OnCue: Temporally Controlled Occlusion Hazards}

\subsection{Defining an Occluded Encounter}\label{sec:encounter}

To make hazard timing independent of the tested planner, OnCue measures it against a fixed reference ego, an IDM-driven vehicle that follows the logged route and responds to recorded traffic but not to the inserted target. In contrast, the tested planner drives the evaluated ego in closed-loop evaluation. 
\looseness=-1
\par

With this reference fixed, a request specifies the scene, target class and route, contact point, and requested window $W^*$. Reference arrival at the contact point gives the requested contact time $t_c^*$, and $t_e^*=t_c^*-W^*$. A generated target's realized window $W=t_c-t_e$ runs from its confirmed exposure $t_e$ to its first bounding-box contact $t_c$ with the reference ego.
\par

\par\noindent\textbf{Exposure time.} Following RSS, which defines exposure as first visibility~\cite{rss2017}, OnCue confirms exposure over three consecutive frames and sets $t_e$ to the first. A frame is visible when rays from the reference ego's 2.3 m sensor reach part of the target within an 80 m perception range. Recorded vehicles occlude as road users, and areas off roads, crosswalks, and traffic paths act as static occluders. Construction and evaluation share this rule.
\par

\subsection{Geometry-Guided Trajectory Generation}\label{sec:generation}

\par\noindent\textbf{From event times to motion conditions.}
We express temporal control through two timed waypoints (Figure~\ref{fig2}), where the contact waypoint fixes the conflict within recorded traffic and the exposure waypoint determines when the reference ego first sees the target. Together they form the condition set:
\looseness=-1
\begin{equation}
\mathcal C=\{(\mathbf p_e,t_e^*),(\mathbf p_c,t_c^*)\},\qquad W^*=t_c^*-t_e^*.
\label{eq:event-conditions}
\end{equation}
For event-time errors $e_e=t_e-t_e^*$ and $e_c=t_c-t_c^*$,
\begin{equation}
W-W^*=e_c-e_e.
\label{eq:timing-shift}
\end{equation}
Equal errors preserve $W$ while shifting both events relative to surrounding traffic, so matching the window alone can move the entire encounter to a different point in the recorded traffic sequence. The two timed waypoints therefore constrain both the response interval and the \oncuetail{timing of the encounter.}
\par

To build the two waypoints, we work backward from the requested contact point $\mathbf p_c$ and place $\mathbf p_e$ at the first point along the target route that the reference ego can see at $t_e^*=t_c^*-W^*$ (Section~\ref{sec:encounter}). Figure~\ref{fig14} shows pedestrian, cyclist, and vehicle examples.
\par

A constructed start then fixes the target's state at replay start and its 1\,s history. It is derived from a simple profile that passes both waypoints on time. Moving targets start at training-log speeds and change speed uniformly within class acceleration limits until exposure, then keep the mean speed from $\mathbf p_e$ to $\mathbf p_c$. Standing targets wait hidden and depart at a solved movement start time with a logged acceleration. This profile only sets the start, history, and movement start time, and also serves as the rule-based variant. We keep starts whose approach stays hidden and rank them by agreement with logged motion.
\par

\par\noindent\textbf{Conditional trajectory generation.}
The constructed start and waypoints condition the generator, which fills in the rest of the motion from traffic context, namely target and surrounding histories $\mathcal H$, map and signals $\mathcal M$, and route $\mathcal R$. Starting at the constructed start for moving targets or at movement start time for standing targets, \oncuetail{the generator predicts}
\begin{equation}
\{(\hat\tau_k,s_k)\}_{k=1}^{K}
=f_\theta(\mathcal H,\mathcal M,\mathcal R,\mathcal C).
\label{eq:conditional-completion}
\end{equation}
An MTR-based encoder~\cite{mtr} represents the scene, and route and timed-waypoint tokens enter condition attention at each decoding stage, keeping the requested locations and times available throughout motion refinement. Each of $K=64$ candidates has a score and spline control points, and the spline preserves initial position and velocity so that generated motion connects continuously to the \oncuetail{constructed start}.
\looseness=-1
\par

\par\noindent\textbf{Training objective.}
During training, waypoints are sampled from recorded WOMD futures in the same format as the geometry-derived waypoints used at generation. The six refinement stages use
\begin{equation}
\mathcal L=\frac{1}{D}\sum_{d=1}^{D}
\left(\mathcal L_{\mathrm{reg}}^{(d)}+\mathcal L_{\mathrm{cls}}^{(d)}+\mathcal L_{\mathrm{wp}}^{(d)}\right)
+\mathcal L_{\mathrm{dense}}.
\label{eq:training-objective}
\end{equation}
Here $D=6$. As in MTR, $k^*$ is the mode whose intention point is nearest the ground-truth endpoint, and regression and classification supervise its motion and the scores. The auxiliary $\mathcal{L}_{\mathrm{dense}}$ supervises future positions and velocities of the target and neighboring road users to encode traffic motion, while $\mathcal{L}_{\mathrm{wp}}$ constrains supplied waypoints:
\begin{equation}
\mathcal{L}^{(d)}_{\mathrm{wp}}
= \frac{1}{|\mathcal{J}|} \sum_{j \in \mathcal{J}}
\lambda_j \left\| \hat{\mathbf{p}}^{(d)}_{k^*}(t_j) - \mathbf{p}_j \right\|_1 ,
\label{eq:wp}
\end{equation}
where $\mathcal{J}$ indexes the supplied waypoints $(\mathbf p_j,t_j)$, $\hat{\mathbf p}$ is a predicted position, and $\lambda_j$ weights each position error.
Training samples zero to three waypoints, with $\mathcal{L}^{(d)}_{\mathrm{wp}} = 0$ when $\mathcal{J} = \emptyset$, so one checkpoint serves all classes and ablations. Trajectory supervision learns the motion between waypoints.
\par

At inference, we check candidates in score order within each constructed start. Hard checks require a hidden approach, confirmed exposure before reference contact, satisfied motion constraints, and a reaction-window error within 0.2\,s. Candidates passing these checks are \emph{valid}, while additional background, rule, and motion-quality violations are recorded separately. Search selects the passing candidate with the fewest additional violations. A valid output is \emph{on time} when both events also fall within $\pm0.2$\,s of their requested times. The rule-based variant shares the same construction, checks, \oncuetail{and labels.}
\par

\subsection{Checking Safe Responses}\label{sec:oracle}

For each generated conflict, the offline oracle searches for safe maneuvers after exposure and under earlier visibility. Target and background futures stay fixed, including the generated target motion before exposure, and are available to the oracle. The ego follows the reference until it acts at exposure $t_e$ or at $t_v$, the first confirmed visibility time when the occluders hiding the target are transparent ($t_v=t_e$ if this is not earlier). Both conditions use the same range and visibility rules, and occluders remain collision obstacles. Each search ends when the target first leaves the ego route band after reference contact.
\looseness=-1
\par

The response library combines longitudinal speed changes and lateral avoidance into 1,316 maneuvers. Each maneuver runs through a shared tracker and kinematic bicycle model, and its rollout is checked for vehicle and road limits, target contact, and at-fault background contact. Search first evaluates a 288-action core from both start times and expands to the full library only when no core action is safe after exposure. Labels therefore refer to this library, budget, \oncuetail{and reference protocol.}
\par

Safe actions begun at $t$ among $B$ candidates are
\begin{equation}
\mathcal S_B(t)=\{a\in\mathcal A_B:\operatorname{Safe}(a,t)=1\}.
\label{eq:safe-responses}
\end{equation}
Evaluating safe actions at the two visibility times gives \oncuetail{the response label:}
\begin{equation}
 y=\begin{cases}
 \text{Reactive}, & \mathcal S_B(t_e)\ne\varnothing,\\
 \text{Anticipatory}, & \mathcal S_B(t_e)=\varnothing,\ \mathcal S_B(t_v)\ne\varnothing,\\
 \text{Unresolved}, & \mathcal S_B(t_e)=\mathcal S_B(t_v)=\varnothing.
 \end{cases}
\label{eq:response-label}
\end{equation}
Reactive tests have at least one safe post-exposure maneuver in the evaluated action library, whereas Anticipatory tests have a safe action under earlier visibility but none found after exposure within that library. Tests with insufficient recording or background coverage are labeled Not judged. The least-braking safe rollouts of labeled tests serve as executable demonstrations for planner adaptation (Section~\ref{sec:exp-training}).
\par

\par\noindent\textbf{Planner evaluation.}
We test planners under \emph{masked} observations, which hide occluded road users, and under \emph{unoccluded} observations, which make occluders transparent. Both modes use the same 0.1\,s replay and vehicle executor.
\par

Standing targets start when the evaluated ego meets a position-and-speed trigger in both observation modes, so their start times can differ with ego motion, whereas moving targets keep fixed timetables. Background traffic follows its original timetable, and target-background interactions are rechecked. Additional runs place standing targets on the reference timetable to isolate observation effects. Requested $W^*$ and labels thus describe the reference encounter, while closed-loop outcomes measure planner execution.
\par

\section{Experiments}

\par\noindent\textbf{Setup.}
Driving logs come from the Waymo Open Motion Dataset (WOMD)~\cite{womd}, with simulation in MetaDrive~\cite{metadrive} and generator training on the WOMD training split. Evaluation retains 33,648 of the 40,543 logs in WOMD validation \emph{shards} 12 through 149, excluding \emph{driving episodes} (longer drives split into several logs) shared with development shards 0 through 11. Closed-loop evaluation samples 644 tests across target classes, response labels, and windows, drawing them from 402 logs and 370 driving episodes. Of these tests, 247 have initially standing \oncuetail{targets.}
\par

\par\noindent\textbf{Baselines.}
To test temporal control, we compare CAT~\cite{cat}, KING~\cite{king}, GOOSE~\cite{goose}, SEAL~\cite{seal}, and SAGE~\cite{sage} on vehicle targets, matching their released adversary implementations. Across all classes, preset speed plans, grid-sampled positions or movement start times, and waypoint removal isolate learned motion, start construction, and timed waypoints, respectively (Table~\ref{tab:timing-ablation}). Three IDM variants, which share the reference IDM limits and controller, test planners' use of visibility and reaction time: IDM-base responds to observed traffic, IDM-conservative drives slower, and IDM-occlusion-aware adds virtual hidden hazards. We also test proposal-based PDM-Closed~\cite{pdm} and learned LatentDriver~\cite{latentdriver}.
\par

\par\noindent\textbf{Metrics.}
Over requests with constructed starts, generation success and on-time generation count those yielding valid and on-time outputs, respectively, while on-time share is the on-time fraction of valid outputs. Generator window match and safe-response coverage use all shared requests, where coverage requires an oracle Reactive or Anticipatory label. Masked PDM-Closed runs measure collision and adversarial reward. Following SAGE~\cite{sage}, motion quality uses Wasserstein distances (WD) between generated and logged speed, acceleration, and yaw-rate distributions. Road-edge crossing measures map compliance, and background contact measures target overlap with replayed road users through the test cutoff, \oncuetail{excluding reference-ego contact.}
\par

Closed-loop metrics are ego-target collision, safe-response success, and no-target progress. Progress is ego distance with the target removed, using the 4\,s median for planner comparisons and 8\,s mean for adaptation. Success requires a triggered run without target or at-fault background contact, road violations, or vehicle violations, so stopping can qualify. The run must also continue until the earlier of the test cutoff and 1\,s past the target's arrival at the contact point, and a triggered standing target must add no new background violations. Rates count every attempted test, and untriggered standing targets and \emph{late} masked fixed-timetable runs, where the evaluated ego trails the reference by more than 0.2\,s at exposure, never count as success. Generator rates count missing outputs as failures, and planner comparisons weight strata by eligible-test counts. Window, timetable, and adaptation comparisons use sample rates on their stated subsets.
\par

\subsection{Q1. Can Hazard Timing Be Controlled?}\label{sec:exp-control}

\par
\begin{figure}[!t]
\centering
\includegraphics[width=\columnwidth]{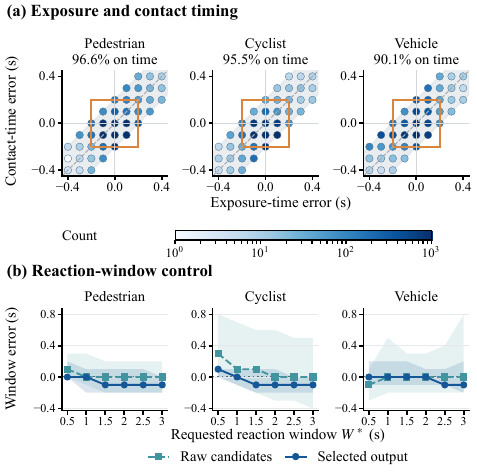}
\caption{\textbf{Temporal control.} (a) Exposure and contact timing errors and percentages for valid outputs. Orange marks both errors within $\pm0.2$\,s. Gray marks the 0.2\,s window tolerance. (b) Median window error with 10th to 90th percentile bands. Raw candidates are evaluated before checks at the first constructed start.}\label{fig3}\label{fig-window-control}
\end{figure}

\begin{figure*}[!t]
\centering
\includegraphics[width=\textwidth]{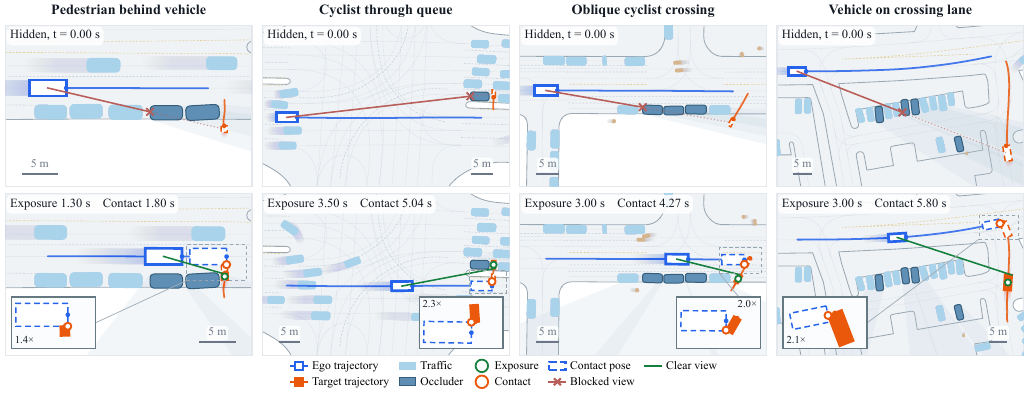}
\caption{\textbf{Generated occluded encounters across pedestrian, cyclist, and vehicle scenarios.} The top row shows hidden approaches, while the bottom row shows first exposure before reference contact. Contact times are interpolated between 10\,Hz frames, and window checks use frame times.}\label{fig4}\label{fig14}
\end{figure*}

\noindent\textbf{Results.} OnCue places both events at the requested times for every target class (Figure~\ref{fig3}), with at least 90.1\% of valid outputs on time and a median absolute window error of 0.1\,s in each \oncuetail{class.}
\par

\par\noindent\textbf{Finding.} For pedestrians and cyclists, timing comes from the event conditions rather than from candidate filtering. On a random tenth of requests, sampled independently of outcomes, the top-scored raw candidate at the first constructed start meets both times in 97.1\% of pedestrian and 89.7\% of cyclist requests before any check. Vehicles are harder in two ways. First, their top-scored raw candidate is on time in only 45.1\% of requests, so their on-time share relies on checks and later starts. Second, only 13.4\% of vehicle requests have a constructed start, which needs a mapped lane crossing the ego route and an occluder hiding the approach, so on-time generation over all vehicle requests is 5.2\%. Even so, OnCue generates 13,572 valid vehicle tests from 33,648 logs, whereas direct replay of logged road users yields \oncuetail{only three qualifying reference-ego conflicts.}
\looseness=-1
\par

\subsection{Q2. How Do Generators Compare?}\label{sec:exp-baselines}

External comparisons test whether existing generators can be adapted to temporal control. None of them accepts requested event times, so each runs its native search against the reference ego from OnCue's constructed start for the same successful vehicle request, with the same one-second constant-speed history (Table~\ref{table2}). Window match thus shows whether this search preserves the requested encounter. Reference replay checks target contact, while masked PDM-Closed runs measure planner \oncuetail{challenge.}
\par

\begin{table*}[!t]
\centering
\caption{\textbf{Generator comparison.} OnCue achieves the highest window match, coverage, and masked PDM-Closed collision on 2,624 shared requests from 446 logs. Requests are selected from valid OnCue outputs. OnCue (rule) keeps its first candidate when none passes the hard checks (900 requests), so both OnCue variants return an output for each request; requests without an output from another generator count as failures. Blue shades mark the top three values across all rows, darkest first, with the best in bold.}\label{table2}\label{table-vehicle-quality}
{\fontsize{7.35}{8.9}\selectfont\setlength{\tabcolsep}{2pt}\renewcommand{\arraystretch}{1.10}
\oncuetablealign
\def\oncuecomparisontablerows{%
\toprule
& \multicolumn{2}{c}{Controlled encounters} & \multicolumn{2}{c}{PDM-Closed} & \multicolumn{5}{c}{Motion quality}\\
\cmidrule(lr){2-3}\cmidrule(lr){4-5}\cmidrule(l){6-10}
Method & $W$ match & Coverage & Collision & Adv. reward & Speed WD & Accel. WD & Yaw-rate WD & Road edge & BG contact\\
& (\%) $\uparrow$ & (\%) $\uparrow$ & (\%) $\uparrow$ & $\uparrow$ & (m/s) $\downarrow$ & (m/s$^2$) $\downarrow$ & (rad/s) $\downarrow$ & (\%) $\downarrow$ & (\%) $\downarrow$\\
\midrule
\multicolumn{10}{l}{\itshape External generators}\\
CAT~\cite{cat} & \oncuenum{1}{26}{8} & \oncuerank{Three}{\oncuenum{1}{57}{5}} & \oncuerank{Three}{\oncuenum{1}{19}{4}} & \oncuerank{Three}{\oncuenum{3}{0}{943}} & \oncuenum{3}{4}{086} & \oncuenum{3}{1}{151} & \oncuenum{3}{0}{339} & \oncuenum{1}{26}{5} & \oncuenum{1}{56}{6}\\
KING~\cite{king} & \oncuerank{Three}{\oncuenum{1}{42}{1}} & \oncuenum{1}{57}{3} & \oncuenum{1}{12}{0} & \oncuenum{3}{0}{707} & \oncuenum{3}{4}{033} & \oncuenum{3}{2}{014} & \oncuenum{3}{0}{352} & \oncuenum{1}{25}{2} & \oncuenum{1}{48}{1}\\
GOOSE~\cite{goose} & \oncuenum{1}{38}{9} & \oncuenum{1}{47}{2} & \oncuenum{1}{9}{2} & \oncuenum{3}{0}{576} & \oncuenum{3}{3}{753} & \oncuerank{Three}{\oncuenum{3}{0}{939}} & \oncuenum{3}{0}{284} & \oncuenum{1}{30}{2} & \oncuenum{1}{49}{0}\\
SEAL~\cite{seal} & \oncuenum{1}{22}{2} & \oncuenum{1}{49}{9} & \oncuenum{1}{12}{5} & \oncuenum{3}{0}{609} & \oncuenum{3}{3}{597} & \oncuerank{One}{\textbf{\oncuenum{3}{0}{382}}} & \oncuerank{One}{\textbf{\oncuenum{3}{0}{269}}} & \oncuenum{1}{42}{7} & \oncuenum{1}{50}{0}\\
SAGE~\cite{sage} ($w=0.5$) & \oncuenum{1}{11}{6} & \oncuenum{1}{22}{9} & \oncuenum{1}{8}{7} & \oncuenum{3}{0}{478} & \oncuenum{3}{2}{979} & \oncuenum{3}{1}{165} & \oncuenum{3}{0}{305} & \oncuerank{Three}{\oncuenum{1}{20}{4}} & \oncuerank{Two}{\oncuenum{1}{43}{6}}\\
SAGE~\cite{sage} ($w=1.0$) & \oncuenum{1}{5}{6} & \oncuenum{1}{13}{8} & \oncuenum{1}{8}{2} & \oncuenum{3}{0}{473} & \oncuerank{One}{\textbf{\oncuenum{3}{1}{549}}} & \oncuenum{3}{2}{084} & \oncuerank{Two}{\oncuenum{3}{0}{279}} & \oncuenum{1}{38}{5} & \oncuerank{One}{\textbf{\oncuenum{1}{38}{7}}}\\
\midrule
\multicolumn{10}{l}{\itshape OnCue variants}\\
OnCue (rule) & \oncuerank{Two}{\oncuenum{1}{81}{7}} & \oncuerank{Two}{\oncuenum{1}{86}{7}} & \oncuerank{Two}{\oncuenum{1}{22}{9}} & \oncuerank{Two}{\oncuenum{3}{1}{076}} & \oncuerank{Two}{\oncuenum{3}{2}{691}} & \oncuerank{Two}{\oncuenum{3}{0}{848}} & \oncuenum{3}{0}{343} & \oncuerank{One}{\textbf{\oncuenum{1}{0}{9}}} & \oncuenum{1}{54}{5}\\
\textbf{OnCue (model)} & \oncuerank{One}{\textbf{\oncuenum{1}{92}{8}}} & \oncuerank{One}{\textbf{\oncuenum{1}{92}{6}}} & \oncuerank{One}{\textbf{\oncuenum{1}{34}{1}}} & \oncuerank{One}{\textbf{\oncuenum{3}{1}{333}}} & \oncuerank{Three}{\oncuenum{3}{2}{935}} & \oncuenum{3}{0}{993} & \oncuerank{Three}{\oncuenum{3}{0}{283}} & \oncuerank{Two}{\oncuenum{1}{3}{9}} & \oncuerank{Three}{\oncuenum{1}{45}{7}}\\
\bottomrule
}
\sbox0{\begin{tabular}{lrrrrrrrrr}\oncuecomparisontablerows\end{tabular}}
\setlength{\tabcolsep}{\dimexpr\tabcolsep+(\textwidth-\wd0)/20\relax}
\begin{tabular}{lrrrrrrrrr}
\oncuecomparisontablerows
\end{tabular}
\par\vspace{2pt}\raggedright\fontsize{7.2}{8.5}\selectfont
WD compares generated motion with logged moving vehicles in the same scenes. BG: background. Adv. reward is against masked PDM-Closed. $W$ match requires reference contact and a realized window within 0.2\,s of $W^*$. Coverage counts Reactive or Anticipatory tests. Task rates include skips, quality uses outputs, and reward uses completed runs. GOOSE uses SEAL's WOMD implementation. SEAL, GOOSE, and KING continue at constant velocity after early termination.\par
}
\end{table*}

\par\noindent\textbf{Results.} OnCue combines accurate timing with the strongest planner challenge on 2,624 shared requests (Table~\ref{table2}). It matches the requested window in 92.8\% of cases, compared with at most 42.1\% for external generators, and crosses road edges far less often than \oncuetail{every external generator.}
\looseness=-1
\par

Even when counting missing outputs as failures over all 6,054 vehicle requests with a constructed start on the same shards, OnCue still has higher window match than every external generator (40.2\% versus at most 27.3\%), coverage similar to KING's, but lower masked PDM-Closed collision than CAT, KING, and \oncuetail{SEAL.}
\par

Background contact, however, remains frequent with replayed traffic, at similar rates for OnCue and SAGE ($w=1.0$) before reference contact (36.9\% and 36.0\%). Replayed background vehicles cannot \oncuetail{yield to the inserted target.} Excluding 124 of 644 closed-loop tests with target-background contact preserves pedestrian and cyclist collision trends (Table~\ref{table3}, Figure~\ref{fig7}) and the adaptation gain in Section~\ref{sec:exp-training}.
\par

\par\noindent\textbf{Finding.} KING contacts the reference ego in 52.9\% of its executed tests before optimization, so collision alone cannot establish temporal control. Most baseline outputs without Reactive or Anticipatory labels lack reference contact or exposure. The comparisons cover vehicles, OnCue-constructed \oncuetail{starts, and PDM-Closed.}
\par

\subsection{Q3. Which Designs Enable Control?}\label{sec:exp-ablation}

On matched requests, ablations replace one component while holding the others fixed. The rule baseline uses preset speed plans, grid starts sample positions or movement start times from logged median speed and acceleration, and waypoint ablations remove the exposure waypoint or both \oncuetail{waypoints.}
\par

\par\noindent\textbf{Results.} Each component improves a different part of control (Table~\ref{tab:timing-ablation}). Constructed starts add 21.2 points of cyclist on-time generation over grid starts, timed waypoints add 30.5 points of pedestrian on-time share over no waypoints, and learned motion adds 13.8 points of vehicle generation success over preset speed plans. Preset speed plans retain higher pedestrian generation success and, in every class, a higher on-time share of valid \oncuetail{outputs.}
\par

Learned motion also lowers pedestrian speed WD from 0.648 to 0.305\,m/s, with paired 95\% intervals below zero for five of the six pedestrian and cyclist WD differences, while background-contact estimates rise by \oncuetail{about one point with intervals spanning zero.}
\par

\begin{table}[!t]
\centering
\caption{\textbf{Components of temporal control.} Starts and waypoints improve timing, and learned motion improves cyclist and vehicle generation. Rates are percentages, with gray point changes from each panel's first row. Panels (b, c) use shards 12--35, unlike Figure~\ref{fig3}.}\label{table9}\label{tab:timing-ablation}
{\fontsize{8.0}{9.5}\selectfont\setlength{\tabcolsep}{2pt}\renewcommand{\arraystretch}{1.10}
\oncuetablealign
\def\oncueablationrows{%
\toprule
Variant & \oncuevalueheading{Pedestrian} & \oncuevalueheading{Cyclist} & \oncuevalueheading{Vehicle}\\
\midrule
\multicolumn{4}{l}{\itshape (a) Learned motion: generation success $\uparrow$}\\
Rule & \textbf{94.2}\oncuenochange & 86.2\oncuenochange & 29.5\oncuenochange\\
OnCue & 90.5\oncuechange{-3.7} & \textbf{93.7}\oncuechange{+7.5} & \textbf{43.3}\oncuechange{+13.8}\\
\midrule
\multicolumn{4}{l}{\itshape (b) Constructed starts: on-time generation $\uparrow$}\\
Grid & 84.4\oncuenochange & 69.1\oncuenochange & 27.7\oncuenochange\\
OnCue & \textbf{87.4}\oncuechange{+3.0} & \textbf{90.3}\oncuechange{+21.2} & \textbf{38.7}\oncuechange{+11.1}\\
\midrule
\multicolumn{4}{l}{\itshape (c) Timed waypoints: on-time share $\uparrow$}\\
None & 66.3\oncuenochange & 68.5\oncuenochange & 69.5\oncuenochange\\
Contact only & 93.5\oncuenochange & 86.3\oncuenochange & 86.9\oncuenochange\\
OnCue & \textbf{96.8}\oncuechange{+30.5} & \textbf{96.0}\oncuechange{+27.5} & \textbf{89.4}\oncuechange{+19.9}\\
\bottomrule}
\sbox0{\begin{tabular}{lrrr}\oncueablationrows\end{tabular}}
\setlength{\tabcolsep}{\dimexpr\tabcolsep+(\columnwidth-\wd0)/8\relax}
\begin{tabular}{lrrr}\oncueablationrows\end{tabular}}
\end{table}

\par

\par\noindent\textbf{Finding.} Without waypoints, pedestrian on-time share falls from 96.8\% to 66.3\%, although starts, search, and checks are unchanged, so construction and selection cannot recover timing alone, consistent with Eq.~\eqref{eq:timing-shift}. Starts make a hidden approach feasible, waypoints place the two events, and learned motion \oncuetail{improves cyclist and vehicle generation success.}
\par

\setcounter{topnumber}{3}

\subsection{Q4. What Do Response Labels Add?}\label{sec:exp-responses}

\begin{figure}[!t]
\centering
\includegraphics[width=\columnwidth]{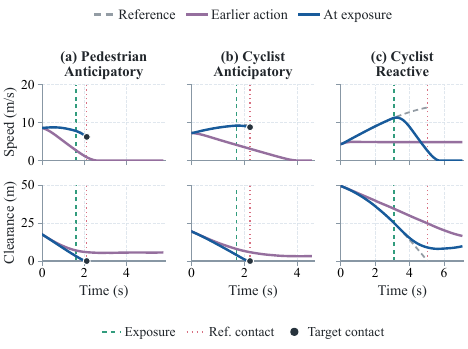}
\caption{\textbf{Oracle action timing.} In (a, b), the same braking action is safe when started at $t_v$ (purple) and collides when started at exposure (blue). Dots mark target contact. In (c), the post-exposure action is safe. Guides mark exposure and reference contact.}\label{fig05}
\end{figure}

With temporal control established, we ask whether planners collide despite safe post-exposure responses, using the oracle (Figure~\ref{fig05}) to analyze masked and unoccluded PDM-Closed runs. For comparison, OCC-Gen~\cite{dykhne2024} inserts a conflicting vehicle and occluder and retains planner collisions under occlusion that disappear with unoccluded observations. We apply this criterion to fixed-timetable OnCue candidates.
\par
\par\noindent\textbf{Results.} Occlusion filtering keeps 56 of 515 candidates and raises the Anticipatory share from 9.9\% to 35.7\%. Yet 35 of the 56 retained failures (62.5\%) still admit a safe post-exposure \oncuetail{action.} When oracle responses are delayed, 412 of the 484 tests initially labeled Reactive (85.1\%) retain a safe action at 0.3\,s and 378 (78.1\%) at 0.5\,s, so most Reactive tests allow delayed avoidance.
\par

\par

\par
Likewise, regenerating tests with a 20\% slower reference ego or a 2.0\,s IDM time headway (desired following time) rarely changes Reactive labels. IDM-conservative and IDM-occlusion-aware still collide least under both reference settings, while about half of pedestrian and cyclist Anticipatory labels become Reactive with the slower reference.
\par

\par\noindent\textbf{Finding.} Many collisions retained by occlusion filtering still admit safe post-exposure maneuvers within the oracle's action library. Response labels distinguish these failures from those where safe actions are found only under earlier visibility. Most Reactive labels survive response delay or a different reference ego, whereas Anticipatory labels depend on reference speed. Oracle labels depend on the action library, known traffic futures, \oncuetail{and geometric visibility.}
\par

\subsection{Q5. How Does Hazard Timing Affect Planners?}\label{sec:exp-planners}

Having identified avoidance opportunities, we vary $W^*\in\{1,2,3\}$\,s across 291 situations, fixing scene, replay start, reference-ego motion, target route, and contact point. Per observation mode, analyzed situations have on-time, labeled outputs and complete runs at every window. With fixed target timetables, shorter windows delay exposure and require faster target motion.
\par

\begin{figure}[!t]
\centering
\includegraphics[width=\columnwidth]{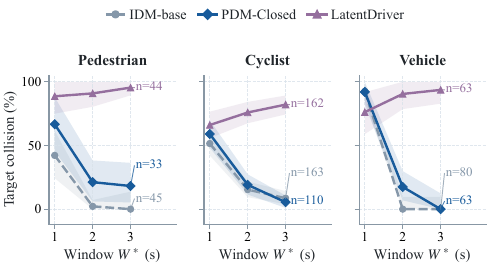}
\caption{\textbf{Response time separates planner behavior.} Target collision under masked observations. Each curve keeps situations complete at all windows, so counts differ by planner. Bands: pointwise 95\% driving-episode bootstrap \oncuetail{intervals.}}\label{fig7}\label{fig11}
\end{figure}

\par\noindent\textbf{Results.} Longer windows reduce PDM-Closed collisions but increase LatentDriver collisions (Figure~\ref{fig7}). PDM-Closed cyclist collision falls from 59.1\% to 5.5\%, with a 52.3-point decrease retained when every window is labeled Reactive. In 29 cyclist situations, LatentDriver avoids collision at $W^*=1$\,s but collides at $W^*=3$\,s.
\par

Passing before the target arrives can explain this short-window avoidance. In these situations, LatentDriver leads the reference ego by a median of about 1\,s at short-window exposure, and its median peak braking at $W^*=3$\,s is zero. By contrast, the 104 cyclist situations that collide at both windows lead by only 0.15\,s, so large leads accompany short-window avoidance.
Turning to observation modes, earlier visibility reduces collision for IDM-base and PDM-Closed (Table~\ref{table3}), and the IDM-base contrast persists on scenes valid for every planner. LatentDriver's collision is unchanged, likely because its released $80\times20$\,m object crop follows visibility filtering and can leave a visible target outside model input.
\par

\begin{table}[!t]
\centering
\caption{\textbf{Anticipatory planner tests.} Earlier visibility reduces collision for IDM-base and PDM-Closed. Rates are weighted by eligible-test counts, and progress is median 4\,s no-target distance. Gray parentheses give unoccluded minus masked collision (points). Blue marks each class's largest reduction and highest masked \oncuetail{success.}}\label{table3}
{\fontsize{7.6}{9.2}\selectfont\setlength{\tabcolsep}{1.5pt}\renewcommand{\arraystretch}{1.08}
\oncuetablealign
\begin{tabular*}{\columnwidth}{@{\extracolsep{\fill}}lrrrr@{}}
\toprule
\multirow{2}{*}{Planner} & \multicolumn{2}{c}{Collision (\%) $\downarrow$} & \multirow{2}{*}{\shortstack{Success $\uparrow$\\(\%)}} & \multirow{2}{*}{\shortstack{Progress $\uparrow$\\(m)}}\\
\cmidrule(lr){2-3}
& Masked & Unoccluded & & \\
\midrule\multicolumn{5}{@{}l}{\itshape Pedestrian: 61 tests, 43 driving episodes}\\
IDM-base & \oncuenum{1}{100}{0} & \oncuekey{\oncuenum{1}{7}{2}}\oncuechange{\oncuenum{1}{-92}{8}} & \oncuenum{1}{0}{0} & \oncuenum{1}{40}{9}\\
IDM-conservative & \oncuenum{1}{11}{6} & \oncuenum{1}{3}{1}\oncuechange{\oncuenum{1}{-8}{4}} & \oncuenum{1}{6}{1} & \oncuenum{1}{23}{9}\\
IDM-occlusion-aware & \oncuenum{1}{23}{7} & \oncuenum{1}{7}{2}\oncuechange{\oncuenum{1}{-16}{5}} & \oncuenum{1}{9}{9} & \oncuenum{1}{17}{1}\\
PDM-Closed & \oncuenum{1}{83}{8} & \oncuenum{1}{53}{5}\oncuechange{\oncuenum{1}{-30}{3}} & \oncuekey{\textbf{\oncuenum{1}{11}{1}}} & \oncuenum{1}{41}{1}\\
LatentDriver & \oncuenum{1}{73}{9} & \oncuenum{1}{73}{9}\oncuechange{\oncuenum{1}{0}{0}} & \oncuenum{1}{7}{4} & \oncuenum{1}{45}{6}\\
\midrule\multicolumn{5}{@{}l}{\itshape Cyclist: 53 tests, 38 driving episodes}\\
IDM-base & \oncuenum{1}{100}{0} & \oncuekey{\oncuenum{1}{26}{8}}\oncuechange{\oncuenum{1}{-73}{2}} & \oncuenum{1}{0}{0} & \oncuenum{1}{42}{8}\\
IDM-conservative & \oncuenum{1}{21}{2} & \oncuenum{1}{14}{4}\oncuechange{\oncuenum{1}{-6}{7}} & \oncuekey{\textbf{\oncuenum{1}{24}{0}}} & \oncuenum{1}{24}{1}\\
IDM-occlusion-aware & \oncuenum{1}{32}{6} & \oncuenum{1}{26}{8}\oncuechange{\oncuenum{1}{-5}{8}} & \oncuenum{1}{10}{3} & \oncuenum{1}{20}{8}\\
PDM-Closed & \oncuenum{1}{87}{9} & \oncuenum{1}{61}{8}\oncuechange{\oncuenum{1}{-26}{1}} & \oncuenum{1}{11}{2} & \oncuenum{1}{42}{6}\\
LatentDriver & \oncuenum{1}{79}{2} & \oncuenum{1}{79}{2}\oncuechange{\oncuenum{1}{0}{0}} & \oncuenum{1}{1}{7} & \oncuenum{1}{45}{6}\\
\midrule\multicolumn{5}{@{}l}{\itshape Vehicle: 31 tests, 29 driving episodes}\\
IDM-base & \oncuenum{1}{100}{0} & \oncuekey{\oncuenum{1}{3}{7}}\oncuechange{\oncuenum{1}{-96}{3}} & \oncuenum{1}{0}{0} & \oncuenum{1}{62}{7}\\
IDM-conservative & \oncuenum{1}{11}{1} & \oncuenum{1}{2}{3}\oncuechange{\oncuenum{1}{-8}{8}} & \oncuenum{1}{19}{9} & \oncuenum{1}{36}{7}\\
IDM-occlusion-aware & \oncuenum{1}{16}{0} & \oncuenum{1}{3}{7}\oncuechange{\oncuenum{1}{-12}{3}} & \oncuekey{\textbf{\oncuenum{1}{35}{7}}} & \oncuenum{1}{49}{3}\\
PDM-Closed & \oncuenum{1}{69}{4} & \oncuenum{1}{44}{7}\oncuechange{\oncuenum{1}{-24}{7}} & \oncuenum{1}{9}{4} & \oncuenum{1}{60}{8}\\
LatentDriver & \oncuenum{1}{90}{3} & \oncuenum{1}{90}{3}\oncuechange{\oncuenum{1}{0}{0}} & \oncuenum{1}{1}{4} & \oncuenum{1}{63}{7}\\
\bottomrule
\end{tabular*}}

\end{table}

Among the IDM variants, IDM-conservative collides less than IDM-occlusion-aware in every class but has lower vehicle success and vehicle progress, so for vehicles, slowing trades progress against hazard \oncuetail{exposure.}
\par

\par

\par\noindent\textbf{Finding.} Window comparisons distinguish avoidance that benefits from extra response time from avoidance consistent with passing first. A single-window collision rate hides this distinction. Changing $W^*$ also changes target motion, so the trends reflect both information timing \oncuetail{and encounter dynamics.}
\looseness=-1
\par

\subsection{Q6. Can the Tests Improve a Planner?}\label{sec:exp-training}

To test whether oracle demonstrations improve planner avoidance, we fine-tune released LatentDriver on equal parts demonstrations and logs from 16 WOMD training shards disjoint from evaluation. The logs-only control matches the learning rate ($2\times10^{-5}$), five epochs, training steps, and seeds 0, 1, and 2.
\looseness=-1
\par

\par\noindent\textbf{Results.} Oracle demonstrations make LatentDriver avoid targets far more often (Table~\ref{tab:planner-training}). On 484 held-out Reactive tests, adding them to logs-only training lowers masked collision by $54.8\pm0.4$ points (driving-episode 95\% interval $[49.0, 60.3]$) and raises safe-response success from $6.5\pm0.4\%$ to $37.8\pm0.2\%$.
\par

The collision reduction is not explained by slower driving alone. The reduction remains $53.8\pm0.8$ points on 332--336 tests per seed where neither run is untriggered or late, and $35.1\pm0.8$ points on 114--116 tests per seed where, in target-free runs, the demonstration-trained model travels at least as far as the logs-only control. Excluding the 124 tests with target-background contact, the reduction is $56.6\pm0.3$ \oncuetail{points.}
\looseness=-1
\par
\begin{table}[!t]
\centering

\caption{\textbf{LatentDriver adaptation.} OnCue demonstrations reduce collision by $54.8\pm0.4$ points on held-out Reactive tests and $35.1\pm0.8$ points on a subset with equal or greater no-target progress. Differences use unrounded values.}\label{tab:planner-training}
{\fontsize{7.5}{9.0}\selectfont\setlength{\tabcolsep}{1pt}\renewcommand{\arraystretch}{1.10}
\oncuetablealign
\newlength{\oncuetrainingcellwidth}
\settowidth{\oncuetrainingcellwidth}{88.8\oncuenostd}
\newlength{\oncuetrainingprogresswidth}
\settowidth{\oncuetrainingprogresswidth}{103.1\oncuenostd}
\newcommand{\oncuetrainingcell}[1]{\makebox[\oncuetrainingcellwidth][r]{#1}}
\newcommand{\oncuetrainingprogress}[1]{\makebox[\oncuetrainingprogresswidth][r]{#1}}
\def\oncuetrainingrows{%
\toprule
& \multicolumn{4}{c}{Rates (\%)} & \shortstack{Progress}\\
\cmidrule(lr){2-5}\cmidrule(l){6-6}
Training & \shortstack{Target\\coll. $\downarrow$} & \shortstack{Safe resp.\\success $\uparrow$} & \shortstack{Untriggered\\or late $\downarrow$} & \shortstack{Other-agent\\contact $\downarrow$} & \shortstack{No-target\\dist. (m) $\uparrow$}\\
\midrule
\multicolumn{6}{@{}l}{\textbf{(a) Full evaluation set}}\\
Pretrained & \oncuenum{1}{73}{3} & \oncuenum{1}{8}{1} & \oncuenum{1}{0}{2} & \oncuenum{1}{17}{5} & \oncuenum{1}{103}{1}\\
\midrule
\hspace{4pt}Logs only & \oncuetrainingcell{\oncuenum{1}{78}{4}\oncuestd{0.2}} & \oncuetrainingcell{\oncuenum{1}{6}{5}\oncuestd{0.4}} & \oncuetrainingcell{\oncuenum{1}{11}{1}\oncuestd{0.3}} & \oncuetrainingcell{\textbf{\oncuenum{1}{0}{4}}\oncuestd{0.1}} & \oncuetrainingprogress{\oncuenum{1}{71}{1}\oncuestd{0.1}}\\
\hspace{4pt}\textbf{Logs + OnCue} & \oncuetrainingcell{\oncuerank{One}{\textbf{\oncuenum{1}{23}{7}}}\oncuestd{0.1}} & \oncuetrainingcell{\oncuerank{One}{\textbf{\oncuenum{1}{37}{8}}}\oncuestd{0.2}} & \oncuetrainingcell{\oncuenum{1}{28}{6}\oncuestd{0.4}} & \oncuetrainingcell{\oncuenum{1}{4}{6}\oncuestd{0.2}} & \oncuetrainingprogress{\oncuenum{1}{58}{4}\oncuestd{0.2}}\\
\midrule
\multicolumn{6}{@{}l}{\textbf{(b) No-target distance: OnCue $\geq$ Logs only}}\\
\hspace{4pt}Logs only & \oncuetrainingcell{\oncuenum{1}{77}{1}\oncuestd{0.5}} & \oncuetrainingcell{\oncuenum{1}{7}{2}\oncuestd{1.3}} & \oncuetrainingcell{\oncuenum{1}{11}{6}\oncuestd{1.4}} & \oncuetrainingcell{\textbf{\oncuenum{1}{0}{0}}\oncuestd{0.0}} & \oncuetrainingprogress{\oncuenum{1}{54}{0}\oncuestd{0.7}}\\
\hspace{4pt}\textbf{Logs + OnCue} & \oncuetrainingcell{\oncuerank{One}{\textbf{\oncuenum{1}{42}{0}}}\oncuestd{1.0}} & \oncuetrainingcell{\oncuerank{One}{\textbf{\oncuenum{1}{37}{4}}}\oncuestd{0.3}} & \oncuetrainingcell{\oncuenum{1}{7}{8}\oncuestd{0.9}} & \oncuetrainingcell{\oncuenum{1}{7}{0}\oncuestd{0.1}} & \oncuetrainingprogress{\oncuenum{1}{59}{9}\oncuestd{0.7}}\\
\bottomrule
}
\sbox0{\begin{tabular}{lccccc}\oncuetrainingrows\end{tabular}}
\setlength{\tabcolsep}{\dimexpr\tabcolsep+(\columnwidth-\wd0)/12\relax}
\begin{tabular}{lccccc}\oncuetrainingrows\end{tabular}
\par\vspace{2pt}\raggedright\fontsize{7.0}{8.2}\selectfont
Mean $\pm$ sample SD over three training seeds. Coll., resp., and dist. denote collision, response, and distance. Contact excludes the target. In (a), rates (\%) use 484 Reactive tests, except contact and mean 8\,s no-target distance over all 644 tests. In (b), all metrics use 114--116 Reactive tests per seed where Logs + OnCue travels at least as far as Logs only without the target.\par
}
\end{table}

\par\noindent\textbf{Finding.} Part of the avoidance gain persists when no-target progress matches or exceeds the logs-only control's. Lower progress, more untriggered or late runs, and increased other-agent contact accompany the gain, though contact remains below the released model's. Removing training windows in which the ego stands still leaves both the avoidance gain and the lower progress in place.
\par

\section{Conclusion}
We presented OnCue, a framework for generating occluded hazards with controlled exposure and reference-contact times in real driving logs.
It combines geometry-guided trajectory generation with an offline response oracle to identify feasible avoidance maneuvers.
Experiments reveal contrasting planner behaviors across reaction windows. Many occlusion-induced failures still admit safe post-exposure maneuvers, suggesting missed responses rather than insufficient visibility alone. Oracle demonstrations improve target avoidance but reduce driving progress and increase other-agent contacts. OnCue remains limited by non-reactive background traffic in log replay, motivating future extensions to interactive traffic environments. \ifOnCueArxiv\else Code, generated tests, and oracle labels will be released upon publication.\fi
\looseness=-1
\par

\label{main-text-end}
\ifOnCueArxiv\section*{Acknowledgements}
This work was supported by the Large-scale Convergence Research Institute Program of the Korea Advanced Institute of Science and Technology (KAIST) (No. N10260180), by the National Research Foundation of Korea (NRF) grant funded by the Korean government (MSIT) under the project ``Development of Risk-Enhanced Continual Learning for Embodied Intelligence in Long-Tail Environments'' (Grant No. RS-2026-25595451), and by the Korea Institute of Science and Technology Information (KISTI) R\&D program through the joint research project ``Development of the Next-Generation Integrated Wired/Wireless Communication Gateway (X-Gateway).''
\fi  %
\ifOnCueArxiv\else\clearpage\nobalance\fi
\section*{LLM Usage Statement}
Large language models (Claude Opus 5.5 and GPT-6.1 Sol) were used to improve manuscript clarity, grammar, spelling, and style and to refine figures and tables. The authors take full responsibility for all scientific content, interpretations, and conclusions.
\par

\setcounter{topnumber}{3}
{\small\renewcommand{\refname}{References}\bibliographystyle{ieeenat_fullname}
\bibliography{references}}

\begin{thebibliography}{57}
\providecommand{\natexlab}[1]{#1}
\providecommand{\url}[1]{\texttt{#1}}
\expandafter\ifx\csname urlstyle\endcsname\relax
  \providecommand{\doi}[1]{doi: #1}\else
  \providecommand{\doi}{doi: \begingroup \urlstyle{rm}\Url}\fi

\bibitem[{AASHTO}(2018)]{aashto2018}
{AASHTO}.
\newblock \emph{{A Policy on Geometric Design of Highways and Streets}}.
\newblock American Association of State Highway and Transportation Officials, Washington, DC, 7th edition, 2018.

\bibitem[Caesar et~al.(2021)Caesar, Kabzan, Tan, Fong, Wolff, Lang, Fletcher, Beijbom, and Omari]{nuplan}
Holger Caesar, Juraj Kabzan, Kok~Seang Tan, Whye~Kit Fong, Eric Wolff, Alex Lang, Luke Fletcher, Oscar Beijbom, and Sammy Omari.
\newblock {nuPlan: A Closed-Loop ML-Based Planning Benchmark for Autonomous Vehicles}.
\newblock In \emph{CVPR ADP3 Workshop}, 2021.
\newblock arXiv preprint arXiv:2106.11810.

\bibitem[Chang et~al.(2024)Chang, Pittaluga, Tomizuka, Zhan, and Chandraker]{safesim}
Wei-Jer Chang, Francesco Pittaluga, Masayoshi Tomizuka, Wei Zhan, and Manmohan Chandraker.
\newblock {SAFE-SIM: Safety-Critical Closed-Loop Traffic Simulation with Diffusion-Controllable Adversaries}.
\newblock In \emph{European Conference on Computer Vision (ECCV)}, pages 242--258, 2024.

\bibitem[Chen et~al.(2025)Chen, Lei, Cheng, Wu, Sun, and Zheng]{frea}
Keyu Chen, Yuheng Lei, Hao Cheng, Haoran Wu, Wenchao Sun, and Sifa Zheng.
\newblock {FREA: Feasibility-Guided Generation of Safety-Critical Scenarios with Reasonable Adversariality}.
\newblock In \emph{Proceedings of the 8th Conference on Robot Learning (CoRL)}, pages 566--586. PMLR, 2025.

\bibitem[Chen et~al.(2026)Chen, Kung, Chang, Chiu, and Chen]{collide}
Pin-Lun Chen, Chi-Hsi Kung, Che-Han Chang, Wei-Chen Chiu, and Yi-Ting Chen.
\newblock {Controllable Collision Scenario Generation via Collision Pattern Prediction}.
\newblock In \emph{IEEE International Conference on Robotics and Automation (ICRA)}, pages 18434--18441, 2026.

\bibitem[Christianos et~al.(2023)Christianos, Karkus, Ivanovic, Albrecht, and Pavone]{bivo}
Filippos Christianos, Peter Karkus, Boris Ivanovic, Stefano~V. Albrecht, and Marco Pavone.
\newblock {Planning with Occluded Traffic Agents using Bi-Level Variational Occlusion Models}.
\newblock In \emph{IEEE International Conference on Robotics and Automation (ICRA)}, pages 5558--5565, 2023.

\bibitem[Cui et~al.(2026)Cui, Liang, Yang, Tang, and Cui]{safer}
Jinlong Cui, Fenghua Liang, Guo Yang, Chengcheng Tang, and Jianxun Cui.
\newblock {SaFeR: Safety-Critical Scenario Generation for Autonomous Driving Test via Feasibility-Constrained Token Resampling}.
\newblock arXiv preprint arXiv:2603.04071, 2026.

\bibitem[Dauner et~al.(2023)Dauner, Hallgarten, Geiger, and Chitta]{pdm}
Daniel Dauner, Marcel Hallgarten, Andreas Geiger, and Kashyap Chitta.
\newblock {Parting with Misconceptions about Learning-based Vehicle Motion Planning}.
\newblock In \emph{Proceedings of the 7th Conference on Robot Learning (CoRL)}, pages 1268--1281. PMLR, 2023.

\bibitem[Davidson et~al.(2026)Davidson, Ramanan, and Peri]{refav}
Cainan Davidson, Deva Ramanan, and Neehar Peri.
\newblock {RefAV: Towards Planning-Centric Scenario Mining}.
\newblock In \emph{IEEE/CVF Conference on Computer Vision and Pattern Recognition (CVPR)}, pages 21537--21548, 2026.

\bibitem[Ding et~al.(2019)Ding, Wang, and Zhao]{mtg}
Wenhao Ding, Wenshuo Wang, and Ding Zhao.
\newblock {A Multi-Vehicle Trajectories Generator to Simulate Vehicle-to-Vehicle Encountering Scenarios}.
\newblock In \emph{IEEE International Conference on Robotics and Automation (ICRA)}, pages 4255--4261, 2019.

\bibitem[Ding et~al.(2020)Ding, Chen, Xu, and Zhao]{learningcollide}
Wenhao Ding, Baiming Chen, Minjun Xu, and Ding Zhao.
\newblock {Learning to Collide: An Adaptive Safety-Critical Scenarios Generating Method}.
\newblock In \emph{IEEE/RSJ International Conference on Intelligent Robots and Systems (IROS)}, pages 2243--2250, 2020.

\bibitem[Ding et~al.(2023)Ding, Xu, Arief, Lin, Li, and Zhao]{dingsurvey}
Wenhao Ding, Chejian Xu, Mansur Arief, Haohong Lin, Bo Li, and Ding Zhao.
\newblock {A Survey on Safety-Critical Driving Scenario Generation---A Methodological Perspective}.
\newblock \emph{IEEE Transactions on Intelligent Transportation Systems}, 24\penalty0 (7):\penalty0 6971--6988, 2023.

\bibitem[Dykhne(2024)]{dykhne2024}
Eli-Henry Dykhne.
\newblock {Automated Generation of Dynamic Occlusion-Caused Collisions}.
\newblock {MMath} thesis, University of Waterloo, 2024.

\bibitem[Ettinger et~al.(2021)Ettinger, Cheng, Caine, Liu, Zhao, Pradhan, Chai, Sapp, Qi, Zhou, Yang, Chouard, Sun, Ngiam, Vasudevan, McCauley, Shlens, and Anguelov]{womd}
Scott Ettinger, Shuyang Cheng, Benjamin Caine, Chenxi Liu, Hang Zhao, Sabeek Pradhan, Yuning Chai, Ben Sapp, Charles Qi, Yin Zhou, Zoey Yang, Aurelien Chouard, Pei Sun, Jiquan Ngiam, Vijay Vasudevan, Alexander McCauley, Jonathon Shlens, and Dragomir Anguelov.
\newblock {Large Scale Interactive Motion Forecasting for Autonomous Driving: The Waymo Open Motion Dataset}.
\newblock In \emph{IEEE/CVF International Conference on Computer Vision (ICCV)}, pages 9690--9699, 2021.

\bibitem[{Euro NCAP}(2024)]{euroncap2024}
{Euro NCAP}.
\newblock {Test Protocol -- AEB/LSS VRU Systems, Implementation 2023, Version 4.5.1}, 2024.

\bibitem[Feng et~al.(2021)Feng, Yan, Sun, Feng, and Liu]{nade}
Shuo Feng, Xintao Yan, Haowei Sun, Yiheng Feng, and Henry~X. Liu.
\newblock {Intelligent Driving Intelligence Test for Autonomous Vehicles with Naturalistic and Adversarial Environment}.
\newblock \emph{Nature Communications}, 12\penalty0 (1):\penalty0 748, 2021.

\bibitem[Feng et~al.(2023)Feng, Sun, Yan, Zhu, Zou, Shen, and Liu]{d2rl}
Shuo Feng, Haowei Sun, Xintao Yan, Haojie Zhu, Zhengxia Zou, Shengyin Shen, and Henry~X. Liu.
\newblock {Dense Reinforcement Learning for Safety Validation of Autonomous Vehicles}.
\newblock \emph{Nature}, 615\penalty0 (7953):\penalty0 620--627, 2023.

\bibitem[Fisher et~al.(2006)Fisher, Pollatsek, and Pradhan]{fisher2006}
D.~L. Fisher, A.~P. Pollatsek, and A. Pradhan.
\newblock {Can Novice Drivers Be Trained to Scan for Information That Will Reduce Their Likelihood of a Crash?}
\newblock \emph{Injury Prevention}, 12\penalty0 (suppl 1):\penalty0 i25--i29, 2006.

\bibitem[Gehrig and Scaramuzza(2024)]{eventvision}
Daniel Gehrig and Davide Scaramuzza.
\newblock {Low-Latency Automotive Vision with Event Cameras}.
\newblock \emph{Nature}, 629\penalty0 (8014):\penalty0 1034--1040, 2024.

\bibitem[Ghodsi et~al.(2021)Ghodsi, Hari, Frosio, Tsai, Troccoli, Keckler, Garg, and Anandkumar]{ghodsi2021}
Zahra Ghodsi, Siva Kumar~Sastry Hari, Iuri Frosio, Timothy Tsai, Alejandro Troccoli, Stephen~W. Keckler, Siddharth Garg, and Anima Anandkumar.
\newblock {Generating and Characterizing Scenarios for Safety Testing of Autonomous Vehicles}.
\newblock In \emph{IEEE Intelligent Vehicles Symposium (IV)}, pages 157--164, 2021.

\bibitem[Gipps(1981)]{gipps1981}
P.~G. Gipps.
\newblock {A Behavioural Car-Following Model for Computer Simulation}.
\newblock \emph{Transportation Research Part B: Methodological}, 15\penalty0 (2):\penalty0 105--111, 1981.

\bibitem[Gulino et~al.(2023)Gulino, Fu, Luo, Tucker, Bronstein, Lu, Harb, Pan, Wang, Chen, Co-Reyes, Agarwal, Roelofs, Lu, Montali, Mougin, Yang, White, Faust, McAllister, Anguelov, and Sapp]{waymax}
Cole Gulino, Justin Fu, Wenjie Luo, George Tucker, Eli Bronstein, Yiren Lu, Jean Harb, Xinlei Pan, Yan Wang, Xiangyu Chen, John~D. Co-Reyes, Rishabh Agarwal, Rebecca Roelofs, Yao Lu, Nico Montali, Paul Mougin, Zoey Yang, Brandyn White, Aleksandra Faust, Rowan McAllister, Dragomir Anguelov, and Benjamin Sapp.
\newblock {Waymax: An Accelerated, Data-Driven Simulator for Large-Scale Autonomous Driving Research}.
\newblock In \emph{Advances in Neural Information Processing Systems (NeurIPS), Datasets and Benchmarks Track}, 2023.

\bibitem[Hallgarten et~al.(2024)Hallgarten, Zapata, Stoll, Renz, and Zell]{interplan}
Marcel Hallgarten, Julian Zapata, Martin Stoll, Katrin Renz, and Andreas Zell.
\newblock {Can Vehicle Motion Planning Generalize to Realistic Long-tail Scenarios?}
\newblock In \emph{IEEE/RSJ International Conference on Intelligent Robots and Systems (IROS)}, pages 5388--5395, 2024.

\bibitem[Hanselmann et~al.(2022)Hanselmann, Renz, Chitta, Bhattacharyya, and Geiger]{king}
Niklas Hanselmann, Katrin Renz, Kashyap Chitta, Apratim Bhattacharyya, and Andreas Geiger.
\newblock {KING: Generating Safety-Critical Driving Scenarios for Robust Imitation via Kinematics Gradients}.
\newblock In \emph{European Conference on Computer Vision (ECCV)}, pages 335--352, 2022.

\bibitem[Jia et~al.(2026)Jia, Su, Bao, Hong, Gao, Gan, and Ding]{urm}
Jie Jia, Yaofeng Su, Zeyu Bao, Yun Hong, Bingzhao Gao, Zhongxue Gan, and Wenchao Ding.
\newblock {Learning A Unified Risk Map for Autonomous Driving in Partially Observable Environments}.
\newblock \emph{IEEE Robotics and Automation Letters}, 11\penalty0 (4):\penalty0 4457--4464, 2026.

\bibitem[Jiang et~al.(2023)Jiang, Cornman, Park, Sapp, Zhou, and Anguelov]{motiondiffuser}
Chiyu~Max Jiang, Andre Cornman, Cheolho Park, Benjamin Sapp, Yin Zhou, and Dragomir Anguelov.
\newblock {MotionDiffuser: Controllable Multi-Agent Motion Prediction Using Diffusion}.
\newblock In \emph{IEEE/CVF Conference on Computer Vision and Pattern Recognition (CVPR)}, pages 9644--9653, 2023.

\bibitem[Kahn et~al.(2022)Kahn, Sarkar, and Czarnecki]{kahn2022}
Maximilian Kahn, Atrisha Sarkar, and Krzysztof Czarnecki.
\newblock {I Know You Can't See Me: Dynamic Occlusion-Aware Safety Validation of Strategic Planners for Autonomous Vehicles Using Hypergames}.
\newblock In \emph{IEEE International Conference on Robotics and Automation (ICRA)}, pages 11202--11208, 2022.

\bibitem[Li et~al.(2025)Li, Li, Lan, Xie, Zhang, Liu, Wu, Yu, and Alvarez]{hydramdppp}
Kailin Li, Zhenxin Li, Shiyi Lan, Yuan Xie, Zhizhong Zhang, Jiayi Liu, Zuxuan Wu, Zhiding Yu, and Jose~M. Alvarez.
\newblock {Hydra-MDP++: Advancing End-to-End Driving via Expert-Guided Hydra-Distillation}.
\newblock arXiv preprint arXiv:2503.12820, 2025.

\bibitem[Li et~al.(2023)Li, Peng, Feng, Zhang, Xue, and Zhou]{metadrive}
Quanyi Li, Zhenghao Peng, Lan Feng, Qihang Zhang, Zhenghai Xue, and Bolei Zhou.
\newblock {MetaDrive: Composing Diverse Driving Scenarios for Generalizable Reinforcement Learning}.
\newblock \emph{IEEE Transactions on Pattern Analysis and Machine Intelligence}, 45\penalty0 (3):\penalty0 3461--3475, 2023.

\bibitem[Mahjourian et~al.(2024)Mahjourian, Mu, Likhosherstov, Mougin, Huang, Messias, and Whiteson]{unigen}
Reza Mahjourian, Rongbing Mu, Valerii Likhosherstov, Paul Mougin, Xiukun Huang, Joao Messias, and Shimon Whiteson.
\newblock {UniGen: Unified Modeling of Initial Agent States and Trajectories for Generating Autonomous Driving Scenarios}.
\newblock In \emph{IEEE International Conference on Robotics and Automation (ICRA)}, pages 16367--16373, 2024.

\bibitem[Mordfin(1975)]{mordfin1975}
Leonard Mordfin.
\newblock {The CPSC Road Test of Bicycle Braking Performance: Kinetic and Error Analyses}.
\newblock Report NBSIR 75-786, National Bureau of Standards, 1975.

\bibitem[{National Highway Traffic Safety Administration}(2019)]{nhtsa2019paeb}
{National Highway Traffic Safety Administration}.
\newblock {Pedestrian Automatic Emergency Brake System Confirmation Test}.
\newblock Technical report, {U.S. Department of Transportation}, 2019.
\newblock Working draft.

\bibitem[Nie et~al.(2026)Nie, Mei, Tang, He, Sun, Shi, Ma, and Sun]{sage}
Tong Nie, Yuewen Mei, Yihong Tang, Junlin He, Jie Sun, Haotian Shi, Wei Ma, and Jian Sun.
\newblock {Steerable Adversarial Scenario Generation through Test-Time Preference Alignment}.
\newblock In \emph{International Conference on Learning Representations (ICLR)}, 2026.

\bibitem[Priisalu et~al.(2022)Priisalu, Pirinen, Paduraru, and Sminchisescu]{priisalu2022}
Maria Priisalu, Aleksis Pirinen, Ciprian Paduraru, and Cristian Sminchisescu.
\newblock {Generating Scenarios with Diverse Pedestrian Behaviors for Autonomous Vehicle Testing}.
\newblock In \emph{Proceedings of the 5th Conference on Robot Learning (CoRL)}, pages 1247--1258. PMLR, 2022.

\bibitem[Ransiek et~al.(2024)Ransiek, Plaum, Langner, and Sax]{goose}
Joshua Ransiek, Johannes Plaum, Jacob Langner, and Eric Sax.
\newblock {GOOSE: Goal-Conditioned Reinforcement Learning for Safety-Critical Scenario Generation}.
\newblock In \emph{IEEE International Conference on Intelligent Transportation Systems (ITSC)}, pages 2651--2658, 2024.

\bibitem[Rempe et~al.(2022)Rempe, Philion, Guibas, Fidler, and Litany]{strive}
Davis Rempe, Jonah Philion, Leonidas~J. Guibas, Sanja Fidler, and Or Litany.
\newblock {Generating Useful Accident-Prone Driving Scenarios via a Learned Traffic Prior}.
\newblock In \emph{IEEE/CVF Conference on Computer Vision and Pattern Recognition (CVPR)}, pages 17284--17294, 2022.

\bibitem[Renz et~al.(2023)Renz, Chitta, Mercea, Koepke, Akata, and Geiger]{plant}
Katrin Renz, Kashyap Chitta, Otniel-Bogdan Mercea, A.~Sophia Koepke, Zeynep Akata, and Andreas Geiger.
\newblock {PlanT: Explainable Planning Transformers via Object-Level Representations}.
\newblock In \emph{Proceedings of the 6th Conference on Robot Learning (CoRL)}, pages 459--470. PMLR, 2023.

\bibitem[S\'anchez et~al.(2022)S\'anchez, Nyberg, Pek, Tumova, and T\"orngren]{foresee}
Jos\'e~Manuel~Gaspar S\'anchez, Truls Nyberg, Christian Pek, Jana Tumova, and Martin T\"orngren.
\newblock {Foresee the Unseen: Sequential Reasoning about Hidden Obstacles for Safe Driving}.
\newblock In \emph{IEEE Intelligent Vehicles Symposium (IV)}, pages 255--264, 2022.

\bibitem[Scanlon et~al.(2022)Scanlon, Kusano, Engstr{\"o}m, and Victor]{nieon}
John~M. Scanlon, Kristofer~D. Kusano, Johan Engstr{\"o}m, and Trent Victor.
\newblock {Collision Avoidance Effectiveness of an Automated Driving System Using a Human Driver Behavior Reference Model in Reconstructed Fatal Collisions}.
\newblock Technical report, Waymo LLC, 2022.

\bibitem[Shalev-Shwartz et~al.(2017)Shalev-Shwartz, Shammah, and Shashua]{rss2017}
Shai Shalev-Shwartz, Shaked Shammah, and Amnon Shashua.
\newblock {On a Formal Model of Safe and Scalable Self-driving Cars}.
\newblock arXiv preprint arXiv:1708.06374, 2017.

\bibitem[Shi et~al.(2022)Shi, Jiang, Dai, and Schiele]{mtr}
Shaoshuai Shi, Li Jiang, Dengxin Dai, and Bernt Schiele.
\newblock {Motion Transformer with Global Intention Localization and Local Movement Refinement}.
\newblock In \emph{Advances in Neural Information Processing Systems (NeurIPS)}, pages 6531--6543, 2022.

\bibitem[Stavridis et~al.(2022)Stavridis, Economou, Walker, Bissas, Tsopanidou, and Paradisis]{stavridis2022}
Ioannis Stavridis, Theodosia Economou, Josh Walker, Athanassios Bissas, Angela Tsopanidou, and Giorgos Paradisis.
\newblock {Sprint Mechanical Characteristics of Sub-Elite and Recreational Sprinters}.
\newblock \emph{Journal of Physical Education and Sport}, 22\penalty0 (5):\penalty0 1126--1133, 2022.

\bibitem[Stoler et~al.(2025)Stoler, Navarro, Francis, and Oh]{seal}
Benjamin Stoler, Ingrid Navarro, Jonathan Francis, and Jean Oh.
\newblock {SEAL: Towards Safe Autonomous Driving via Skill-Enabled Adversary Learning for Closed-Loop Scenario Generation}.
\newblock \emph{IEEE Robotics and Automation Letters}, 10\penalty0 (9):\penalty0 9320--9327, 2025.

\bibitem[{\v{S}}tuhec et~al.(2023){\v{S}}tuhec, Planj{\v{s}}ek, {\v{C}}oh, and Mackala]{stuhec2023}
Stanislav {\v{S}}tuhec, Peter Planj{\v{s}}ek, Milan {\v{C}}oh, and Krzysztof Mackala.
\newblock {Multicomponent Velocity Measurement for Linear Sprinting: Usain Bolt's 100 m World-Record Analysis}.
\newblock \emph{Bioengineering}, 10\penalty0 (11):\penalty0 1254, 2023.

\bibitem[Tan et~al.(2026)Tan, Lei, and Ukkusuri]{hcmts}
Ruichen Tan, Zengxiang Lei, and Satish Ukkusuri.
\newblock {Worst-Case Hidden-Vehicle Trajectory Search in Spatiotemporal Occlusion Regions}.
\newblock In \emph{ECCV Workshop on Safe and Defensive Autonomous Driving}, 2026.

\bibitem[Tan et~al.(2025)Tan, Lambert, Jeon, Kulshrestha, Bai, Luo, Anguelov, Tan, and Jiang]{scenediffuserpp}
Shuhan Tan, John Lambert, Hong Jeon, Sakshum Kulshrestha, Yijing Bai, Jing Luo, Dragomir Anguelov, Mingxing Tan, and Chiyu~Max Jiang.
\newblock {SceneDiffuser++: City-Scale Traffic Simulation via a Generative World Model}.
\newblock In \emph{IEEE/CVF Conference on Computer Vision and Pattern Recognition (CVPR)}, pages 1570--1580, 2025.

\bibitem[Treiber et~al.(2000)Treiber, Hennecke, and Helbing]{idm}
Martin Treiber, Ansgar Hennecke, and Dirk Helbing.
\newblock {Congested Traffic States in Empirical Observations and Microscopic Simulations}.
\newblock \emph{Physical Review E}, 62\penalty0 (2):\penalty0 1805--1824, 2000.

\bibitem[{UNECE}(2021)]{unr157}
{UNECE}.
\newblock {UN Regulation No 157 -- Uniform Provisions Concerning the Approval of Vehicles with Regards to Automated Lane Keeping Systems}.
\newblock Official Journal of the European Union, L 82:75--137, 2021.

\bibitem[Wang et~al.(2021)Wang, Pun, Tu, Manivasagam, Sadat, Casas, Ren, and Urtasun]{advsim}
Jingkang Wang, Ava Pun, James Tu, Sivabalan Manivasagam, Abbas Sadat, Sergio Casas, Mengye Ren, and Raquel Urtasun.
\newblock {AdvSim: Generating Safety-Critical Scenarios for Self-Driving Vehicles}.
\newblock In \emph{IEEE/CVF Conference on Computer Vision and Pattern Recognition (CVPR)}, pages 9904--9913, 2021.

\bibitem[Xiao et~al.(2025)Xiao, Liu, Yang, Li, Ye, Yang, and Wang]{latentdriver}
Lingyu Xiao, Jiang-Jiang Liu, Sen Yang, Xiaofan Li, Xiaoqing Ye, Wankou Yang, and Jingdong Wang.
\newblock {Learning Multiple Probabilistic Decisions from Latent World Model in Autonomous Driving}.
\newblock In \emph{IEEE International Conference on Robotics and Automation (ICRA)}, pages 1279--1285, 2025.

\bibitem[Xu et~al.(2022)Xu, Ding, Lyu, Liu, Wang, He, Hu, Zhao, and Li]{safebench}
Chejian Xu, Wenhao Ding, Weijie Lyu, Zuxin Liu, Shuai Wang, Yihan He, Hanjiang Hu, Ding Zhao, and Bo Li.
\newblock {SafeBench: A Benchmarking Platform for Safety Evaluation of Autonomous Vehicles}.
\newblock In \emph{Advances in Neural Information Processing Systems (NeurIPS)}, pages 25667--25682, 2022.

\bibitem[Xu et~al.(2025)Xu, Petiushko, Zhao, and Li]{diffscene}
Chejian Xu, Aleksandr Petiushko, Ding Zhao, and Bo Li.
\newblock {DiffScene: Diffusion-Based Safety-Critical Scenario Generation for Autonomous Vehicles}.
\newblock In \emph{AAAI Conference on Artificial Intelligence (AAAI)}, pages 8797--8805, 2025.

\bibitem[Yan et~al.(2023)Yan, Zou, Feng, Zhu, Sun, and Liu]{neuralnde}
Xintao Yan, Zhengxia Zou, Shuo Feng, Haojie Zhu, Haowei Sun, and Henry~X. Liu.
\newblock {Learning Naturalistic Driving Environment with Statistical Realism}.
\newblock \emph{Nature Communications}, 14\penalty0 (1):\penalty0 2037, 2023.

\bibitem[Zhang et~al.(2024)Zhang, Xu, and Li]{chatscene}
Jiawei Zhang, Chejian Xu, and Bo Li.
\newblock {ChatScene: Knowledge-Enabled Safety-Critical Scenario Generation for Autonomous Vehicles}.
\newblock In \emph{IEEE/CVF Conference on Computer Vision and Pattern Recognition (CVPR)}, pages 15459--15469, 2024.

\bibitem[Zhang et~al.(2023{\natexlab{a}})Zhang, Peng, Li, and Zhou]{cat}
Linrui Zhang, Zhenghao Peng, Quanyi Li, and Bolei Zhou.
\newblock {CAT: Closed-loop Adversarial Training for Safe End-to-End Driving}.
\newblock In \emph{Proceedings of the 7th Conference on Robot Learning (CoRL)}, pages 2357--2372. PMLR, 2023{\natexlab{a}}.

\bibitem[Zhang et~al.(2023{\natexlab{b}})Zhang, Jiang, and Chen]{mpcocclusion}
Zheyu Zhang, Jingjing Jiang, and Wen-Hua Chen.
\newblock {Unveiling the Road Ahead: An MPC-Based Approach for Autonomous Intersection Navigation with Occlusion}.
\newblock In \emph{IEEE International Automated Vehicle Validation Conference (IAVVC)}, pages 1--6, 2023{\natexlab{b}}.

\bibitem[Zhong et~al.(2023)Zhong, Rempe, Xu, Chen, Veer, Che, Ray, and Pavone]{ctg}
Ziyuan Zhong, Davis Rempe, Danfei Xu, Yuxiao Chen, Sushant Veer, Tong Che, Baishakhi Ray, and Marco Pavone.
\newblock {Guided Conditional Diffusion for Controllable Traffic Simulation}.
\newblock In \emph{IEEE International Conference on Robotics and Automation (ICRA)}, pages 3560--3566, 2023.

\end{thebibliography}

\ifOnCueArxiv
\clearpage
\setcounter{table}{0}\setcounter{figure}{0}\setcounter{equation}{0}\setcounter{algorithm}{0}
\renewcommand{\thetable}{S\arabic{table}}\renewcommand{\theHtable}{S\arabic{table}}
\renewcommand{\thefigure}{S\arabic{figure}}\renewcommand{\theHfigure}{S\arabic{figure}}
\renewcommand{\theequation}{S\arabic{equation}}\renewcommand{\theHequation}{S\arabic{equation}}
\renewcommand{\thealgorithm}{S\arabic{algorithm}}\renewcommand{\theHalgorithm}{S\arabic{algorithm}}
\definecolor{OnCueGain}{HTML}{DCECF9}
\definecolor{OnCueCost}{HTML}{F9E0DB}
\makeatletter
\renewcommand{\oncueci}[1]{\unskip\nobreak\hspace{1.2pt}{\normalfont\fontsize{5.4}{6.0}\selectfont\color{black!38}\oncue@ci#1\@nil}}
\def\oncue@ci#1,#2\@nil{\makebox[16.5pt][r]{[#1,}\makebox[16.5pt][r]{#2]}}
\makeatother

\newcommand{\oncuesupplementtitle}{\newpage\twocolumn[{\centering
  {\Large\textbf{\raisebox{-0.18\height}{\includegraphics[height=19pt]{figures/oncue-logo.pdf}}: \oncuetitlerest}}\\
  \vspace{0.5em}{\Large Supplementary Material}\\
  \vspace{1.0em}}]}
\ifOnCueArxiv\oncuesupplementtitle\else\clearpage\oncuesupplementtitle\switchlinenumbers\fi
\setcounter{page}{1}\renewcommand{\thepage}{S\arabic{page}}  %
\hyphenation{commands Anticipatory recovers execution location highest returns observation protocol lowers extends generation reserved occurs difference waypoint}
\ifOnCueArxiv\else\setlength{\linenumbersep}{.75cm}\fi
\captionsetup[table]{position=top,skip=4pt}
\newcommand{\oncuetablenote}[1]{\par\vspace{3pt}{\footnotesize\leftskip=0pt\rightskip=0pt\parfillskip=0pt plus 1fil\noindent #1\par}}
\noindent Appendices A to C describe how tests are generated, labeled, and evaluated. Appendices D to F follow the main paper\textquotesingle{}s questions: D extends the generation results (Q1 to Q3), E tests the response labels (Q4), and F extends the planner and training results (Q5 and Q6). Appendix G shows six generated encounters. Unless a caption says otherwise, light blue in tables marks favorable or best values, darker shades rank the top three where shown, and underlining marks values quoted in the main paper. Table, figure, and equation numbers without the prefix S refer to the main \ifOnCueArxiv\oncuetail{paper.}\else paper, and citation numbers to this supplement\textquotesingle{}s reference \oncuetail{list.}\fi
\makeatletter
\newcommand{\oncuetoc}[4]{\csname l@#1\endcsname{\numberline{#2}#3}{\pageref{#4}}}
\renewcommand{\@pnumwidth}{2em}\renewcommand{\@tocrmarg}{3em plus 1fil}  %
\makeatother
\section*{Contents}
\oncuetoc{section}{A}{Generation Pipeline and Model}{toc:A}
\oncuetoc{subsection}{A.1}{Notation and Time Reference}{toc:A.1}
\oncuetoc{subsection}{A.2}{Starts, Visibility, and Trajectory Selection}{toc:A.2}
\oncuetoc{subsection}{A.3}{Trajectory Generator and Training}{toc:A.3}
\oncuetoc{section}{B}{Safe-Response Labeling}{toc:B}
\oncuetoc{subsection}{B.1}{Evaluation Interval and Coverage}{toc:B.1}
\oncuetoc{subsection}{B.2}{Reference Ego and Response Library}{toc:B.2}
\oncuetoc{subsection}{B.3}{Event Times and Response Coverage}{toc:B.3}
\oncuetoc{section}{C}{Evaluation Protocol and Metrics}{toc:C}
\oncuetoc{subsection}{C.1}{Comparison Scope and Intervals}{toc:C.1}
\oncuetoc{subsection}{C.2}{Motion-Quality Metrics}{toc:C.2}
\oncuetoc{subsection}{C.3}{Planners, Target Execution, and Outcomes}{toc:C.3}
\oncuetoc{section}{D}{Additional Generation Results}{toc:D}
\oncuetoc{subsection}{D.1}{Where Requests Are Lost {\color{black!55}(Q1,~Q3)}}{toc:D.1}
\oncuetoc{subsection}{D.2}{Timing before Checks and Selection {\color{black!55}(Q1,~Q3)}}{toc:D.2}
\oncuetoc{subsection}{D.3}{Timing across Generators {\color{black!55}(Q2)}}{toc:D.3}
\oncuetoc{subsection}{D.4}{Motion Quality across Generators {\color{black!55}(Q2)}}{toc:D.4}
\oncuetoc{subsection}{D.5}{Missing Events versus Failed Action Search {\color{black!55}(Q2,~Q4)}}{toc:D.5}
\oncuetoc{section}{E}{Response-Label Sensitivity}{toc:E}
\oncuetoc{subsection}{E.1}{Varying the Oracle Settings {\color{black!55}(Q4)}}{toc:E.1}
\oncuetoc{subsection}{E.2}{Varying the Reference Ego {\color{black!55}(Q4)}}{toc:E.2}
\oncuetoc{section}{F}{Planner Behavior, Training, and Transfer}{toc:F}
\oncuetoc{subsection}{F.1}{Visibility and Avoidance Opportunities {\color{black!55}(Q5)}}{toc:F.1}
\oncuetoc{subsection}{F.2}{Observation Contrast with Fixed Timing {\color{black!55}(Q5)}}{toc:F.2}
\oncuetoc{subsection}{F.3}{More Response Time in the Same Situations {\color{black!55}(Q5)}}{toc:F.3}
\oncuetoc{subsection}{F.4}{Results without Target-Background Contact {\color{black!55}(Q5)}}{toc:F.4}
\oncuetoc{subsection}{F.5}{Training a Planner on the Tests {\color{black!55}(Q6)}}{toc:F.5}
\oncuetoc{subsubsection}{F.5.1}{Slower Driving and Triggered Starts}{toc:F.5.1}
\oncuetoc{subsubsection}{F.5.2}{Transfer to Other Tests, Planners, and Conditions}{toc:F.5.2}
\oncuetoc{subsubsection}{F.5.3}{Costs and the Source of Hesitation}{toc:F.5.3}
\oncuetoc{subsubsection}{F.5.4}{Natural Logged Encounters}{toc:F.5.4}
\oncuetoc{section}{G}{Encounter Examples}{toc:G}
\par
\section*{Appendix A. Generation Pipeline and Model}\label{toc:A}
This appendix follows the generation pipeline: notation, starts and visibility, and the conditional trajectory model. Figure~\ref{fig:test-construction} walks through these steps on one test.
\begin{figure*}[t]
\centering
{\footnotesize
\raisebox{-0.22em}{\includegraphics[height=1.05em]{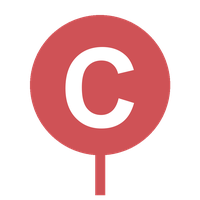}}\,contact C\quad
\raisebox{-0.22em}{\includegraphics[height=1.05em]{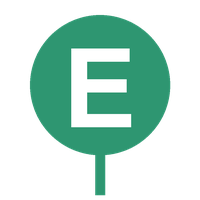}}\,exposure E\quad
\raisebox{-0.22em}{\includegraphics[height=1.05em]{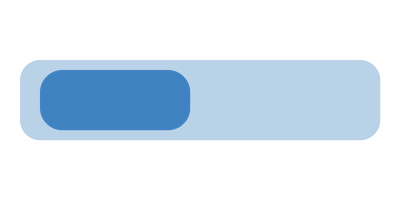}}\,reference ego and path\quad
\raisebox{-0.22em}{\includegraphics[height=1.05em]{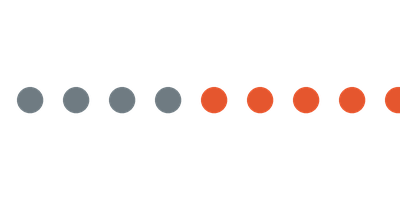}}\,target route (gray: hidden)\quad
\raisebox{-0.22em}{\includegraphics[height=1.05em]{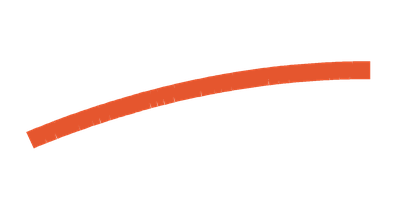}}\,generated trajectory\quad
\raisebox{-0.22em}{\includegraphics[height=1.05em]{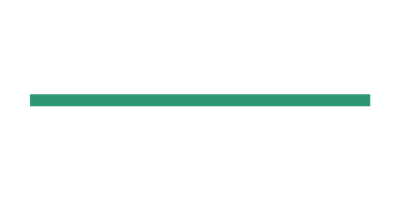}}\,line of sight}\\[4pt]
\setlength{\tabcolsep}{1pt}
\begin{tabular}{@{}cccc@{}}
\includegraphics[width=.244\textwidth]{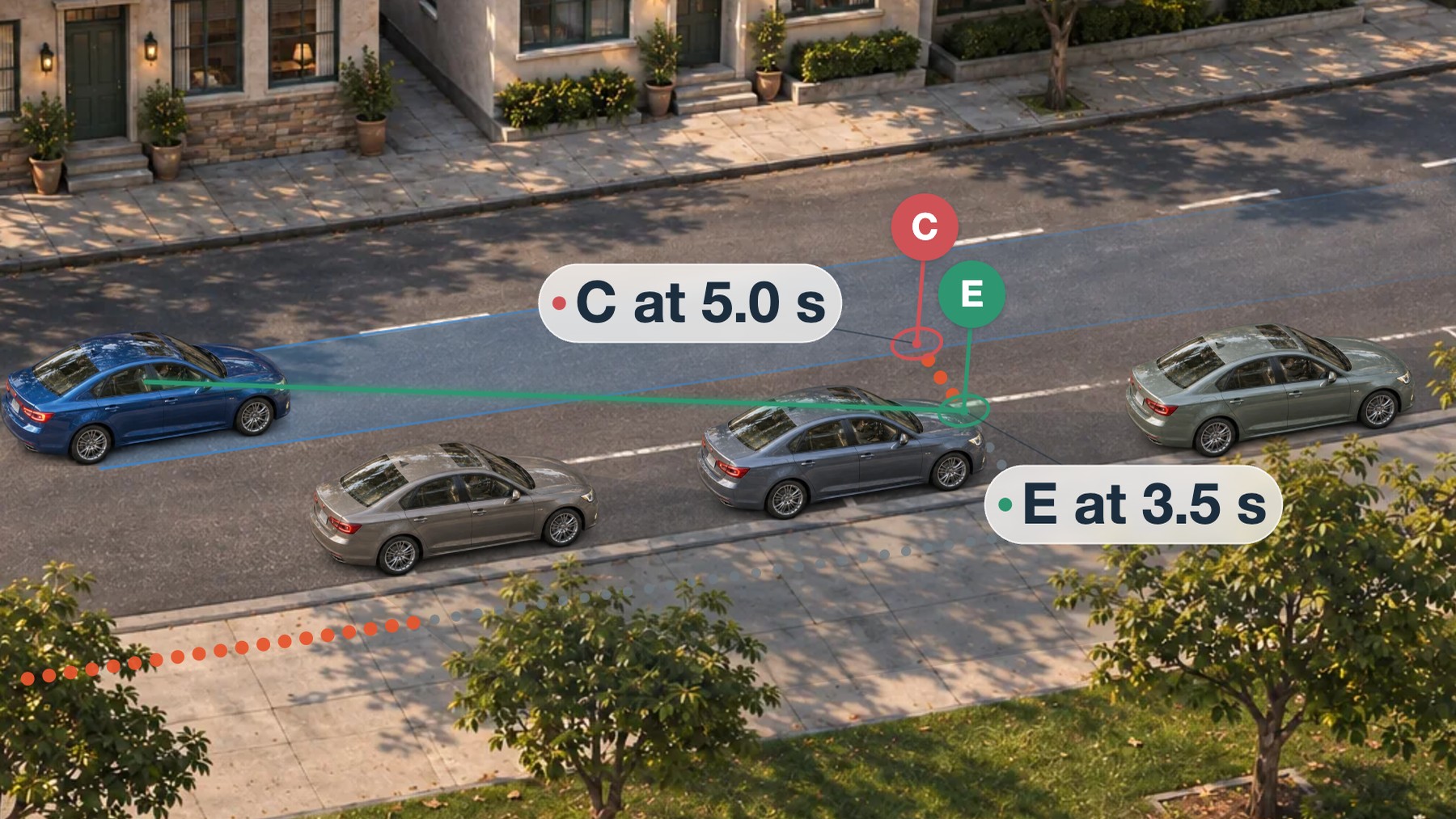} &
\includegraphics[width=.244\textwidth]{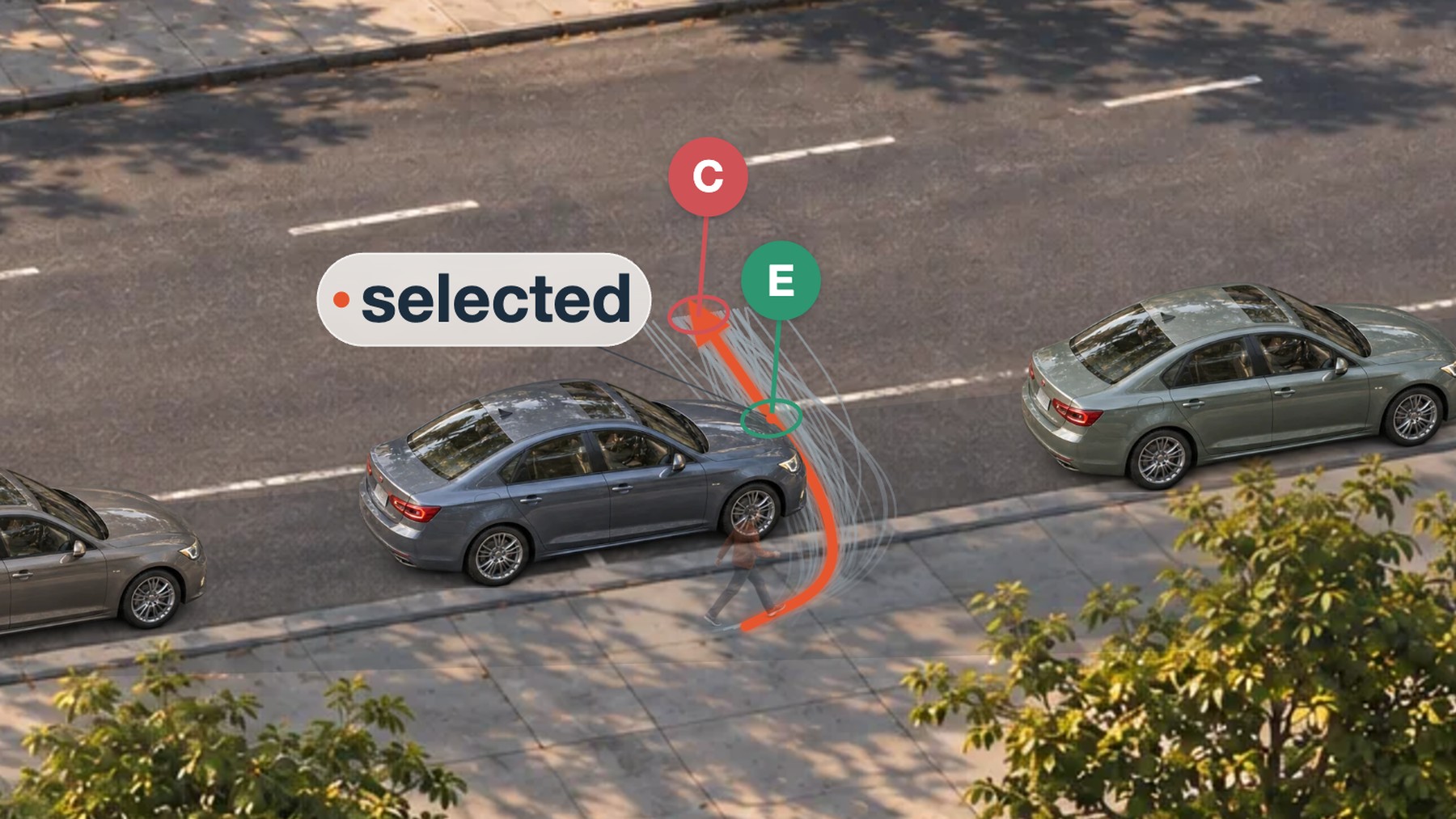} &
\includegraphics[width=.244\textwidth]{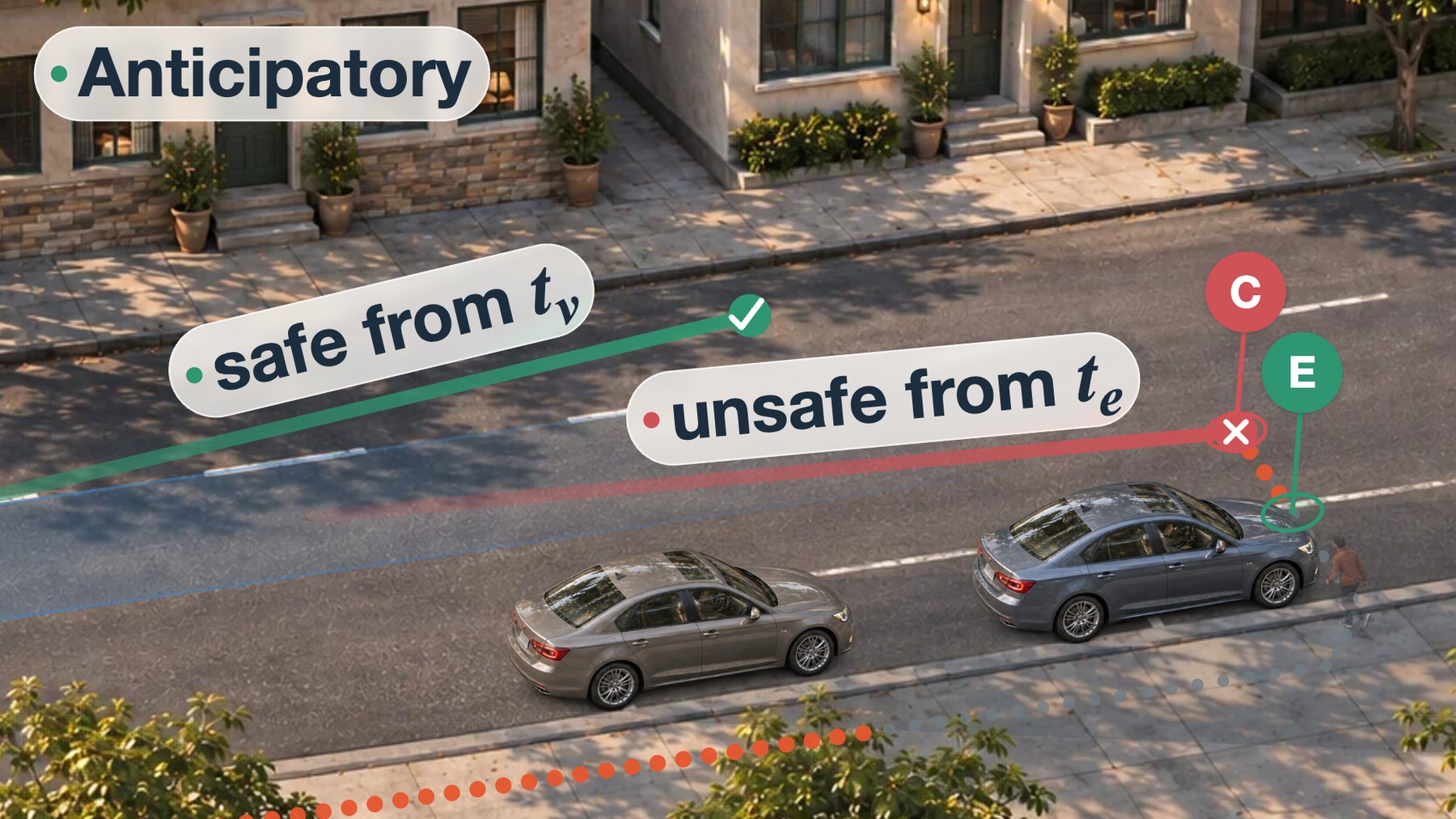} &
\includegraphics[width=.244\textwidth]{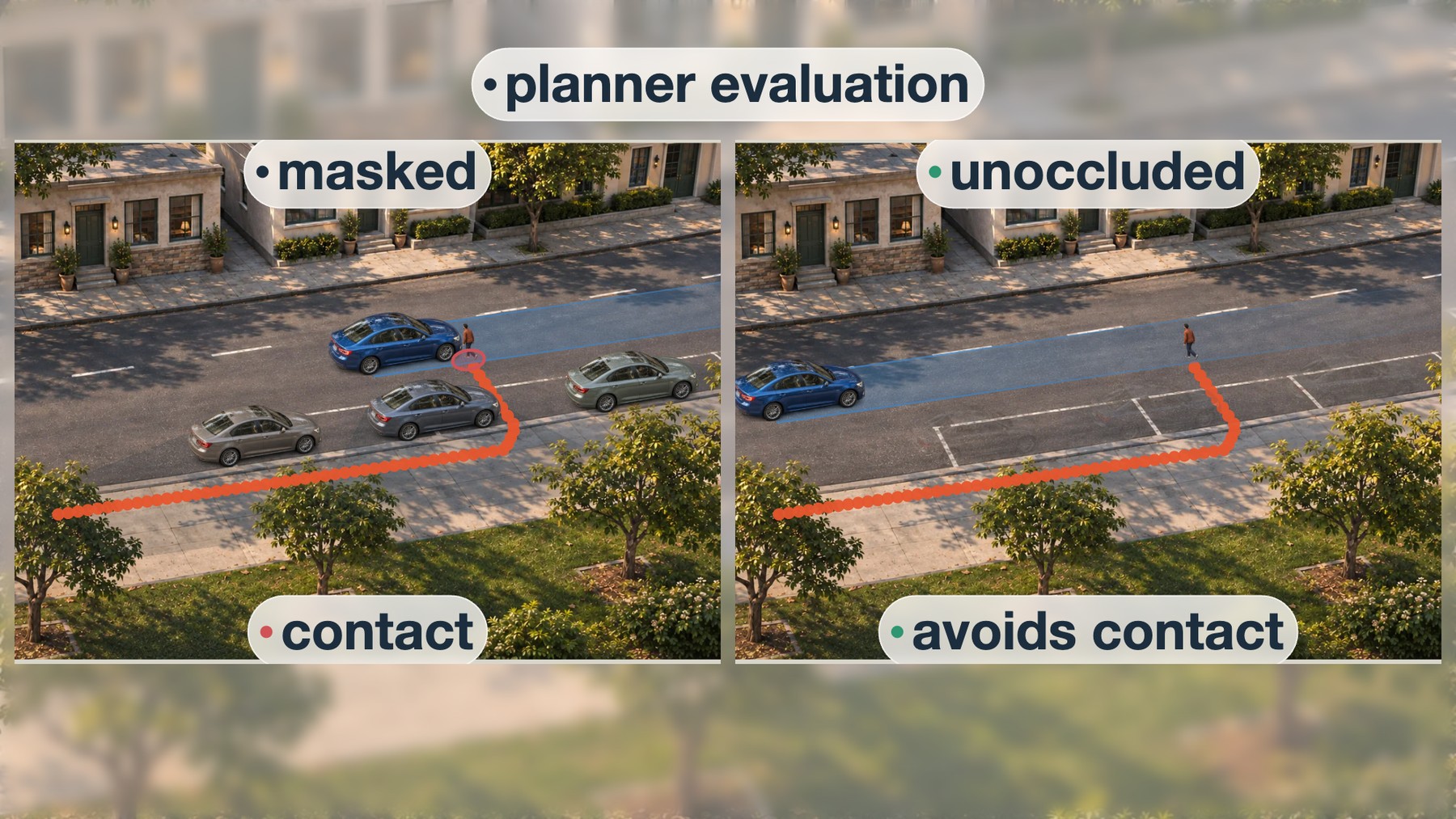} \\
{\footnotesize (a) Request C and E} & {\footnotesize (b) Generate and select} & {\footnotesize (c) Search safe responses} & {\footnotesize (d) Evaluate planners} \\[6pt]
\end{tabular}\\
{\footnotesize\textbf{Start construction for (b):} 19 candidates $\rightarrow$ 9 valid $\rightarrow$ top 5}\\[2pt]
{\footnotesize
\raisebox{-0.22em}{\includegraphics[height=0.75em]{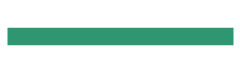}}\,kept, ranks 1--5\quad
\raisebox{-0.22em}{\includegraphics[height=0.75em]{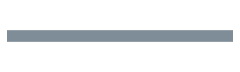}}\,valid, ranks 6--9\quad
\raisebox{-0.22em}{\includegraphics[height=0.75em]{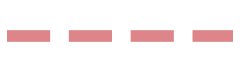}}\,removed: moves before replay start\quad
\raisebox{-0.22em}{\includegraphics[height=0.75em]{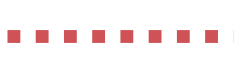}}\,removed: seen before exposure}\\[3pt]
\begin{tabular}{@{}c@{\hspace{4pt}}c@{\hspace{12pt}}c@{\hspace{4pt}}c@{}}
\includegraphics[height=1.6in]{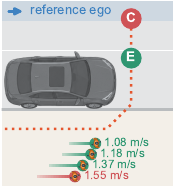} & \includegraphics[height=1.6in]{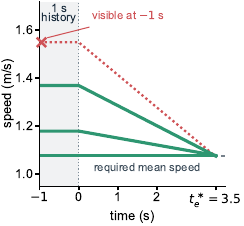} &
\includegraphics[height=1.6in]{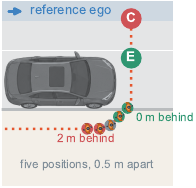} & \includegraphics[height=1.6in]{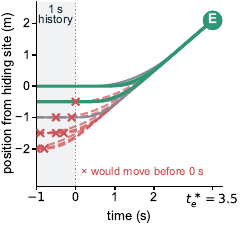} \\
{\footnotesize (e) 4 moving candidates} & {\footnotesize (f) Moving-start speeds} & {\footnotesize (g) 15 standing candidates} & {\footnotesize (h) Standing starts over time}
\end{tabular}
\caption{\textbf{How one test is built,} on an illustrative pedestrian scene. (a) The request places contact C on the reference ego\textquotesingle{}s path at $t_c^\ast=5.0$\,s and exposure E at $t_e^\ast=3.5$\,s, so $W^\ast=1.5$\,s. (b) From a kept start, E, and C, the trajectory generator proposes candidates (gray), checked in score order, and one is selected (orange). (c) The reference ego has no safe response from $t_e$ but has one from $t_v$, the first confirmed visibility with relevant occluders transparent to observation, so the test is Anticipatory. (d) A planner with masked observation contacts the target, whereas with unoccluded observation it avoids contact. (e, f) Moving starts, at three logged speed percentiles and at the required mean speed, keep their speed through a 1\,s history (drawn apart for legibility) and then change it uniformly to the required mean speed by $t_e^\ast$. (g, h) Standing starts wait 0 to 2\,m behind the hiding site and accelerate at one of three rates to pass E at $t_e^\ast$. Crosses mark removed starts.}\label{fig:test-construction}
\end{figure*}

\subsection*{A.1 Notation and Time Reference}\label{toc:A.1}
Table~\ref{table4} lists the symbols used throughout. A shard is one WOMD data file, and a driving episode groups the logs and tests from one recorded drive. A valid output passes every hard check of Appendix~A.2, and it is on time when both events fall within $\pm0.2$\,s of their requested times.

\begin{table}[!htbp]
\captionsetup{type=table}
\centering
\caption{\textbf{Notation.} Event times are measured from replay start. \emph{Target start} denotes a position, and \emph{movement start time} denotes when motion begins.}\label{table4}
{\fontsize{7.6}{8.8}\selectfont\setlength{\tabcolsep}{2pt}\renewcommand{\arraystretch}{1.00}
\begin{tabular}{@{}>{\raggedright\arraybackslash}p{63pt}>{\raggedright\arraybackslash}p{164pt}@{}}
\toprule
Symbol & Definition \\
\midrule
\multicolumn{2}{@{}l}{\textbf{Event times}}\\
\(t_e^\ast,t_c^\ast\) & Requested exposure \oncuetail{and reference-contact times.} \\
\(t_e,t_c\) & Realized confirmed exposure and \oncuetail{first reference contact.} \\
\(W^\ast,W\) & Requested and realized windows: \(t_c^\ast-t_e^\ast\) and \(t_c-t_e\). \\
\(t_v\) & First confirmed visibility with relevant occluders transparent to observation. \\
\midrule
\multicolumn{2}{@{}l}{\textbf{Trajectory generation}}\\
\(\mathcal R\) & Target route. \\
\(f_\theta\) & Trajectory generator of main Eq.~\eqref{eq:conditional-completion}, conditioned on scene, start history, route, and event conditions. \\
\(\mathbf p_e,\mathbf p_c,\mathcal C\) & Exposure and contact points (positions of the two timed waypoints) and the event conditions they form. \\
\midrule
\multicolumn{2}{@{}l}{\textbf{Response search}}\\
\(B,\mathcal A_B\) & Search budget and the first $B$ responses of the ordered library. \\
\(\mathcal S_B(t)\) & Candidates in the prefix that are safe under intervention at $t$. \\
\midrule
\multicolumn{2}{@{}l}{\textbf{Evaluation and experiments}}\\
\(W_{\mathrm{run}}\) & Window realized in a closed-loop run (Appendix F). \\
\(\rho,\alpha\) & Action scale of the slowed control (F.5.1) and desired-speed scale of IDM-conservative (C.3). \\
\bottomrule
\end{tabular}
}

\end{table}

\par

\subsection*{A.2 Starts, Visibility, and Trajectory Selection}\label{toc:A.2}

\noindent\textbf{Routes and admissible requests.} Pedestrians and cyclists emerge near roadside parking, median-side stopped vehicles, off-road parking edges, or queue gaps, crossing directly, following the road edge and turning, or, for cyclists, pulling out. Vehicles follow map lane chains that cross the ego route, and hiding vehicles stay within 0.5\,m of their initial position until 1\,s after the reference ego reaches the crossing. A request needs a hidden route segment and a required mean speed within class capability. Class limits follow published measurements: pedestrian speed and acceleration from sprint data~\cite{stuhec2023,stavridis2022}, cyclist braking from pitch-over thresholds~\cite{mordfin1975}, vehicle braking, lateral acceleration, and jerk from the NIEON reference driver~\cite{nieon}, vehicle acceleration and curvature from Waymax~\cite{waymax}, and tracking error from Hydra-MDP++~\cite{hydramdppp}. The same limits bound start transitions and generated motion, and vehicle requests whose required mean exceeds the highest pre-contact lane limit plus tolerance are dropped.
\par

\noindent\textbf{Visibility and static layout.} A frame is visible when at least one of five rays from a sensor at the logged ego\textquotesingle{}s height (2.3\,m) to the target center and corners is clear within 80\,m, about where WOMD~\cite{womd} tracks end. Exposure needs three such frames in a row. Besides road users, everything outside the open region (road surface, lanes, crosswalks, and recorded road-user positions, dilated by 4\,m) occludes, except sightlines from the recorded ego to tracked road users.
\par

\noindent\textbf{Motion statistics.} From 484,710 scenes in 1,000 WOMD training shards, common speeds span the 5th to 95th logged percentiles, acceleration, jerk, lateral acceleration, and yaw rate beyond their 99th percentiles are flagged as rare motion, and vehicles may exceed the lane speed limit by up to 0.655\,m/s in any frame, the 95th percentile of the largest excess of logged vehicles.
\par

\noindent\textbf{Moving and standing starts.} Moving starts use the 25th, 50th, and 75th percentiles of logged speed (for vehicles, of the speed-to-limit ratio) and, below the common-speed bound, a constant-speed start at the required mean. Before requested exposure, the hidden target changes speed uniformly to the required mean within class acceleration limits. Standing pedestrians and cyclists wait just behind the hiding site and vehicles at the stop line, with movement start time and cruising speed solved from logged accelerations to pass both waypoints. Ranking compares each start\textquotesingle{}s speed, start acceleration, and transition acceleration with logged medians.
\par

\noindent\textbf{Checks and selection.} Ranking keeps at most five starts. For each in rank order, the generator\textquotesingle{}s candidates are checked in descending model-score order. Hard checks cover finite states, vehicle continuity at motion start, approximate timing for pedestrians and cyclists, a first reference contact, motion coverage and class limits, no static or standing-road-user contact, a hidden duration of at least 0.5\,s, confirmed exposure before contact and within range with re-occlusion of at most 0.5\,s, and a realized window within 0.2\,s of $W^\ast$, from the first valid history frame through the test cutoff (Appendix B). Passing candidates carry quality flags for moving-background contact, near misses, rare motion, curbs, signals, stop signs, wrong-way travel, speeding, and inferred signal phases. Selection prefers fewer flags, candidate order breaks ties, and a flag-free candidate ends the search. A request without a passing candidate is rejected with its reason.
\par

\noindent\textbf{Grid-start control.} Grid starts replace constructed starts with positions every 5\,m upstream of contact, up to the distance covered at median training speed in 8\,s, or standing start times every 0.5\,s before requested exposure, at logged median speed and start acceleration. Up to five are retained uniformly, and the model still receives both timed waypoints.
\par

\subsection*{A.3 Trajectory Generator and Training}\label{toc:A.3}
\noindent\textbf{Condition sampling and losses.} Training samples 0 to 3 waypoints with probabilities 0.1, 0.2, 0.5, and 0.2, filled in the order contact, exposure, then a logged position before contact. Generation uses only the first two. Waypoints receive 0.3\,m position noise, and route and signal inputs are dropped with probabilities 0.5 and 0.2. The waypoint term of main Eq.~\eqref{eq:wp} divides its weighted L1 sum over the minibatch by the number of available slots, at least one, with weights 1, 3, and 1 for contact, exposure, and the third slot. The other terms average over batch samples and valid frames.
\par
\noindent\textbf{Architecture and training.} The encoder and decoder have six layers and eight heads (widths 256 and 512). Per class, 64 intentions produce 8\,s trajectories at 10\,Hz in the frame of the target at the prediction origin. We train with AdamW on eight GPUs (batch size 128 per process) for 30 epochs on the 1,000 WOMD training shards with MTR\textquotesingle{}s learning-rate schedule~\cite{mtr}, keeping the epoch with the lowest loss on held-out training scenes.
\par
\section*{Appendix B. Safe-Response Labeling}\label{toc:B}
This appendix describes the oracle's evaluation interval, reference ego, response library, and two properties of event timing and library expansion. The oracle labels a test Reactive if a safe response exists from $t_e$, Anticipatory if one exists only from $t_v$, and Unresolved if a complete search finds none. Each label is therefore relative to the reference ego\textquotesingle{}s state, the response library, and the time from which a response may start. Appendix~E varies each of them.

\subsection*{B.1 Evaluation Interval and Coverage}\label{toc:B.1}

\noindent\textbf{Visibility and shared cutoff.} If removing the responsible occlusion gives no earlier confirmed visibility, $t_v=t_e$. The test cutoff ends evaluation when the target first leaves the ego route band (ego half-width plus 0.4\,m per side) after reference contact, at run end if overlap persists, or when target data are missing. Generation, quality, response, and main collision checks share it, shifted with a triggered target start.
\par

\noindent\textbf{Coverage.} Requests require at least 1\,s of recording after reference contact, and main-protocol actions start at their designated time without delay. A test is judged only if a post-exposure action avoids at-fault background contact. Insufficient duration, pre-action contact, or incomplete core search yields Not judged. Both searches run the 288 core responses first and extend to all 1,316 only when no response from $t_e$ is \oncuetail{yet safe.}
\par

\subsection*{B.2 Reference Ego and Response Library}\label{toc:B.2}

\noindent\textbf{Reference ego.} The reference ego follows the logged path and map continuation, accounting for speed limit, curvature, signals, stop signs, and recorded leaders, and gives way to crossing and oncoming traffic with a 1.5\,s gap, with one configuration for all target classes. It responds to recorded traffic but not to the inserted target and brakes at up to 8.3\,m/s$^2$ with a jerk limit of 30\,m/s$^3$. Scenes with reference-background contact are excluded.
\par

\noindent\textbf{Action representation and execution.} Longitudinal plans set station along the reference route and lateral offsets depend on traveled distance, so stopped vehicles hold their position. Direct-steering candidates supply steering commands. nuPlan\textquotesingle{}s two-stage controller~\cite{nuplan}, an LQR tracker driving a kinematic bicycle, executes the motion.
\par

\par

\noindent\textbf{Response library.} Each response pairs a longitudinal plan with a lateral plan. Longitudinal plans accelerate, keep speed, brake at up to 8.3\,m/s$^2$ (emergency braking at a jerk of 30\,m/s$^3$), stop before the first target conflict, or brake and then recover. Lateral plans keep the lane, hold an offset, swerve and return, change one lane, or steer directly. Offsets, swerves, and lane changes start immediately or 0.5\,s later. The 288 core responses, all required for classification, cover these families coarsely. Refinements add finer braking levels, offset amplitudes, and durations, giving 1,316 distinct responses ordered from coarse to fine. Each budget uses a prefix of this list and reuses prior executions.
\par

\noindent\textbf{Execution and safety.} Execution keeps each response within the class limits of A.2, speed below 40\,m/s, and tracking error below 0.5\,m, with road and contact checks. Every target contact counts, whereas background contacts follow nuPlan\textquotesingle{}s fault rules~\cite{nuplan}. Outcomes are inapplicable, infeasible, background unsafe, target unsafe, or safe.
\par

\subsection*{B.3 Event Times and Response Coverage}\label{toc:B.3}

\noindent\textbf{Remark 1: a common temporal shift.} Let $e_e=t_e-t_e^*$ and $e_c=t_c-t_c^*$. We define the common shift $\bar e=(e_e+e_c)/2$ and the window error $\delta=e_c-e_e=W-W^*$. Then $\max(|e_e|,|e_c|)=|\bar e|+|\delta|/2$, because substituting $e_e=\bar e-\delta/2$ and $e_c=\bar e+\delta/2$ gives this identity. Window tolerance bounds $|\delta|$, while joint timing bounds $|\bar e|+|\delta|/2$.
\par

\noindent\textbf{Remark 2: library expansion retains safe actions.} Fix scene, observations, dynamics, evaluation interval, and deterministic safety checks. If $B\leq B'$ gives nested candidate prefixes $\mathcal A_B\subseteq\mathcal A_{B'}$, then $\mathcal S_B(t)\subseteq\mathcal S_{B'}(t)$. Hence, for a fixed evaluable test, a Reactive label with a safe post-exposure action persists as the budget grows, whereas new post-exposure safe actions can turn Anticipatory labels into Reactive ones.
\par

\section*{Appendix C. Evaluation Protocol and Metrics}\label{toc:C}
This appendix specifies what is compared and how: comparison scope and intervals, motion-quality metrics, and planner execution and outcomes. Table~\ref{tab:data-sets} lists the data sets.

\begin{table}[!t]
\centering
\caption{\textbf{Data sets.} Shards 0--149 are WOMD validation files.}\label{tab:data-sets}
{\fontsize{7.6}{8.8}\selectfont\setlength{\tabcolsep}{2pt}\renewcommand{\arraystretch}{1.0}
\begin{tabular*}{\columnwidth}{@{\extracolsep{\fill}}>{\raggedright\arraybackslash}p{0.30\columnwidth}>{\raggedright\arraybackslash}p{0.42\columnwidth}>{\raggedleft\arraybackslash}p{0.22\columnwidth}@{}}
\toprule
Set (used in) & Source and selection & Size\\
\midrule
Generator training (A) & 1,000 WOMD training shards & 484,710 scenes\\
Development (C.3, F.5) & shards 0--11 & 93 scenes, 515 tests\\
Evaluation logs (all) & shards 12--149, no driving episode shared with development & 33,648 logs\\
Vehicle requests (D) & with a constructed start, shards 12--35 & 6,054\\
Shared requests (D) & of these, with a valid OnCue (model) output & 2,624\\
Closed-loop tests (E, F) & $\le$10 per site, class, label, and window & 644 (484 Reactive)\\
Screening pool (E) & $\le$10 per site, class, and window & 515\\
Demonstrations (F.5) & OnCue tests on 16 WOMD training shards & 16 shards\\
Natural encounters (F.5.4) & 300 WOMD training shards, no fine-tuning episode & 148 (79 Reactive)\\
\bottomrule
\end{tabular*}}
\end{table}

\noindent\textbf{Terms and rates.} Requested windows and labels refer to the reference ego. In closed loop the tested planner drives its own ego, which can pass the contact point earlier, arrive later, or stop. A standing target starts by trigger when the tested ego approaches, whereas a fixed timetable keeps the reference schedule (F.2). Rates are sample rates over the stated tests unless population-weighted, which weights each stratum by its test count before the 10-test sampling quota (C.1). Missing events, Unresolved, and Not judged are request counts, and mean $\pm$ SD is the sample standard deviation over three training seeds.

\subsection*{C.1 Comparison Scope and Intervals}\label{toc:C.1}
\noindent\textbf{Shared starts and motion import.} External generators run their native candidate generation and selection from vehicle starts of valid OnCue outputs, with the same 1\,s constant-speed history for every method. The shared evaluation places targets only from replay start, so 187 OnCue outputs are not hidden long enough to confirm exposure, and OnCue\textquotesingle{}s window match is below 100\%. Rollouts that end early, including SEAL\textquotesingle{}s final skill-policy rollout, continue at terminal speed and heading, and the shared checks include this continuation. Candidate search uses 64 proposals per OnCue start, 32 for CAT~\cite{cat} and SAGE~\cite{sage}, five rounds of 32 for SEAL, and five policy-guided rounds for GOOSE.
\par

\par

\noindent\textbf{GOOSE adaptation.} We use the WOMD implementation and checkpoint released with SEAL~\cite{seal} for GOOSE~\cite{goose}, with the final rollout as output and the shared continuity checks. Of 2,624 requests, GOOSE returns no output for 63, including 53 targets its released code treats as static, SEAL for 62, and each other external method for 4. Main Table~\ref{table2} counts these requests as failures, and Table~\ref{tab:baseline-timing-decomposition}(a) as Not judged.
\par

\noindent\textbf{KING adaptation.} KING~\cite{king} optimizes target acceleration and steering against fixed reference replay (151 Adam iterations, 4\,Hz simulation, an 8\,s horizon, and actions initialized at zero because WOMD has no recorded controls) and returns the first collision-producing iteration or the last iterate.
\par

\par

\noindent\textbf{Driving-episode resampling and weighting.} For 10,000 bootstrap replicates with seed 0, we resample driving episodes with replacement, keeping all their logs and tests, and paired comparisons use identical draws. The 2.5th and 97.5th percentiles give the 95\% confidence intervals of the main paper. Closed-loop lists sample up to 10 tests per stratum of hiding-site type (A.2), target class, response label, and requested window, with population weights fixed in resampling. Tables report test and driving-episode counts separately.
\par
\subsection*{C.2 Motion-Quality Metrics}\label{toc:C.2}

\noindent\textbf{Motion metrics.} Speed, acceleration, and yaw rate use 0.1\,s differences, computed separately across gaps, with valid segments weighted by frame count. The logged reference, shared by all methods, contains same-class moving road users from the contributing WOMD scenes, excluding the recorded ego and always-stationary tracks, over valid frames after replay start. Each quantity gives a one-dimensional Wasserstein distance, and the kinematic penalty is SAGE\textquotesingle{}s trajectory-averaged one.
\par

\noindent\textbf{Behavior and challenge.} Behavior penalty follows SAGE\textquotesingle{}s sharp-turn and low-speed-turn terms. The adversarial reward of main Table~\ref{table2}, against masked PDM-Closed~\cite{pdm}, is $10(1-t/N)$ for first contact at step $t$, with a shared $N=80$ frames after replay start. Without contact, it is $\exp(-0.2d_{\min})$ for a minimum center distance $d_{\min}\leq20$\,m and zero otherwise. Reference contact and PDM collision measure target contact with the reference ego and with masked PDM-Closed.
\par

\noindent\textbf{Background and road edges.} Background contact is the fraction of returned target trajectories with a bounding-box overlap with a replayed background road user, other than the intended reference contact, through the test cutoff (over the whole generated trajectory for Table~\ref{table-vru-intervals}(b)). Road-edge crossing uses WOMD boundary road edges, not median edges. Main Table~\ref{table2} reports both as shares of flagged outputs, which SAGE\textquotesingle{}s penalties weight by 10 and 50.
\par

\subsection*{C.3 Planners, Target Execution, and Outcomes}\label{toc:C.3}

\noindent\textbf{Our IDM variants.} All three variants follow the Intelligent Driver Model~\cite{idm} with the reference limits in B.2 and the same controller, and resolve corridor conflicts from 3\,s constant-velocity predictions under shared traffic controls and curvature limits. IDM-conservative scales the desired speed by $\alpha=0.30$, calibrated on 93 development scenes so that its median 4\,s no-target distance matches the occlusion-aware variant\textquotesingle{}s. IDM-occlusion-aware adds virtual hazards near parked vehicles and crossing lanes with 5\,m/s$^2$ cautious braking. With unoccluded observations, it keeps only virtual vehicles beyond sensing range, and its target collision matches IDM-base in every \oncuetail{class.}
\par

\noindent\textbf{PDM-Closed adaptation.} The public tuplan\_garage port keeps proposal generation and scoring with WOMD maps and shared perception, with three changes: lateral-offset proposal paths ease in over 2\,s of travel, at least 10\,m, instead of starting fully offset. Proposals, which keep the upstream 3\,m/s$^2$ deceleration, are simulated with the common executor, which then executes the selected trajectory once. The emergency brake, still triggered by an at-fault collision expected within 2\,s, acts at all speeds instead of only at or below 5\,m/s, with its deceleration raised from 4.05 to the 8.3\,m/s$^2$ vehicle limit used by the IDM planners and the oracle. Without these changes, population-weighted masked collision rises from 29.8\% to 49.8\% on Reactive and from 75.8\% to 83.2\% on Anticipatory tests.
\par

\noindent\textbf{LatentDriver inputs.} The released LatentDriver~\cite{latentdriver} J3 checkpoint receives road-user states in an $80\times20$\,m ego-centered crop after observation filtering, with the route continued along the map. Native displacement execution or the logged route gives population-weighted masked Reactive collision of 72.6\% and 83.8\%, versus 71.7\% for the main configuration.
\par

\noindent\textbf{Target start time and validity.} Standing targets trigger on the evaluated ego\textquotesingle{}s predicted arrival at the original contact location, then replay generated motion with the original background timetable, whereas moving targets keep fixed timetables. Both observation modes share this rule, but each run triggers on its own ego progress, so observation can alter ego motion and thus the target timetable. Fixed-timetable pairs compare observations under identical target motion. Background contact or near misses newly introduced by the shift make the scene invalid, whereas those already flagged at generation leave it valid. The main comparison includes untriggered, invalid, and unfinished runs.
\par

\noindent\textbf{Collision interval.} Collision is measured through the test cutoff, and full-run contact is recorded separately. A run is untriggered if its standing target never starts, and late if, in a masked fixed-timetable run, the tested ego trails the reference ego by more than 0.2\,s at reference exposure without contact. The lag divides the route-position gap by reference speed, floored at 1\,m/s. Safe-response success counts triggered, not late runs on valid scenes that, observed through the earlier of the test cutoff and 1\,s after scheduled reference contact, have no target contact before the cutoff and no at-fault background contact, road violation, or vehicle violation during the run.
\par

\noindent\textbf{Road checks.} Outcomes beyond collision use the vehicle-capability checks and WOMD road boundaries, directions, and limits, with off-road, wrong-way, and reverse-heading tolerances at the 99.9th percentiles of recorded ego behavior.
\par

\FloatBarrier

\section*{Appendix D. Additional Generation Results}\label{toc:D}
This appendix extends the generation results of Q1 to Q3 along the pipeline: requests and starts, candidate timing, timing and motion quality against other generators, and safe responses.

\begin{table}[!t]
\centering
\caption{\textbf{Generation rates and raw-candidate timing.} (a) The share of requests that get a start and, for each OnCue variant, the share of these with a valid output and the share of valid outputs on time. (b) OnCue (model)\textquotesingle{}s own timing before checks and selection, on 1/10 of the requests with a start. Selected outputs may come from later starts and pass every check. A valid output passes every hard check of Appendix~A.2, and an output is on time if both events are within $\pm0.2$\,s.}\label{table7}\label{tab:component-ablation-full}\label{tab:raw-candidate-timing}
\begin{minipage}[t]{\columnwidth}
\vspace{0pt}
{\fontsize{7.6}{8.8}\selectfont\setlength{\tabcolsep}{2pt}\renewcommand{\arraystretch}{1.00}
\oncuetablealign
\begin{tabular*}{\columnwidth}{@{\extracolsep{\fill}}lrrrrrr@{}}
\toprule
\multicolumn{7}{@{}l}{\itshape (a) Generation rates, full evaluation set}\\
& & & \multicolumn{2}{c}{OnCue (rule)} & \multicolumn{2}{c}{OnCue (model)}\\
\cmidrule(lr){4-5}\cmidrule(l){6-7}
Target & Requests & \shortstack{Start /\\request (\%)} & \shortstack{Valid /\\start (\%)} & \shortstack{On time /\\valid (\%)} & \shortstack{Valid /\\start (\%)} & \shortstack{On time /\\valid (\%)}\\
\midrule
Pedestrian & \oncuenum{0}{130,140}{} & \oncuenum{1}{22}{6} & \oncuenum{1}{94}{2} & \oncuenum{1}{99}{9} & \oncuenum{1}{90}{5} & \underline{\oncuenum{1}{96}{6}}\\
Cyclist & \oncuenum{0}{81,468}{} & \oncuenum{1}{16}{5} & \oncuenum{1}{86}{2} & \oncuenum{1}{99}{7} & \oncuenum{1}{93}{7} & \oncuenum{1}{95}{5}\\
Vehicle & \oncuenum{0}{234,366}{} & \underline{\oncuenum{1}{13}{4}} & \oncuenum{1}{29}{5} & \oncuenum{1}{95}{9} & \oncuenum{1}{43}{3} & \underline{\oncuenum{1}{90}{1}}\\
\bottomrule
\end{tabular*}}
\end{minipage}\par\vspace{5pt}
{\fontsize{7.6}{8.8}\selectfont\setlength{\tabcolsep}{2pt}\renewcommand{\arraystretch}{1.00}
\oncuetablealign
\begin{tabular*}{\columnwidth}{@{\extracolsep{\fill}}lrrrrr@{}}
\toprule
\multicolumn{6}{@{}l}{\itshape (b) Raw-candidate timing: OnCue (model)\textquotesingle{}s 64 candidates at the first start}\\
& & \multicolumn{3}{c}{Raw candidates on time (\%)} & \multicolumn{1}{c}{Selected on time (\%)}\\
\cmidrule(lr){3-5}\cmidrule(l){6-6}
Target & \shortstack{Sampled\\requests} & \shortstack{Average\\candidate} & \shortstack{Top-scored\\candidate} & \shortstack{Any of\\the 64} & \shortstack{Of valid\\outputs}\\
\midrule
Pedestrian & \oncuenum{0}{2,900}{} & \oncuenum{1}{41}{5} & \underline{\oncuenum{1}{97}{1}} & \oncuenum{1}{98}{0} & \oncuenum{1}{96}{7}\\
Cyclist & \oncuenum{0}{1,325}{} & \oncuenum{1}{24}{6} & \underline{\oncuenum{1}{89}{7}} & \oncuenum{1}{96}{5} & \oncuenum{1}{95}{6}\\
Vehicle & \oncuenum{0}{3,090}{} & \oncuenum{1}{6}{1} & \underline{\oncuenum{1}{45}{1}} & \oncuenum{1}{49}{7} & \oncuenum{1}{89}{8}\\
\bottomrule
\end{tabular*}}
\end{table}
\subsection*{D.1 Where Requests Are Lost}\label{toc:D.1}
We locate where requests are lost and check whether recorded logs already contain the targeted conflicts. Original logs rarely contain qualifying conflicts, which satisfy the generation protocol's visibility and contact requirements. In replays of 33,648 logs with recorded background motion, the recorded ego encounters hidden road users able to reach its path in 18.3\% of logs, but none is a qualifying Reactive or Anticipatory conflict. With the reference ego instead, the share is 19.0\%, with three qualifying conflicts, all Reactive, whose natural windows are 4.0, 4.7, and \oncuetail{6.7\,s.}

Because a request needs a hidden start before trajectory generation can return an output, Table~\ref{table7} follows this process on the full evaluation set and compares the two OnCue variants on the same starts.

\par\noindent\textbf{Results.}
Few requests get a constructed start: 13.4\% to 22.6\% in each class. Once a start exists, OnCue (model) returns a valid output for over 90\% of pedestrian and cyclist requests but for only 43.3\% of vehicle requests (Table~\ref{table7}(a)), or 13,572 vehicle outputs.

\par\noindent\textbf{Finding.}
Most requests are thus lost before generation, because a start needs an occluder that hides the approach until the requested exposure and a speed profile that reaches both waypoints within class limits (A.2). For vehicles, generation also fails on most requests that have a start (D.2).

\begin{table}[!t]
\centering
\caption{\textbf{Conditional timing and rates over all requests.} (a) Window match given reference contact on the shared requests, and why the other requests lack the Reactive or Anticipatory label that coverage requires: events unmet (no reference contact, confirmed exposure, or long enough hidden interval), Unresolved (no safe response in the library), or not judged (no output, too little recording, contact before any action, or incomplete search, see B.1). (b) Main Table~\ref{table2} over all requests, counting missing outputs as failures. Coverage counts requests whose output is labeled Reactive or Anticipatory, and shades rank the top three per column. OnCue (rule) keeps a failing first candidate on 4,321 requests, so only 1,733 of its outputs (28.6\% of requests) pass every check, whereas all OnCue (model) outputs do. $W$ match requires reference contact and $|W-W^\ast|\leq0.2$\,s. Higher PDM-Closed collision means a harder test.}\label{tab:baseline-timing-decomposition}\label{fig6}\label{table10}\label{tab:all-vehicle-requests}
{\fontsize{7.6}{8.8}\selectfont\setlength{\tabcolsep}{2pt}\renewcommand{\arraystretch}{1.00}
\oncuetablealign
\begin{tabular*}{\columnwidth}{@{\extracolsep{\fill}}lrrrrr@{}}
\toprule
\multicolumn{6}{@{}l}{\itshape (a) Conditional timing: 2,624 requests shared by all generators, same starts}\\
& \multicolumn{2}{c}{Rates (\%)} & \multicolumn{3}{c}{Uncovered requests ($n$) $\downarrow$}\\
\cmidrule(lr){2-3}\cmidrule(l){4-6}
Method & \shortstack{Contact /\\request $\uparrow$} & \shortstack{$W$ match /\\contact $\uparrow$} & \shortstack{Events\\unmet} & \shortstack{Unre-\\solved} & \shortstack{Not\\judged}\\
\midrule
CAT~\cite{cat} & \oncuenum{1}{70}{0} & \oncuenum{1}{38}{2} & \oncuenum{0}{1,107}{} & \oncuenum{0}{3}{} & \oncuenum{0}{4}{}\\
KING~\cite{king} & \oncuenum{1}{69}{9} & \oncuenum{1}{60}{2} & \oncuenum{0}{1,114}{} & \oncuenum{0}{2}{} & \oncuenum{0}{4}{}\\
GOOSE~\cite{goose} & \oncuenum{1}{58}{0} & \oncuenum{1}{67}{0} & \oncuenum{0}{1,322}{} & \oncuenum{0}{1}{} & \oncuenum{0}{63}{}\\
SEAL~\cite{seal} & \oncuenum{1}{59}{9} & \oncuenum{1}{37}{1} & \oncuenum{0}{1,251}{} & \oncuenum{0}{1}{} & \oncuenum{0}{62}{}\\
SAGE~\cite{sage} ($w=0.5$) & \oncuenum{1}{32}{1} & \oncuenum{1}{36}{2} & \oncuenum{0}{2,015}{} & \oncuenum{0}{3}{} & \oncuenum{0}{5}{}\\
SAGE~\cite{sage} ($w=1.0$) & \oncuenum{1}{21}{3} & \oncuenum{1}{26}{4} & \oncuenum{0}{2,255}{} & \oncuenum{0}{4}{} & \oncuenum{0}{4}{}\\
\midrule
OnCue (rule) & \oncuenum{1}{100}{0} & \oncuenum{1}{81}{7} & \oncuenum{0}{349}{} & \oncuenum{0}{1}{} & \oncuenum{0}{0}{}\\
OnCue (model) & \oncuenum{1}{100}{0} & \cellcolor{OnCueGain}\oncuenum{1}{92}{9} & \oncuenum{0}{187}{} & \oncuenum{0}{3}{} & \oncuenum{0}{3}{}\\
\bottomrule
\end{tabular*}}\par\vspace{4pt}
{\fontsize{7.6}{8.8}\selectfont\setlength{\tabcolsep}{2pt}\renewcommand{\arraystretch}{1.00}
\oncuetablealign
\begin{tabular*}{\columnwidth}{@{\extracolsep{\fill}}lrrrr@{}}
\toprule
\multicolumn{5}{@{}l}{\itshape (b) Rates over all requests: 6,054 vehicle requests with a start, shards 12--35}\\
& \multicolumn{4}{c}{Rates (\%)}\\
\cmidrule(l){2-5}
Method & \shortstack{Contact /\\request $\uparrow$} & \shortstack{$W$ match /\\request $\uparrow$} & \shortstack{Coverage /\\request $\uparrow$} & \shortstack{PDM-Closed collision\\/ request $\uparrow$}\\
\midrule
CAT & \oncuerank{Three}{\oncuenum{1}{72}{0}} & \oncuenum{1}{15}{8} & \oncuenum{1}{38}{2} & \oncuerank{Two}{\oncuenum{1}{21}{1}}\\
KING & \oncuerank{Two}{\oncuenum{1}{76}{2}} & \oncuerank{Three}{\underline{\oncuenum{1}{27}{3}}} & \oncuerank{Three}{\oncuenum{1}{40}{0}} & \oncuerank{Three}{\oncuenum{1}{18}{1}}\\
GOOSE & \oncuenum{1}{57}{4} & \oncuenum{1}{24}{3} & \oncuenum{1}{31}{6} & \oncuenum{1}{13}{6}\\
SEAL & \oncuenum{1}{59}{5} & \oncuenum{1}{13}{1} & \oncuenum{1}{33}{0} & \oncuenum{1}{16}{7}\\
SAGE ($w=0.5$) & \oncuenum{1}{29}{1} & \oncuenum{1}{7}{0} & \oncuenum{1}{15}{0} & \oncuenum{1}{9}{1}\\
SAGE ($w=1.0$) & \oncuenum{1}{23}{1} & \oncuenum{1}{3}{9} & \oncuenum{1}{10}{3} & \oncuenum{1}{9}{0}\\
\midrule
OnCue (rule) & \oncuerank{One}{\oncuenum{1}{100}{0}} & \oncuerank{One}{\oncuenum{1}{50}{7}} & \oncuerank{One}{\oncuenum{1}{56}{7}} & \oncuerank{One}{\oncuenum{1}{24}{1}}\\
OnCue (model) & \oncuenum{1}{43}{3} & \oncuerank{Two}{\underline{\oncuenum{1}{40}{2}}} & \oncuerank{Two}{\oncuenum{1}{40}{2}} & \oncuenum{1}{14}{8}\\
\bottomrule
\end{tabular*}}
\end{table}
\begin{table}[!t]
\centering
\caption{\textbf{Motion quality for vehicles and for pedestrians and cyclists.} (a) Behavior penalties for sharp and low-speed turns up to the test cutoff, and background contact before reference contact (before the cutoff if contact is absent). Main Table~\ref{table2} gives it over the full interval. (b) Wasserstein distances to class-matched moving road users in the same WOMD scenes, and background contact, both over the whole generated trajectory. Main Table~\ref{table2} and (a) compare the two variants on vehicles. Lower is better in every column.}\label{table8}\label{tab:paired-motion-distances}\label{table-vru-intervals}
{\fontsize{7.6}{8.8}\selectfont\setlength{\tabcolsep}{2pt}\renewcommand{\arraystretch}{1.00}
\oncuetablealign
\begin{tabular*}{\columnwidth}{@{\extracolsep{\fill}}lrrr@{}}
\toprule
\multicolumn{4}{@{}l}{\itshape (a) Vehicles: every generator, 2,624 shared requests}\\
& \multicolumn{2}{c}{Behavior penalty} & \\
\cmidrule(lr){2-3}
Method & Mean & Median & \shortstack{Background\\contact (\%)}\\
\midrule
CAT~\cite{cat} & \oncuenum{3}{1}{361} & \oncuenum{3}{0}{513} & \oncuenum{1}{42}{6}\\
KING~\cite{king} & \cellcolor{OnCueGain}\oncuenum{3}{0}{370} & \cellcolor{OnCueGain}\oncuenum{3}{0}{211} & \oncuenum{1}{37}{7}\\
GOOSE~\cite{goose} & \oncuenum{3}{4}{934} & \oncuenum{3}{0}{225} & \oncuenum{1}{38}{5}\\
SEAL~\cite{seal} & \oncuenum{3}{1}{763} & \oncuenum{3}{0}{539} & \oncuenum{1}{39}{9}\\
SAGE~\cite{sage} ($w=0.5$) & \oncuenum{3}{1}{163} & \oncuenum{3}{0}{510} & \oncuenum{1}{38}{3}\\
SAGE~\cite{sage} ($w=1.0$) & \oncuenum{3}{2}{391} & \oncuenum{3}{1}{053} & \cellcolor{OnCueGain}\underline{\oncuenum{1}{36}{0}}\\
\midrule
OnCue (rule) & \oncuenum{3}{0}{518} & \oncuenum{3}{0}{326} & \oncuenum{1}{37}{9}\\
OnCue (model) & \oncuenum{3}{2}{870} & \oncuenum{3}{0}{319} & \underline{\oncuenum{1}{36}{9}}\\
\bottomrule
\end{tabular*}}
\par\vspace{5pt}
{\fontsize{7.6}{8.8}\selectfont\setlength{\tabcolsep}{2pt}\renewcommand{\arraystretch}{1.00}
\oncuetablealign
\begin{tabular*}{\columnwidth}{@{\extracolsep{\fill}}llrrrr@{}}
\toprule
\multicolumn{6}{@{}l}{\itshape (b) Pedestrians and cyclists: the two OnCue variants, shards 12--35}\\
& & \multicolumn{3}{c}{Wasserstein distance} & \\
\cmidrule(lr){3-5}
Target & Method & \shortstack{Speed\\(m/s)} & \shortstack{Acceleration\\(m/s$^2$)} & \shortstack{Yaw rate\\(rad/s)} & \shortstack{Background\\contact (\%)}\\
\midrule
\multirow{2}{*}{Pedestrian} & OnCue (rule) & \underline{\oncuenum{3}{0}{648}} & \oncuenum{3}{2}{069} & \oncuenum{3}{0}{985} & \cellcolor{OnCueGain}\oncuenum{1}{13}{9}\\
 & OnCue (model) & \cellcolor{OnCueGain}\underline{\oncuenum{3}{0}{305}} & \cellcolor{OnCueGain}\oncuenum{3}{1}{892} & \cellcolor{OnCueGain}\oncuenum{3}{0}{914} & \oncuenum{1}{14}{9}\\
\midrule
\multirow{2}{*}{Cyclist} & OnCue (rule) & \oncuenum{3}{2}{174} & \oncuenum{3}{2}{396} & \oncuenum{3}{0}{265} & \cellcolor{OnCueGain}\oncuenum{1}{19}{9}\\
 & OnCue (model) & \cellcolor{OnCueGain}\oncuenum{3}{2}{145} & \cellcolor{OnCueGain}\oncuenum{3}{2}{190} & \cellcolor{OnCueGain}\oncuenum{3}{0}{230} & \oncuenum{1}{21}{1}\\
\bottomrule
\end{tabular*}}
\end{table}
\subsection*{D.2 Timing before Checks and Selection}\label{toc:D.2}
To separate the learned generator\textquotesingle{}s own timing from that of checks and selection, we measure all 64 OnCue (model) candidates from the first start on 1/10 of the requests, sampled by identifier independently of outcomes. Exposure and contact follow the hard-check definitions even when other checks fail. Table~\ref{tab:raw-candidate-timing}(b) reports within-request mean on-time share, top-score success, and any-candidate success.

\par\noindent\textbf{Results.}
Individual candidates are on time far less often than the top-scored candidate reported in the main text, and for vehicles even the best of all 64 candidates is on time in only 49.7\% of requests. The selected-output share instead allows later starts, requires all checks, and divides by valid outputs rather than by every diagnostic request.

\par\noindent\textbf{Finding.}
Scores favor on-time candidates in every class, but about half of vehicle requests have no on-time candidate among the 64 at the first start, so start construction is important.

\noindent\textbf{Q3 ablation.} The pedestrian 96.8\% and 66.3\% in the waypoint ablation (main Table~\ref{tab:timing-ablation}(c), shards 12--35) are likewise on-time shares of valid outputs. Over requests with a start, removing both waypoints lowers on-time generation from 87.4\% to 53.4\%.
\par

\subsection*{D.3 Timing across Generators}\label{toc:D.3}
Table~\ref{tab:baseline-timing-decomposition} separates missing reference encounters from timing error on the 2,624 shared requests (a) and repeats the comparison on all 6,054 vehicle requests with a start, using the first start beyond the shared requests (b).

\par\noindent\textbf{Results.}
Once reference contact occurs, CAT, SEAL, and SAGE ($w=0.5$) have similar window match on their own outputs despite different contact rates. Over all requests, OnCue (model) returns no test beyond the shared 2,624, which lowers its coverage and collision, whereas OnCue (rule) scores higher because it keeps its first candidate even when none passes the hard checks: only 9 of the 3,430 added requests have a rule output passing every check.

\par\noindent\textbf{Finding.}
Window match therefore depends on both constructing an encounter and scheduling its events.

\noindent\textbf{Collisions before and after optimization.} Of the 2,235 collisions that KING reports in 3,221 executed tests, 76.3\% are present at initialization (52.9\% of tests), because all methods receive the same OnCue starts, whose positions, headings, and speeds already support reference contact. Optimization adds 530 (16.5\% of tests), and 30.6\% of tests have no native collision. These counts precede exclusion of development-overlapping driving episodes and use KING\textquotesingle{}s native checks.
\par

\subsection*{D.4 Motion Quality across Generators}\label{toc:D.4}
Table~\ref{table8} compares motion quality over matched intervals for every generator on vehicles (a) and for the two OnCue variants on pedestrians and cyclists (b).

\par\noindent\textbf{Results.}
OnCue (model)\textquotesingle{}s high mean behavior penalty comes mostly from low-speed turning, which amplifies heading changes at small displacement (68 requests requiring mean speed below 0.5\,m/s contribute 1.160 of its 2.870), and from extension to the cutoff on the other requests (1.270). Its median is below those of CAT, SEAL, and both SAGE settings.

On pedestrians and cyclists, the learned model has the lower Wasserstein distance (WD) in all six cells (178--179 pedestrian and 145--146 cyclist logs), and paired 95\% intervals lie below zero for five of these six differences (learned minus rule-based), all but cyclist speed ($-0.028$\,m/s, $[-0.075, 0.024]$). The rule-based variant has the lower background contact, but both paired intervals span zero (pedestrian $+1.0$ points, $[-1.8, 3.8]$, and cyclist $+1.2$ points, $[-2.4, 5.0]$).

On vehicles, the rule-based variant has the lower mean penalty and the lower speed and acceleration WD (main Table~\ref{table2}), whereas the learned model has the slightly lower median penalty, the lower yaw-rate WD, and the lower background contact before reference contact. The rule-based variant is also on time for more of its valid outputs in every class (Table~\ref{table7}(a)).

\par\noindent\textbf{Finding.}
Motion-quality conclusions therefore depend on the metric and the evaluation period, because background contact rises after reference contact, when recorded traffic continues without responding to the inserted target. Between the OnCue variants, learned motion, trained on logged futures, is closer to logged pedestrian and cyclist motion than preset speed plans are, whereas on vehicles neither variant is better on every measure.

\subsection*{D.5 Missing Events versus Failed Action Search}\label{toc:D.5}
To see whether uncovered requests reflect missing events or failed action search, Table~\ref{table10}(a) separates outputs with a Reactive or Anticipatory safe response from the two failure kinds.

\par\noindent\textbf{Results.}
Most uncovered requests lack contact, exposure, or the required hidden interval: OnCue (model) misses these events on 187 of the 2,624 shared requests and leaves 3 Unresolved, whereas external generators miss them on 1,107 to 2,255.

\par\noindent\textbf{Finding.}
Requests are therefore lost mostly to missing events rather than to failed action search. External generators miss these events more often because none takes the requested event times, so their native search need not produce reference contact, confirmed exposure, or the required hidden interval.
\section*{Appendix E. Response-Label Sensitivity}\label{toc:E}
This appendix tests how much the response labels of Q4 depend on oracle settings and on the reference ego.
Among the 515 OnCue candidates screened in Q4, 463 are Reactive, 51 Anticipatory, and one Unresolved. Keeping only tests where masked PDM-Closed collides but unoccluded PDM-Closed avoids the target leaves 56 tests: 35 Reactive, 20 Anticipatory, and one Unresolved. Thus 62.5\% of retained failures occur in encounters where the reference ego keeps a safe post-exposure action.

\begin{table}[!t]
\centering
\caption{\textbf{Response-label sensitivity to oracle settings.} (a) Transitions not shown are zero. Of the tests, 61, 53, and 31 are Anticipatory, 15 vehicle tests are Unresolved, and the rest are Reactive. (b) Pooled over the 101,323 labeled outputs of both OnCue variants. The full library has 1,316 responses (B.2).}\label{table12}
{\fontsize{7.6}{8.8}\selectfont\setlength{\tabcolsep}{2pt}\renewcommand{\arraystretch}{1.00}\oncuetablealign
\begin{tabular*}{\columnwidth}{@{\extracolsep{\fill}}lrrrr@{}}
\toprule
\multicolumn{5}{@{}l}{\itshape (a) Braking jerk 30 to 13.9\,m/s$^3$: labels of the 644 tests}\\
& & \multicolumn{3}{c}{Labels ($n$)}\\
\cmidrule(l){3-5}
Target & Tests & Unchanged & \shortstack{Reactive to\\Anticipatory} & \shortstack{Anticipatory to\\Unresolved}\\
\midrule
Pedestrian & \oncuenum{0}{277}{} & \oncuenum{0}{269}{} & \oncuenum{0}{2}{} & \oncuenum{0}{6}{}\\
Cyclist & \oncuenum{0}{266}{} & \oncuenum{0}{264}{} & \oncuenum{0}{2}{} & \oncuenum{0}{0}{}\\
Vehicle & \oncuenum{0}{101}{} & \oncuenum{0}{93}{} & \oncuenum{0}{8}{} & \oncuenum{0}{0}{}\\
\midrule
All & \oncuenum{0}{644}{} & \oncuenum{0}{626}{} & \oncuenum{0}{12}{} & \oncuenum{0}{6}{}\\
\bottomrule
\end{tabular*}}
\par\vspace{4pt}
\begin{minipage}[t]{\linewidth}
\vspace{0pt}
{\fontsize{7.6}{8.8}\selectfont\setlength{\tabcolsep}{2pt}\renewcommand{\arraystretch}{1.00}\oncuetablealign
\begin{tabular*}{\textwidth}{@{\extracolsep{\fill}}lrr@{}}
\toprule
\multicolumn{3}{@{}l}{\itshape (b) Action library: Anticipatory labels kept}\\
& \multicolumn{2}{c}{Kept (\%), 288 actions grown to}\\
\cmidrule(l){2-3}
Target & 512 & 1,316\\
\midrule
Pedestrian & \oncuenum{1}{94}{2} & \oncuenum{1}{91}{6}\\
Cyclist & \oncuenum{1}{86}{8} & \oncuenum{1}{82}{0}\\
Vehicle & \oncuenum{1}{99}{3} & \oncuenum{1}{97}{6}\\
\midrule
All & \oncuenum{1}{94}{6} & \oncuenum{1}{91}{9}\\
\bottomrule
\end{tabular*}}
\end{minipage}
\par\vspace{4pt}
\begin{minipage}[t]{\linewidth}
\vspace{0pt}
{\fontsize{7.6}{8.8}\selectfont\setlength{\tabcolsep}{2pt}\renewcommand{\arraystretch}{1.00}\oncuetablealign
\begin{tabular*}{\textwidth}{@{\extracolsep{\fill}}lrrr@{}}
\toprule
\multicolumn{4}{@{}l}{\itshape (c) Response delay: Reactive tests still safe}\\
& & \multicolumn{2}{c}{Still safe (\%), actions start after exposure}\\
\cmidrule(l){3-4}
Target & \shortstack{Reactive\\tests} & 0.3\,s & 0.5\,s\\
\midrule
Pedestrian & \oncuenum{0}{216}{} & \oncuenum{1}{88}{0} & \oncuenum{1}{81}{9}\\
Cyclist & \oncuenum{0}{213}{} & \oncuenum{1}{85}{4} & \oncuenum{1}{77}{5}\\
Vehicle & \oncuenum{0}{55}{} & \oncuenum{1}{72}{7} & \oncuenum{1}{65}{5}\\
\midrule
All & \oncuenum{0}{484}{} & \underline{\oncuenum{1}{85}{1}} & \underline{\oncuenum{1}{78}{1}}\\
\bottomrule
\end{tabular*}}
\end{minipage}
\end{table}

\begin{table}[!t]
\centering
\caption{\textbf{Response-label sensitivity to the reference ego.} The evaluation set is regenerated with another reference ego: 0.8$\times$ speed (0.8$\times$ in (b) and (c)) scales its desired speed by 0.8, and 2.0\,s headway (2.0\,s in (b) and (c)) raises its time headway from 1.2\,s. Main is the main run\textquotesingle{}s reference ego. (c) Target classes are pooled without population weighting. Blue marks the two lowest rates in each column.}\label{tab:reference-ego-sensitivity}
{\fontsize{7.6}{8.8}\selectfont\setlength{\tabcolsep}{2pt}\renewcommand{\arraystretch}{1.00}\oncuetablealign
\begin{tabular*}{\columnwidth}{@{\extracolsep{\fill}}lrrr@{}}
\toprule
\multicolumn{4}{@{}l}{\itshape (a) Generation rates}\\
Rate (\%) $\uparrow$ & \shortstack{Main\\ego} & \shortstack{0.8$\times$\\speed} & \shortstack{2.0\,s\\headway}\\
\midrule
\multicolumn{4}{@{}l}{\textbf{Pedestrian}}\\
\leftskip=6pt\relax Valid / start & \oncuenum{1}{90}{5} & \oncuenum{1}{90}{9} & \oncuenum{1}{90}{6}\\
\leftskip=6pt\relax On time / valid & \oncuenum{1}{96}{6} & \oncuenum{1}{96}{4} & \oncuenum{1}{96}{2}\\
\multicolumn{4}{@{}l}{\textbf{Cyclist}}\\
\leftskip=6pt\relax Valid / start & \oncuenum{1}{93}{7} & \oncuenum{1}{94}{5} & \oncuenum{1}{94}{2}\\
\leftskip=6pt\relax On time / valid & \oncuenum{1}{95}{5} & \oncuenum{1}{94}{6} & \oncuenum{1}{94}{9}\\
\multicolumn{4}{@{}l}{\textbf{Vehicle}}\\
\leftskip=6pt\relax Valid / start & \oncuenum{1}{43}{3} & \oncuenum{1}{39}{0} & \oncuenum{1}{42}{6}\\
\leftskip=6pt\relax On time / valid & \oncuenum{1}{90}{1} & \oncuenum{1}{88}{8} & \oncuenum{1}{91}{1}\\
\bottomrule
\end{tabular*}
\par\vspace{2pt}
\begin{tabular*}{\columnwidth}{@{\extracolsep{\fill}}lrrrrr@{}}
\toprule
\multicolumn{6}{@{}l}{\itshape (b) Labels of requests labeled in both runs}\\
\shortstack{Reference\\ego} & \shortstack{Requests\\($n$)} & \shortstack{Unchanged\\(\%)} & \shortstack{Reactive to\\Anticipatory\\($n$)} & \shortstack{Anticipatory\\in main run\\($n$)} & \shortstack{Anticipatory\\kept\\(\%)}\\
\midrule
\multicolumn{6}{@{}l}{\textbf{Pedestrian}}\\
0.8$\times$ & \oncuenum{0}{20,728}{} & \oncuenum{1}{99}{5} & \oncuenum{0}{27}{} & \oncuenum{0}{149}{} & \oncuenum{1}{53}{7}\\
2.0\,s & \oncuenum{0}{23,312}{} & \oncuenum{1}{99}{7} & \oncuenum{0}{21}{} & \oncuenum{0}{162}{} & \oncuenum{1}{72}{2}\\
\multicolumn{6}{@{}l}{\textbf{Cyclist}}\\
0.8$\times$ & \oncuenum{0}{9,644}{} & \oncuenum{1}{99}{4} & \oncuenum{0}{10}{} & \oncuenum{0}{83}{} & \oncuenum{1}{47}{0}\\
2.0\,s & \oncuenum{0}{10,860}{} & \oncuenum{1}{99}{7} & \oncuenum{0}{11}{} & \oncuenum{0}{83}{} & \oncuenum{1}{71}{1}\\
\multicolumn{6}{@{}l}{\textbf{Vehicle}}\\
0.8$\times$ & \oncuenum{0}{8,020}{} & \oncuenum{1}{98}{9} & \oncuenum{0}{35}{} & \oncuenum{0}{229}{} & \oncuenum{1}{79}{5}\\
2.0\,s & \oncuenum{0}{9,608}{} & \oncuenum{1}{99}{0} & \oncuenum{0}{33}{} & \oncuenum{0}{313}{} & \oncuenum{1}{83}{1}\\
\bottomrule
\end{tabular*}
\par\vspace{2pt}
\begin{tabular*}{\columnwidth}{@{\extracolsep{\fill}}lrrrrrr@{}}
\toprule
\multicolumn{7}{@{}l}{\itshape (c) Masked target collision on each run\textquotesingle{}s own tests}\\
& \multicolumn{3}{c}{Reactive tests (\%) $\downarrow$} & \multicolumn{3}{c}{Anticipatory tests (\%) $\downarrow$}\\
\cmidrule(lr){2-4}\cmidrule(l){5-7}
Planner & Main & 0.8$\times$ & 2.0\,s & Main & 0.8$\times$ & 2.0\,s\\
\midrule
IDM-base & \oncuenum{1}{23}{3} & \oncuenum{1}{31}{6} & \oncuenum{1}{34}{2} & \oncuenum{1}{100}{0} & \oncuenum{1}{93}{8} & \oncuenum{1}{96}{7}\\
IDM-conservative & \cellcolor{OnCueGain}\oncuenum{1}{4}{3} & \cellcolor{OnCueGain}\oncuenum{1}{4}{4} & \cellcolor{OnCueGain}\oncuenum{1}{5}{8} & \cellcolor{OnCueGain}\oncuenum{1}{17}{2} & \cellcolor{OnCueGain}\oncuenum{1}{24}{2} & \cellcolor{OnCueGain}\oncuenum{1}{23}{6}\\
IDM-occlusion-aware & \cellcolor{OnCueGain}\oncuenum{1}{6}{0} & \cellcolor{OnCueGain}\oncuenum{1}{8}{2} & \cellcolor{OnCueGain}\oncuenum{1}{9}{5} & \cellcolor{OnCueGain}\oncuenum{1}{33}{1} & \cellcolor{OnCueGain}\oncuenum{1}{28}{9} & \cellcolor{OnCueGain}\oncuenum{1}{27}{6}\\
PDM-Closed & \oncuenum{1}{35}{5} & \oncuenum{1}{35}{8} & \oncuenum{1}{37}{3} & \oncuenum{1}{82}{8} & \oncuenum{1}{79}{7} & \oncuenum{1}{87}{0}\\
LatentDriver & \oncuenum{1}{73}{3} & \oncuenum{1}{68}{2} & \oncuenum{1}{67}{5} & \oncuenum{1}{82}{8} & \oncuenum{1}{71}{9} & \oncuenum{1}{82}{1}\\
\bottomrule
\end{tabular*}}
\end{table}
\subsection*{E.1 Varying the Oracle Settings}\label{toc:E.1}
To test whether labels depend on the oracle\textquotesingle{}s settings, we keep generated motion and the evaluation cutoff and change one setting at a time. In the first, emergency-braking jerk falls from 30 to 13.9\,m/s$^3$, derived from the performance model of UN Regulation No.~157~\cite{unr157}, which reaches 0.85\,g within 0.6\,s after full wrap of the ego and cut-in vehicles. In the second, nested action prefixes grow from 288 to 512 and 1,316. In the third, action starts are delayed by 0.3 or 0.5\,s.

\par\noindent\textbf{Results.}
Smoother braking changes few labels: 626 of 644 are unchanged, and every change goes from Reactive to Anticipatory or from Anticipatory to Unresolved (Table~\ref{table12}(a)). A larger library uncovers additional post-exposure avoidance and changes cyclist labels most, and it is not saturated, because the share of Anticipatory labels kept still falls from 94.6\% at 512 actions to 91.9\% at 1,316 (Table~\ref{table12}(b)). Of the 484 tests labeled Reactive without delay, 412 (85.1\%) retain safe avoidance after 0.3\,s delay and 378 (78.1\%) after 0.5\,s, and vehicle retention is lowest at both delays (Table~\ref{table12}(c)).

\par\noindent\textbf{Finding.}
Action capability and response delay thus alter the available avoidance, but within the tested settings most labels hold. A larger nested library retains known safe actions and can discover additional ones.

\subsection*{E.2 Varying the Reference Ego}\label{toc:E.2}
Because event times and labels are relative to reference-ego motion (B.2), we regenerate the evaluation set with reference desired speed reduced to 0.8 of the limit or with the car-following time headway increased from 1.2 to 2.0\,s. Generation, checks, the response library, and tested planners remain fixed.

\par\noindent\textbf{Results.}
Changing the reference ego affects labels, generation, and the planner comparison differently (Table~\ref{tab:reference-ego-sensitivity}). For labels of requests labeled in both runs, fewer than 0.5\% of Reactive labels change, whereas the slower reference ego keeps only 47.0\% to 79.5\% of Anticipatory labels, turning nearly all others Reactive, and longer headway keeps 71.1\% to 83.1\%. The overall agreement of 98.9\% to 99.7\% mainly reflects how few labels are Anticipatory. For generation, pedestrian and cyclist rates stay within one point of the main run, vehicle success falls most with the slower reference ego (43.3\% to 39.0\%), and both modified reference egos construct starts for 4.6\% to 21.9\% fewer requests. For planners, the cautious IDM variants keep the lowest collision rates and LatentDriver the highest Reactive collision on each reference ego\textquotesingle{}s own unweighted masked cohort, although absolute rates change by up to 10.9 points.

\par\noindent\textbf{Finding.}
Reactive labels and which planners collide least therefore barely depend on the tested reference egos, whereas Anticipatory labels depend on how the reference ego approaches, because a slower approach often leaves it a safe post-exposure action.
\section*{Appendix F. Planner Behavior, Training, and Transfer}\label{toc:F}
This appendix extends the planner results of Q5 (F.1 to F.3), checks them without target-background contact (F.4), and extends the training results of Q6 (F.5).

\begin{table}[!t]
\centering
\caption{\textbf{Planner outcomes by response label and scene validity.} (a) The rates main Table~\ref{table3} does not report: unoccluded safe-response success on Anticipatory tests, and collision on Reactive tests, which counts target contact at any ego speed. (b) Keeps the tests valid for every recorded planner and observation mode. The counts show how the denominator changes. Unocc.\ denotes unoccluded observation, and success is safe-response success.}\label{table13}\label{table15}\label{fig13}\label{table14}
{\fontsize{7.6}{8.8}\selectfont\setlength{\tabcolsep}{2pt}\renewcommand{\arraystretch}{1.00}
\oncuetablealign
\begin{tabular*}{\columnwidth}{@{\extracolsep{\fill}}lrrr@{}}
\toprule
\multicolumn{4}{@{}l}{\itshape (a) Response label: population-weighted rates, all attempted tests}\\
& \shortstack{Anticipatory\\success (\%) $\uparrow$} & \multicolumn{2}{c}{Reactive collision (\%) $\downarrow$}\\
\cmidrule(lr){2-2}\cmidrule(l){3-4}
Planner & Unocc. & Masked & Unocc.\\
\midrule
\multicolumn{4}{@{}l}{\textbf{Pedestrian}: 61 Anticipatory tests, 216 Reactive tests}\\
IDM-base & \oncuenum{1}{78}{6} & \oncuenum{1}{6}{5} & \oncuenum{1}{2}{0}\\
IDM-conservative & \oncuenum{1}{57}{3} & \oncuenum{1}{0}{2} & \oncuenum{1}{0}{0}\\
IDM-occlusion-aware & \oncuenum{1}{78}{6} & \oncuenum{1}{0}{4} & \oncuenum{1}{2}{0}\\
PDM-Closed & \oncuenum{1}{39}{8} & \oncuenum{1}{26}{0} & \oncuenum{1}{9}{5}\\
LatentDriver & \oncuenum{1}{7}{4} & \oncuenum{1}{69}{7} & \oncuenum{1}{68}{1}\\
\midrule
\multicolumn{4}{@{}l}{\textbf{Cyclist}: 53 Anticipatory tests, 213 Reactive tests}\\
IDM-base & \oncuenum{1}{72}{2} & \oncuenum{1}{22}{6} & \oncuenum{1}{22}{3}\\
IDM-conservative & \oncuenum{1}{42}{6} & \oncuenum{1}{3}{0} & \oncuenum{1}{2}{4}\\
IDM-occlusion-aware & \oncuenum{1}{72}{2} & \oncuenum{1}{10}{1} & \oncuenum{1}{22}{3}\\
PDM-Closed & \oncuenum{1}{38}{2} & \oncuenum{1}{32}{4} & \oncuenum{1}{23}{4}\\
LatentDriver & \oncuenum{1}{1}{7} & \oncuenum{1}{76}{8} & \oncuenum{1}{78}{3}\\
\midrule
\multicolumn{4}{@{}l}{\textbf{Vehicle}: 31 Anticipatory tests, 55 Reactive tests}\\
IDM-base & \oncuenum{1}{92}{2} & \oncuenum{1}{3}{7} & \oncuenum{1}{0}{2}\\
IDM-conservative & \oncuenum{1}{93}{6} & \oncuenum{1}{0}{2} & \oncuenum{1}{0}{2}\\
IDM-occlusion-aware & \oncuenum{1}{92}{2} & \oncuenum{1}{1}{4} & \oncuenum{1}{0}{2}\\
PDM-Closed & \oncuenum{1}{51}{2} & \oncuenum{1}{34}{9} & \oncuenum{1}{38}{2}\\
LatentDriver & \oncuenum{1}{1}{4} & \oncuenum{1}{71}{2} & \oncuenum{1}{71}{2}\\

\bottomrule
\end{tabular*}}
\par\vspace{4pt}
{\fontsize{7.6}{8.8}\selectfont\setlength{\tabcolsep}{2pt}\renewcommand{\arraystretch}{1.00}
\oncuetablealign
\begin{tabular*}{\columnwidth}{@{\extracolsep{\fill}}lrrrrr@{}}
\toprule
\multicolumn{6}{@{}l}{\itshape (b) Scene validity: unweighted collision on tests valid for every planner}\\
& & & \multicolumn{3}{c}{Collision (\%) $\downarrow$}\\
\cmidrule(l){4-6}
& & & \multicolumn{2}{c}{IDM-base} & \shortstack{IDM-occlusion-\\aware}\\
\cmidrule(lr){4-5}
Label & \shortstack{Tests\\attempted} & \shortstack{Valid for\\every planner} & Masked & Unocc. & Masked\\
\midrule
\multicolumn{6}{@{}l}{\textbf{Pedestrian}}\\
Anticipatory & \oncuenum{0}{61}{} & \oncuenum{0}{61}{} & \oncuenum{1}{100}{0} & \oncuenum{1}{19}{7} & \oncuenum{1}{29}{5}\\
Reactive & \oncuenum{0}{216}{} & \oncuenum{0}{208}{} & \oncuenum{1}{18}{8} & \oncuenum{1}{1}{9} & \oncuenum{1}{2}{4}\\
\midrule
\multicolumn{6}{@{}l}{\textbf{Cyclist}}\\
Anticipatory & \oncuenum{0}{53}{} & \oncuenum{0}{49}{} & \oncuenum{1}{100}{0} & \oncuenum{1}{36}{7} & \oncuenum{1}{40}{8}\\
Reactive & \oncuenum{0}{213}{} & \oncuenum{0}{188}{} & \oncuenum{1}{28}{2} & \oncuenum{1}{21}{8} & \oncuenum{1}{9}{6}\\
\midrule
\multicolumn{6}{@{}l}{\textbf{Vehicle}}\\
Anticipatory & \oncuenum{0}{31}{} & \oncuenum{0}{31}{} & \oncuenum{1}{100}{0} & \oncuenum{1}{12}{9} & \oncuenum{1}{29}{0}\\
Reactive & \oncuenum{0}{55}{} & \oncuenum{0}{55}{} & \oncuenum{1}{21}{8} & \oncuenum{1}{1}{8} & \oncuenum{1}{5}{5}\\

\bottomrule
\end{tabular*}}
\end{table}
\par\noindent\textbf{Realized windows in closed loop.} Masked runs closely reproduce the generated window. Measured from the evaluated ego\textquotesingle{}s first observation of the target to the scheduled reference contact, shifted with a triggered target, masked fixed-timetable runs have a median $W_{\mathrm{run}}-W$ of 0.00\,s for every planner and 91.7\% to 100\% within 0.2\,s, among 6,440 runs of five planners on 644 tests in both observation modes. With triggered starts, medians are 0.01 to 0.04\,s and 88.7\% to 97.2\% are within 0.2\,s, except IDM-conservative (0.09\,s and 69.4\%). Untriggered or late runs, runs ending before the scheduled event, and runs that first see the target after it have no measured window. Unoccluded runs see the target earlier by design, with median windows 1.8 to 2.2\,s longer.
\par

\subsection*{F.1 Visibility and Avoidance Opportunities}\label{toc:F.1}
To relate planner outcomes to visibility and to the avoidance opportunities that labels mark, we compare Reactive and Anticipatory tests under both observation modes, then restrict the comparison to tests valid for every planner and observation mode. Per-window rates retain all attempts and are population weighted, so they describe each window\textquotesingle{}s eligible population.

\par\noindent\textbf{Results.}
With population weighting, IDM-base rarely collides on pedestrian and vehicle Reactive tests (6.5\% and 3.7\%, Table~\ref{table13}(a)), whereas PDM-Closed and LatentDriver miss many of the post-exposure opportunities that the oracle finds from the reference ego\textquotesingle{}s state. Without weighting, IDM-base collides in 18.8\% and 21.8\% of the masked pedestrian and vehicle Reactive tests valid for every planner (Table~\ref{table14}(b)), partly because the test list heavily oversamples short windows relative to their share of eligible tests, and IDM-base collides most at short windows: with population weighting, it collides in all 16 masked pedestrian Reactive tests at 0.5\,s and in 60\% and 70\% of vehicle ones at 0.5 and 1\,s (5 and 10 tests), but in at most 20\% from 1.5\,s on.

LatentDriver\textquotesingle{}s Reactive collision remains high in both observation modes. Its input crop (C.3) is applied after observation filtering, so even a geometrically visible target reaches the model only once it enters the crop.

\par\noindent\textbf{Finding.}
Response labels thus mark when avoidance becomes available, but a planner can act on it only once the target enters its input, which for LatentDriver can be later than geometric visibility.

\begin{table}[!t]
\centering
\caption{\textbf{Target timing versus visibility.} Results cover all 247 triggered-start tests, none of which is a vehicle test: a triggered target starts when the ego meets a position-and-speed trigger, and a fixed timetable starts it at the reference encounter\textquotesingle{}s time. $\Delta$ collision is triggered minus fixed timetable under masked observation, and masked minus unoccluded observation under fixed timetables.}\label{tab:fixed-target-clock}
{\fontsize{7.6}{8.8}\selectfont\setlength{\tabcolsep}{2pt}\renewcommand{\arraystretch}{1.00}
\oncuetablealign
\begin{tabular*}{\columnwidth}{@{\extracolsep{\fill}}lrrr@{}}
\toprule
& & \multicolumn{2}{c}{$\Delta$ collision (pp)}\\
\cmidrule(l){3-4}
Planner & \shortstack{Collision (\%), fixed\\timetable, masked $\downarrow$} & \shortstack{Triggered\\vs fixed} & \shortstack{Masked vs\\unoccluded}\\
\midrule\multicolumn{4}{@{}l}{\textbf{Pedestrian Reactive}: 59 tests}\\
IDM-base & \oncuenum{1}{28}{8} & \oncuenum{1}{$-$3}{4} & \oncuenum{1}{+23}{7}\\
IDM-conservative & \oncuenum{1}{0}{0} & \oncuenum{1}{+3}{4} & \oncuenum{1}{+0}{0}\\
IDM-occlusion-aware & \oncuenum{1}{1}{7} & \oncuenum{1}{+0}{0} & \oncuenum{1}{$-$3}{4}\\
PDM-Closed & \oncuenum{1}{33}{9} & \oncuenum{1}{+11}{9} & \oncuenum{1}{+8}{5}\\
LatentDriver & \oncuenum{1}{57}{6} & \oncuenum{1}{$-$1}{7} & \oncuenum{1}{+1}{7}\\
\midrule\multicolumn{4}{@{}l}{\textbf{Pedestrian Anticipatory}: 16 tests}\\
IDM-base & \oncuenum{1}{87}{5} & \oncuenum{1}{+12}{5} & \oncuenum{1}{+56}{2}\\
IDM-conservative & \oncuenum{1}{0}{0} & \oncuenum{1}{+12}{5} & \oncuenum{1}{+0}{0}\\
IDM-occlusion-aware & \oncuenum{1}{25}{0} & \oncuenum{1}{+0}{0} & \oncuenum{1}{$-$6}{2}\\
PDM-Closed & \oncuenum{1}{43}{8} & \oncuenum{1}{+31}{2} & \oncuenum{1}{+6}{2}\\
LatentDriver & \oncuenum{1}{50}{0} & \oncuenum{1}{+6}{2} & \oncuenum{1}{+0}{0}\\
\midrule\multicolumn{4}{@{}l}{\textbf{Cyclist Reactive}: 140 tests}\\
IDM-base & \oncuenum{1}{29}{3} & \oncuenum{1}{+1}{4} & \oncuenum{1}{+10}{7}\\
IDM-conservative & \oncuenum{1}{0}{0} & \oncuenum{1}{+9}{3} & \oncuenum{1}{+0}{0}\\
IDM-occlusion-aware & \oncuenum{1}{7}{1} & \oncuenum{1}{+2}{9} & \oncuenum{1}{$-$11}{4}\\
PDM-Closed & \oncuenum{1}{32}{9} & \oncuenum{1}{+15}{0} & \oncuenum{1}{+7}{9}\\
LatentDriver & \oncuenum{1}{68}{6} & \oncuenum{1}{+2}{1} & \oncuenum{1}{+0}{0}\\
\midrule\multicolumn{4}{@{}l}{\textbf{Cyclist Anticipatory}: 32 tests}\\
IDM-base & \oncuenum{1}{100}{0} & \oncuenum{1}{+0}{0} & \oncuenum{1}{+62}{5}\\
IDM-conservative & \oncuenum{1}{6}{2} & \oncuenum{1}{+15}{6} & \oncuenum{1}{+3}{1}\\
IDM-occlusion-aware & \oncuenum{1}{25}{0} & \oncuenum{1}{+0}{0} & \oncuenum{1}{$-$12}{5}\\
PDM-Closed & \oncuenum{1}{75}{0} & \oncuenum{1}{+9}{4} & \oncuenum{1}{+18}{8}\\
LatentDriver & \oncuenum{1}{78}{1} & \oncuenum{1}{+3}{1} & \oncuenum{1}{+0}{0}\\
\bottomrule
\end{tabular*}}

\end{table}
\subsection*{F.2 Observation Contrast with Fixed Timing}\label{toc:F.2}
With triggered starts, observation can change ego motion and hence the target\textquotesingle{}s start time (C.3), so the two observation modes may face different target motion. We therefore repeat all 644 tests with target timing fixed by the reference encounter. The 397 tests that already use fixed timing reproduce their outcomes, so Table~\ref{tab:fixed-target-clock} reports the 247 originally triggered standing-target tests.

\par\noindent\textbf{Results.}
On cyclist Reactive tests, fixed timing lowers PDM-Closed collision by 15.0 points, yet a masked-minus-unoccluded gap of 7.9 points remains. Under fixed timing, IDM-base keeps a 10.7-point gap on these tests, IDM-conservative rarely contacts the target in either mode, and IDM-occlusion-aware, which keeps only virtual vehicles beyond sensing range under unoccluded observations (C.3), collides like IDM-base when unoccluded.

\par\noindent\textbf{Finding.}
For IDM-base and PDM-Closed, the observation contrast therefore persists when both observation modes face identical target motion. Absolute collision rates can still differ between schedules, because triggered starts present the hazard as the ego approaches, whereas fixed timing keeps the scene clock regardless of when the ego arrives.

\begin{table}[!t]
\centering
\caption{\textbf{Reaction-window comparisons on the same situations.} (a) Each row keeps the situations with valid and complete runs at all three windows (1, 2, and 3\,s) that, in masked rows, are also not late, so counts can differ by planner and observation mode. Late, defined for masked runs only, means the ego trails the reference ego by more than 0.2\,s at exposure. (b) Restricts the masked columns of (a) to situations whose tests are Reactive at every window. Pedestrian results are unchanged. Rates are not population weighted. $\Delta$ collision is the 1\,s minus the 3\,s rate. Blue marks positive values, which mean fewer collisions at 3\,s. IDM-conservative is late in every masked situation, so its masked cells are empty (--).}\label{tab:matched-masked}
{\fontsize{7.6}{8.8}\selectfont\setlength{\tabcolsep}{2pt}\renewcommand{\arraystretch}{1.00}
\oncuetablealign
\begin{tabular*}{\columnwidth}{@{\extracolsep{\fill}}lrrrrr@{}}
\toprule
\multicolumn{6}{@{}l}{\itshape (a) Same situations: collision at $W^*=1$\,s and its change at 3\,s}\\
& \multicolumn{3}{c}{Masked} & \multicolumn{2}{c}{Unoccluded}\\
\cmidrule(lr){2-4}\cmidrule(l){5-6}
Planner & Situations & \shortstack{At 1\,s\\(\%) $\downarrow$} & \shortstack{$\Delta$ collision\\(pp)} & \shortstack{At 1\,s\\(\%) $\downarrow$} & \shortstack{$\Delta$ collision\\(pp)}\\
\midrule\multicolumn{6}{@{}l}{\textbf{Pedestrian}: 45 unoccluded situations for every planner}\\
IDM-base & \oncuenum{0}{45}{} & \oncuenum{1}{42}{2} & \cellcolor{OnCueGain}\oncuenum{1}{+42}{2} & \oncuenum{1}{0}{0} & \oncuenum{1}{0}{0}\\
IDM-conservative & \oncuenum{0}{0}{} & -- & -- & \oncuenum{1}{0}{0} & \oncuenum{1}{0}{0}\\
IDM-occlusion-aware & \oncuenum{0}{29}{} & \oncuenum{1}{37}{9} & \cellcolor{OnCueGain}\oncuenum{1}{+37}{9} & \oncuenum{1}{0}{0} & \oncuenum{1}{0}{0}\\
PDM-Closed & \oncuenum{0}{33}{} & \oncuenum{1}{66}{7} & \cellcolor{OnCueGain}\oncuenum{1}{+48}{5} & \oncuenum{1}{26}{7} & \cellcolor{OnCueGain}\oncuenum{1}{+15}{6}\\
LatentDriver & \oncuenum{0}{44}{} & \oncuenum{1}{88}{6} & \oncuenum{1}{$-$6}{8} & \oncuenum{1}{86}{7} & \oncuenum{1}{$-$6}{7}\\
\midrule\multicolumn{6}{@{}l}{\textbf{Cyclist}: 166 unoccluded situations for every planner}\\
IDM-base & \oncuenum{0}{163}{} & \oncuenum{1}{51}{5} & \cellcolor{OnCueGain}\oncuenum{1}{+42}{9} & \oncuenum{1}{30}{1} & \cellcolor{OnCueGain}\oncuenum{1}{+21}{1}\\
IDM-conservative & \oncuenum{0}{0}{} & -- & -- & \oncuenum{1}{0}{0} & \oncuenum{1}{0}{0}\\
IDM-occlusion-aware & \oncuenum{0}{74}{} & \oncuenum{1}{44}{6} & \cellcolor{OnCueGain}\oncuenum{1}{+27}{0} & \oncuenum{1}{30}{1} & \cellcolor{OnCueGain}\oncuenum{1}{+21}{1}\\
PDM-Closed & \oncuenum{0}{110}{} & \underline{\oncuenum{1}{59}{1}} & \cellcolor{OnCueGain}\oncuenum{1}{+53}{6} & \oncuenum{1}{25}{3} & \cellcolor{OnCueGain}\oncuenum{1}{+22}{9}\\
LatentDriver & \oncuenum{0}{162}{} & \oncuenum{1}{66}{0} & \oncuenum{1}{$-$16}{0} & \oncuenum{1}{66}{3} & \oncuenum{1}{$-$15}{1}\\
\midrule\multicolumn{6}{@{}l}{\textbf{Vehicle}: 80 unoccluded situations for every planner}\\
IDM-base & \oncuenum{0}{80}{} & \oncuenum{1}{92}{5} & \cellcolor{OnCueGain}\oncuenum{1}{+92}{5} & \oncuenum{1}{0}{0} & \oncuenum{1}{0}{0}\\
IDM-conservative & \oncuenum{0}{0}{} & -- & -- & \oncuenum{1}{0}{0} & \oncuenum{1}{0}{0}\\
IDM-occlusion-aware & \oncuenum{0}{68}{} & \oncuenum{1}{41}{2} & \cellcolor{OnCueGain}\oncuenum{1}{+41}{2} & \oncuenum{1}{0}{0} & \oncuenum{1}{0}{0}\\
PDM-Closed & \oncuenum{0}{63}{} & \oncuenum{1}{92}{1} & \cellcolor{OnCueGain}\oncuenum{1}{+92}{1} & \oncuenum{1}{43}{8} & \cellcolor{OnCueGain}\oncuenum{1}{+43}{8}\\
LatentDriver & \oncuenum{0}{63}{} & \oncuenum{1}{76}{2} & \oncuenum{1}{$-$17}{5} & \oncuenum{1}{68}{8} & \oncuenum{1}{$-$17}{5}\\
\midrule
\multicolumn{6}{@{}l}{\itshape (b) Situations Reactive at every window: masked collision}\\
Planner & Situations & \shortstack{At 1\,s\\(\%) $\downarrow$} & \shortstack{$\Delta$ collision\\(pp)} & & \\
\midrule\multicolumn{6}{@{}l}{\textbf{Cyclist}}\\
IDM-base & \oncuenum{0}{159}{} & \oncuenum{1}{50}{9} & \cellcolor{OnCueGain}\oncuenum{1}{+42}{1} &  & \\
IDM-occlusion-aware & \oncuenum{0}{72}{} & \oncuenum{1}{44}{4} & \cellcolor{OnCueGain}\oncuenum{1}{+26}{4} &  & \\
PDM-Closed & \oncuenum{0}{107}{} & \oncuenum{1}{57}{9} & \cellcolor{OnCueGain}\underline{\oncuenum{1}{+52}{3}} &  & \\
LatentDriver & \oncuenum{0}{159}{} & \oncuenum{1}{65}{4} & \oncuenum{1}{$-$16}{4} &  & \\
\midrule\multicolumn{6}{@{}l}{\textbf{Vehicle}}\\
IDM-base & \oncuenum{0}{20}{} & \oncuenum{1}{70}{0} & \cellcolor{OnCueGain}\oncuenum{1}{+70}{0} &  & \\
IDM-occlusion-aware & \oncuenum{0}{12}{} & \oncuenum{1}{58}{3} & \cellcolor{OnCueGain}\oncuenum{1}{+58}{3} &  & \\
PDM-Closed & \oncuenum{0}{11}{} & \oncuenum{1}{81}{8} & \cellcolor{OnCueGain}\oncuenum{1}{+81}{8} &  & \\
LatentDriver & \oncuenum{0}{18}{} & \oncuenum{1}{66}{7} & \oncuenum{1}{$-$33}{3} &  & \\
\bottomrule
\end{tabular*}}
\end{table}
\subsection*{F.3 More Response Time in the Same Situations}\label{toc:F.3}
To compare requested windows on the same situations, we vary only the requested window within each situation. From the generated outputs on the evaluation shards, we keep 45 pedestrian, 166 cyclist, and 80 vehicle situations that have a labeled, on-time test at each of 1, 2, and 3\,s, taking one test per window. The tests of a situation share the scene, replay start, background, reference-ego motion, route, and contact point, so only the target motion, including its hidden approach, changes. Nine situations also have optional 0.5\,s tests, giving 882 tests. Comparisons use the three common windows with fixed target timetables. Each row keeps the situations with valid and complete runs at all three windows, which in masked rows must also not be late (C.3), so its denominator is fixed across windows but can differ across planners and observation modes. Rates are not population weighted.

\par\noindent\textbf{Results.}
IDM-conservative is late in every masked situation, so only its unoccluded runs form a complete three-window cohort. PDM-Closed\textquotesingle{}s large collision decrease at longer windows remains when every window must also be Reactive, a condition that all pedestrian situations already meet (Table~\ref{tab:matched-masked}).

LatentDriver shows the opposite trend, colliding more at 3\,s than at 1\,s in both observation modes, most for cyclists and vehicles (Table~\ref{tab:matched-masked}). At longer windows it brakes hard (peak deceleration of at least 3\,m/s$^2$) less often, in 25.9\% of masked cyclist runs at 1\,s but 13.3\% at 3\,s (each window\textquotesingle{}s valid, complete runs), and runs with hard braking rarely collide (at most 3.3\%, against about 90\% without). Part of its short-window avoidance can come from passing first: in the 29 cyclist situations where it avoids collision at 1\,s but collides at 3\,s, it leads the reference ego by a median of about 1\,s at exposure in the 1\,s test, against 0.15\,s in the 104 that collide at both windows.

\par\noindent\textbf{Finding.}
A short-window non-collision therefore does not by itself show a good response, because under fixed target timing an ego that arrives early can pass before the target. We therefore report collision together with safe-response success and executed braking.

\subsection*{F.4 Results without Target-Background Contact}\label{toc:F.4}
To check that planner results do not depend on tests whose target also contacts background road users, we exclude the 124 of 644 tests with target-background contact, and every same-situation set containing such a test at any window, and use sample rates without population weighting.

\par\noindent\textbf{Results.}
Pedestrian and cyclist comparisons keep their directions. On the remaining 53 pedestrian and 43 cyclist Anticipatory tests, IDM-base still collides on every masked run, against unoccluded collision of 22.6\% and 34.9\%. Across the same cyclist situations at 1 and 3\,s, PDM-Closed collision still falls by 49.5 points over 91 situations, and LatentDriver collision still rises by 18.3 points over 142. Vehicle evidence becomes thin, because vehicle Anticipatory tests fall from 31 to 6 and vehicle situations roughly halve.

\par\noindent\textbf{Finding.}
The pedestrian and cyclist trends therefore do not depend on tests with target-background contact, whereas the vehicle Anticipatory results rest on too few remaining tests to be checked this way.

\subsection*{F.5 Training a Planner on the Tests}\label{toc:F.5}
Training on OnCue tests adds responses to occluded hazards but also changes ordinary driving (Tables~\ref{tab:q6-costs} and~\ref{tab:q6-natural}, means over three seeds). For LatentDriver on the 484 Reactive tests, Logs + OnCue lowers target collision from 78.4\% with logs only to 23.7\% and raises safe-response success from 6.5\% to 37.8\%. The collision reduction has a driving-episode 95\% interval of $[49.0, 60.3]$ points for LatentDriver and $[41.3, 53.0]$ for PlanT. Logs + OnCue also leaves more runs untriggered or late (28.6\% against 11.1\%), contacts other road users more often on the 644 generated tests (4.6\% against 0.4\%), and covers less no-target distance in 8\,s (58.4 against 71.1\,m). In natural-log replay it shows no demonstrated gain and contacts targets more often (23.4\% against 3.4\%), mostly in follower encounters, where logged followers cannot brake for its slowdowns (F.5.4). Training must therefore balance avoidance against driving efficiency and traffic interaction.

We examine the main-paper training result through four questions. F.5.1 asks whether slower driving or triggered starts explain the reduction, and F.5.2 whether it holds beyond OnCue\textquotesingle{}s own tests, planner, and demonstrated conditions. F.5.3 then asks what the gain costs, and F.5.4 how trained planners behave in natural logged encounters, where no target is inserted. Slowing explains only part of the reduction, since a slowed control at matched distance recovers $57.0\pm2.1\%$ of it (F.5.1). The reduction carries over to other generators\textquotesingle{} tests, a second planner, and nearby held-out conditions (F.5.2), and masked-only demonstrations reduce hesitation at a cost of $3.2\pm0.4$ points of the collision reduction (F.5.3).

On 16 WOMD training shards disjoint from evaluation, demonstrations execute a safe oracle response that never speeds up, preferring lane keeping and then the least peak braking, or the oracle\textquotesingle{}s stored response when none qualifies. Masked demonstrations come only from Reactive tests and respond from $t_e$, whereas unoccluded ones also come from Anticipatory tests and respond from $t_v$. Main training starts from the pretrained checkpoint and mixes demonstration and logged windows equally for five epochs at $2\times10^{-5}$, keeping model inputs, upstream loss, and optimizer unchanged, and the logs-only control uses the same number of optimization steps. This control is needed because fine-tuning also adapts the pretrained checkpoint to our map-continuation route (C.3). The checkpoint was trained with logged-trajectory routes, whose length reveals human progress, so on our route it drives farther, contacts other agents more often, and collides with the target less often than the logs-only control (Table~\ref{tab:q6-costs}). Summaries use three independent training seeds unless a single seed is stated. A reduction is the within-seed paired difference in target collision, logs-only minus trained model, in percentage points computed before rounding.

The learning rate of $2\times10^{-5}$ was chosen on 515 tests from development shards 0--11, separate from the 515 Q4 candidates and drawn from driving episodes disjoint from evaluation.

\begin{table}[!t]
\centering
\caption{\textbf{Training-gain checks and late runs under fixed target timing.} (a) Whether the reduction holds in each subset and setting. (b) How often trained and logs-only models arrive late. Logs only is the pretrained model fine-tuned on logged windows only, for the same number of steps. Observation is masked unless a row says unoccluded. Test-count ranges vary by seed. $\Delta$ collision is logs-only minus trained collision within each seed. Both runs means the trained and the logs-only run, and not slower than logs only keeps tests where the trained model, run without the target, travels at least as far in 8\,s. In the last two rows, the filtered logs-only model collides on the same tests in both observation modes.}\label{tab:q6-gain}
{\fontsize{7.6}{8.8}\selectfont\setlength{\tabcolsep}{2pt}\renewcommand{\arraystretch}{1.0}\oncuetablealign
\begin{tabular*}{\columnwidth}{@{\extracolsep{\fill}}>{\raggedright\arraybackslash}p{118pt}rrr@{}}
\toprule
\multicolumn{4}{@{}l}{\itshape (a) Training-gain checks: target collision, Reactive tests and subsets}\\
& & \multicolumn{2}{c}{Target collision}\\
\cmidrule(l){3-4}
Test subset & Tests & \shortstack{Logs only\\(\%)} & \shortstack{$\Delta$ collision\\(pp) $\uparrow$}\\
\midrule
\multicolumn{4}{@{}l}{\textbf{LatentDriver}, triggered starts (main setting)}\\
\leftskip=6pt\relax All & \oncuenum{0}{484}{} & \oncuenum{1}{78}{4}\oncuestd{0.2} & \underline{\oncuenum{1}{54}{8}}\oncuestd{0.4}\\
\leftskip=6pt\relax All, unoccluded & \oncuenum{0}{484}{} & \oncuenum{1}{78}{2}\oncuestd{0.2} & \oncuenum{1}{55}{4}\oncuestd{0.1}\\
\leftskip=6pt\relax Both runs triggered, not late & 332--336 & \oncuenum{1}{86}{9}\oncuestd{0.5} & \underline{\oncuenum{1}{53}{8}}\oncuestd{0.8}\\
\leftskip=6pt\relax Not slower than logs only & 114--116 & \oncuenum{1}{77}{1}\oncuestd{0.5} & \underline{\oncuenum{1}{35}{1}}\oncuestd{0.8}\\
\leftskip=6pt\relax w/o target-background contact & \oncuenum{0}{418}{} & \oncuenum{1}{78}{6}\oncuestd{0.3} & \underline{\oncuenum{1}{56}{6}}\oncuestd{0.3}\\
\midrule
\multicolumn{4}{@{}l}{\textbf{LatentDriver}, fixed target timing}\\
\leftskip=6pt\relax All & \oncuenum{0}{484}{} & \oncuenum{1}{71}{3}\oncuestd{0.4} & \oncuenum{1}{51}{4}\oncuestd{0.7}\\
\leftskip=6pt\relax Both runs not late & 227--232 & \oncuenum{1}{89}{0}\oncuestd{0.4} & \oncuenum{1}{49}{8}\oncuestd{0.6}\\
\midrule
\multicolumn{4}{@{}l}{\textbf{PlanT}, fixed target timing}\\
\leftskip=6pt\relax All & \oncuenum{0}{484}{} & \oncuenum{1}{81}{1}\oncuestd{0.3} & \oncuenum{1}{51}{2}\oncuestd{0.2}\\
\leftskip=6pt\relax Both runs not late & \oncuenum{0}{280}{} & \oncuenum{1}{86}{5}\oncuestd{0.5} & \oncuenum{1}{36}{2}\oncuestd{0.5}\\
\midrule
\multicolumn{4}{@{}l}{\textbf{LatentDriver}, the 199 standing-target tests}\\
\leftskip=6pt\relax Triggered start & \oncuenum{0}{199}{} & \oncuenum{1}{85}{1}\oncuestd{0.3} & \oncuenum{1}{62}{3}\oncuestd{1.3}\\
\leftskip=6pt\relax Fixed timing & \oncuenum{0}{199}{} & \oncuenum{1}{67}{8}\oncuestd{0.9} & \oncuenum{1}{54}{1}\oncuestd{1.6}\\
\midrule
\multicolumn{4}{@{}l}{\textbf{LatentDriver} trained w/o stationary windows, vs its filtered logs-only model}\\
\leftskip=6pt\relax Not slower than logs only & 102--106 & \oncuenum{1}{76}{3}\oncuestd{1.5} & \oncuenum{1}{32}{7}\oncuestd{2.3}\\
\leftskip=6pt\relax Not slower than logs only, unoccluded & 102--106 & \oncuenum{1}{76}{3}\oncuestd{1.5} & \oncuenum{1}{33}{6}\oncuestd{2.1}\\
\bottomrule
\end{tabular*}}
\par\vspace{4pt}
{\fontsize{7.6}{8.8}\selectfont\setlength{\tabcolsep}{2pt}\renewcommand{\arraystretch}{1.0}\oncuetablealign
\begin{tabular*}{\columnwidth}{@{\extracolsep{\fill}}>{\raggedright\arraybackslash}p{99pt}rrr@{}}
\toprule
\multicolumn{4}{@{}l}{\itshape (b) Late runs under fixed target timing: 484 Reactive tests}\\
& & \multicolumn{2}{c}{Late runs (\%) $\downarrow$}\\
\cmidrule(l){3-4}
Planner & Tests & \shortstack{Logs\\only} & \shortstack{Logs +\\OnCue}\\
\midrule
LatentDriver & \oncuenum{0}{484}{} & \oncuenum{1}{22}{7}\oncuestd{0.5} & \oncuenum{1}{48}{3}\oncuestd{0.6}\\
PlanT & \oncuenum{0}{484}{} & \oncuenum{1}{10}{4}\oncuestd{0.1} & \oncuenum{1}{41}{0}\oncuestd{0.1}\\
\bottomrule
\end{tabular*}}
\end{table}

\begin{figure}[!t]
\centering
\includegraphics[width=\columnwidth]{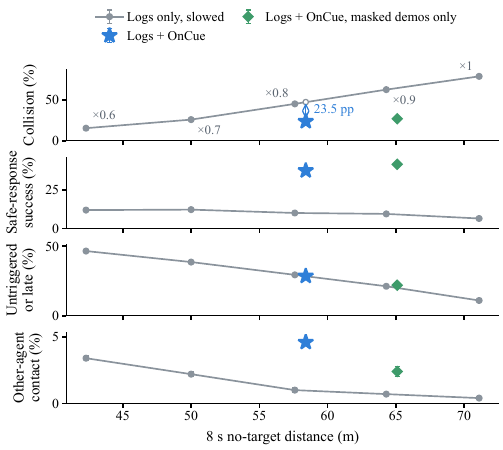}
\caption{\textbf{Slowed logs-only control.} Gray: logs-only actions scaled by the marked factor at evaluation. Untriggered means the standing target never starts, and late that the ego trails the reference ego by more than 0.2\,s at exposure. The two other variants lie within 1\,m of Logs + OnCue\textquotesingle{}s distance and are given in Table~\ref{tab:q6-costs}. The open circle is the slowed control interpolated to Logs + OnCue\textquotesingle{}s distance, and the arrow is the collision gap, $23.5\pm1.0$ points ($43.0\pm2.1\%$ of Logs + OnCue\textquotesingle{}s collision reduction over the unscaled control, so the slowed control recovers $57.0\pm2.1\%$). Against $\times0.8$ the paired gap is $21.3\pm0.4$ points. Triggered starts, means $\pm$ SD over seeds. Rates use the 484 Reactive tests, and contact and distance all 644 tests.}\label{fig:q6-slowed}
\end{figure}
\subsubsection*{F.5.1 Slower Driving and Triggered Starts}\label{toc:F.5.1}
A planner that simply drives more slowly also collides less, and triggered starts could favor a slower ego. Three checks on the 484 Reactive tests address both: a slowed logs-only control, the outcome-conditioned subsets of Table~\ref{tab:q6-gain}, and fixed target timetables, under which no target waits for the ego. Reruns on tests that already use fixed timing reproduce the original outcomes for all six pairs of OnCue-trained and logs-only models.

The slowed control scales each logs-only action at evaluation, that is, the per-step displacement and heading change, by ${\rho\in\{0.9,0.8,0.7,0.6\}}$. Scaling keeps each step\textquotesingle{}s curvature and lowers speed roughly $\rho$-fold, and the scaled actions are fed back as the model\textquotesingle{}s action history. With $\rho=1$, the scaling reproduces all 1,288 seed-0 control runs on generated tests field by field, and the unscaled control runs match the main-paper runs on all 644 generated tests in each of the three seeds.

\par\noindent\textbf{Results.}
At OnCue training\textquotesingle{}s no-target distance, the slowed control collides 23.5 points more and rarely produces safe responses, although it is untriggered or late about as often (Figure~\ref{fig:q6-slowed}). The reduction remains in every subset of Table~\ref{tab:q6-gain} and under fixed timetables for both planners. It is smallest on tests where the trained model is not slower than its logs-only model and for PlanT without late runs. Under fixed timetables, OnCue-trained planners also arrive late more often than their controls.

\par\noindent\textbf{Finding.}
Training therefore adds timely responses rather than only later arrival: a uniformly slowed control at matched distance recovers $57.0\pm2.1\%$ of the reduction, but slowing alone does not produce OnCue training\textquotesingle{}s safe responses. For LatentDriver, triggered starts raise collision for both the trained model and its control, more for the control, because a triggered target waits for the ego whereas a fixed-timetable target can pass before a late ego. Triggered starts therefore add a few points to the reduction, and fixed timetables keep most of it. For PlanT, triggered starts raise the trained model\textquotesingle{}s collision more than the control\textquotesingle{}s, so its reduction is about 4 points larger under fixed timetables.

\begin{table}[!t]
\centering
\caption{\textbf{Transfer across tests, planners, and conditions.} (a) Whether the collision reduction of Logs + OnCue over logs only holds on other generators\textquotesingle{} tests and for PlanT. (b) Whether it holds for conditions left out of the demonstrations. Held-out models use fewer training steps. $\Delta$ collision is logs-only minus Logs + OnCue masked target collision, and the increase is held-out minus all-demonstration collision, in percentage points. Values are mean $\pm$ SD over three seeds. Underlined: OnCue\textquotesingle{}s own tests, for reference.}\label{tab:q6-transfer}
{\fontsize{7.6}{8.8}\selectfont\setlength{\tabcolsep}{2pt}\renewcommand{\arraystretch}{1.0}\oncuetablealign
\begin{tabular*}{\columnwidth}{@{\extracolsep{\fill}}lrrr@{}}
\toprule
\multicolumn{4}{@{}l}{\itshape (a) Tests and planners: Reactive vehicle tests by generator, fixed timing}\\
& & \multicolumn{2}{c}{$\Delta$ collision (pp) $\uparrow$}\\
\cmidrule(l){3-4}
Test generator & Tests & LatentDriver & PlanT\\
\midrule
\underline{OnCue} & \oncuenum{0}{300}{} & \oncuenum{1}{43}{0}\oncuestd{0.9} & \oncuenum{1}{19}{2}\oncuestd{2.6}\\
CAT & \oncuenum{0}{300}{} & \oncuenum{1}{46}{2}\oncuestd{0.7} & \oncuenum{1}{21}{0}\oncuestd{1.5}\\
SAGE ($w=0.5$) & \oncuenum{0}{300}{} & \oncuenum{1}{45}{4}\oncuestd{0.8} & \oncuenum{1}{19}{7}\oncuestd{2.4}\\
SAGE ($w=1.0$) & \oncuenum{0}{179}{} & \oncuenum{1}{45}{1}\oncuestd{0.6} & \oncuenum{1}{19}{9}\oncuestd{1.6}\\
SEAL & \oncuenum{0}{300}{} & \oncuenum{1}{46}{0}\oncuestd{0.3} & \oncuenum{1}{21}{4}\oncuestd{1.7}\\
GOOSE & \oncuenum{0}{300}{} & \oncuenum{1}{49}{8}\oncuestd{1.7} & \oncuenum{1}{25}{7}\oncuestd{1.0}\\
KING & \oncuenum{0}{300}{} & \oncuenum{1}{42}{7}\oncuestd{0.3} & \oncuenum{1}{28}{3}\oncuestd{1.5}\\
All external, pooled & \oncuenum{0}{1,679}{} & \oncuenum{1}{45}{9}\oncuestd{0.6} & \oncuenum{1}{22}{9}\oncuestd{1.5}\\
\midrule\multicolumn{4}{@{}l}{\itshape (b) Conditions left out of demonstrations: that condition\textquotesingle{}s Reactive tests}\\
& & \multicolumn{2}{c}{LatentDriver, triggered starts}\\
\cmidrule(l){3-4}
Left out & Tests & \shortstack{Increase vs\\all demos (pp) $\downarrow$} & \shortstack{$\Delta$ collision\\(pp) $\uparrow$}\\
\midrule
Pedestrian & \oncuenum{0}{216}{} & \oncuenum{1}{4}{0}\oncuestd{0.3} & \oncuenum{1}{52}{2}\oncuestd{1.1} \\
Cyclist & \oncuenum{0}{213}{} & \oncuenum{1}{1}{4}\oncuestd{1.4} & \oncuenum{1}{54}{3}\oncuestd{1.4} \\
Vehicle & \oncuenum{0}{55}{} & \oncuenum{1}{21}{2}\oncuestd{2.8} & \oncuenum{1}{24}{2}\oncuestd{1.0}\\
$W^*=1.0$\,s & \oncuenum{0}{90}{} & \oncuenum{1}{2}{2}\oncuestd{3.3} & \oncuenum{1}{55}{6}\oncuestd{1.1} \\
\bottomrule
\end{tabular*}}
\end{table}

\subsubsection*{F.5.2 Transfer to Other Tests, Planners, and Conditions}\label{toc:F.5.2}
To test whether the gain depends on OnCue\textquotesingle{}s own target motion, we evaluate valid Reactive vehicle tests that the other generators produce from the shared requests. Tests are drawn in the main list\textquotesingle{}s order but without strata, with a separate seed and at most 300 tests per source, from 2,353 eligible tests for OnCue, 1,354 for CAT, 452 and 179 for SAGE at $w=0.5$ and $1.0$, 985 for SEAL, 1,073 for GOOSE, and 1,418 for KING. These tests keep OnCue\textquotesingle{}s requests, occlusion layouts, and starts, and each target follows its generator\textquotesingle{}s fixed timetable under both observation modes. Lateness is not measured on these tests.

To test a second learned planner, we use the Waymax PlanT implementation released with LatentDriver~\cite{latentdriver,plant}, which shares LatentDriver\textquotesingle{}s observation filtering and waypoint execution. Both PlanT training settings use the same data and five-epoch recipe, with PlanT\textquotesingle{}s one-cycle schedule peaking at $2\times10^{-5}$, and three seeds. Evaluation follows the main 644-test protocol under both observation modes, and a rerun of pretrained LatentDriver in this setup reproduces all 1,288 main-cohort runs.

To test conditions absent from demonstrations, we remove all pedestrian, cyclist, vehicle, or $W^*=1.0$\,s demonstrations and evaluate only that condition\textquotesingle{}s Reactive tests. Logged data are unchanged, so the smaller demonstration pools also reduce the training steps in each of the three seeds.

\par\noindent\textbf{Results.}
The training gain carries over to other generators\textquotesingle{} tests and to a second planner. On pooled external tests, the reduction is $45.9\pm0.6$ points for LatentDriver and $22.9\pm1.5$ points for PlanT, at least as large as on the sampled OnCue tests (Table~\ref{tab:q6-transfer}(a)), and LatentDriver keeps $45.9\pm0.8$ points under unoccluded observation. PlanT\textquotesingle{}s smaller external reduction reflects the vehicle-only test sets: its reduction is $47.2\pm0.6$ points on the 484 main Reactive tests (Table~\ref{tab:q6-costs}) but only $24.8\pm1.0$ points on their vehicle tests, about half of its pedestrian and cyclist reductions ($52.8\pm0.8$ and $47.4\pm1.4$ points).

Leaving out one condition\textquotesingle{}s demonstrations keeps a reduction in every case and costs most for vehicles, where about half of the vehicle reduction remains. The other exclusions raise collision by at most 4.0 points (Table~\ref{tab:q6-transfer}(b)).

\par\noindent\textbf{Finding.}
The gain is therefore not specific to OnCue\textquotesingle{}s own target motion or to LatentDriver, although the external tests keep OnCue\textquotesingle{}s starts and occlusion layouts, and their fixed timetables leave lateness unmeasured in both observation modes, so part of the external gain may come from late arrivals. The gain also persists for conditions omitted from demonstrations, but this evidence is weaker: the pedestrian, cyclist, and 1\,s conditions stay close to trained ones (1\,s windows, for example, lie between trained 0.5 and 1.5\,s windows), and the vehicle evidence rests on 55 tests. Because held-out training changes demonstration coverage and training budget together, the cost of leaving out a condition cannot be attributed to coverage alone.

\begin{table*}[!t]
\centering
\caption{\textbf{Training gains, costs, and data ablations.} Pretrained is the checkpoint without fine-tuning. Logs only fine-tunes it on logged windows. Logs + OnCue (bold, the main-paper configuration) mixes these equally, for the same number of steps, with demonstrations, oracle safe responses on OnCue tests from training shards disjoint from evaluation, in both observation modes. Indented rows each change one setting of Logs + OnCue (F.5.3): 25\% demos halves its 50\% demonstration share, and w/o stationary windows is compared with a logs-only model trained on the same filtered windows. $\Delta$ collision is logs-only minus trained collision, paired within each seed. Other-agent contact is at-fault contact with road users other than the target at any time in the run (up to 8\,s). Rates are unweighted, and trained models give mean $\pm$ SD over three seeds.}\label{tab:q6-costs}
{\fontsize{7.6}{8.8}\selectfont\setlength{\tabcolsep}{2pt}\renewcommand{\arraystretch}{1.0}\oncuetablealign
\begin{tabular*}{\textwidth}{@{\extracolsep{\fill}}llrrrrrr@{}}
\toprule
& & \multicolumn{4}{c}{\itshape Reactive tests ($n=484$)} & \multicolumn{2}{c}{\itshape All generated tests ($n=644$)}\\
\cmidrule(lr){3-6}\cmidrule(l){7-8}
Planner & Training & \shortstack{Target\\collision (\%) $\downarrow$} & \shortstack{$\Delta$ collision\\(pp) $\uparrow$} & \shortstack{Safe-response\\success (\%) $\uparrow$} & \shortstack{Untriggered\\or late (\%) $\downarrow$} & \shortstack{Other-agent\\contact (\%) $\downarrow$} & \shortstack{8\,s no-target\\distance (m) $\uparrow$}\\
\midrule
LatentDriver & Pretrained & \oncuenum{1}{73}{3}\oncuenostd & \multicolumn{1}{c}{---} & \oncuenum{1}{8}{1}\oncuenostd & \oncuenum{1}{0}{2}\oncuenostd & \oncuenum{1}{17}{5}\oncuenostd & \oncuenum{1}{103}{1}\oncuenostd\\
 & Logs only & \oncuenum{1}{78}{4}\oncuestd{0.2} & \multicolumn{1}{c}{---} & \underline{\oncuenum{1}{6}{5}}\oncuestd{0.4} & \oncuenum{1}{11}{1}\oncuestd{0.3} & \oncuenum{1}{0}{4}\oncuestd{0.1} & \oncuenum{1}{71}{1}\oncuestd{0.1}\\
 & \textbf{Logs + OnCue} & \oncuenum{1}{23}{7}\oncuestd{0.1} & \underline{\oncuenum{1}{54}{8}}\oncuestd{0.4} & \underline{\oncuenum{1}{37}{8}}\oncuestd{0.2} & \oncuenum{1}{28}{6}\oncuestd{0.4} & \oncuenum{1}{4}{6}\oncuestd{0.2} & \oncuenum{1}{58}{4}\oncuestd{0.2}\\
 & \hspace{6pt}w/o stationary windows & \oncuenum{1}{26}{6}\oncuestd{0.6} & \oncuenum{1}{52}{2}\oncuestd{1.1} & \oncuenum{1}{35}{5}\oncuestd{0.7} & \oncuenum{1}{29}{1}\oncuestd{0.2} & \oncuenum{1}{2}{1}\oncuestd{0.2} & \oncuenum{1}{59}{4}\oncuestd{0.2}\\
 & \hspace{6pt}25\% demos & \oncuenum{1}{24}{3}\oncuestd{0.2} & \oncuenum{1}{54}{1}\oncuestd{0.4} & \oncuenum{1}{36}{8}\oncuestd{0.3} & \oncuenum{1}{28}{6}\oncuestd{0.3} & \oncuenum{1}{2}{3}\oncuestd{0.3} & \oncuenum{1}{59}{0}\oncuestd{0.2}\\
 & \hspace{6pt}masked demos only & \oncuenum{1}{26}{9}\oncuestd{0.5} & \oncuenum{1}{51}{6}\oncuestd{0.8} & \oncuenum{1}{41}{9}\oncuestd{0.2} & \oncuenum{1}{22}{1}\oncuestd{0.2} & \oncuenum{1}{2}{4}\oncuestd{0.4} & \oncuenum{1}{65}{1}\oncuestd{0.1}\\
\midrule\addlinespace[3pt]
PlanT & Pretrained & \oncuenum{1}{25}{4}\oncuenostd & \multicolumn{1}{c}{---} & \oncuenum{1}{7}{6}\oncuenostd & \oncuenum{1}{0}{2}\oncuenostd & \oncuenum{1}{53}{6}\oncuenostd & \oncuenum{1}{125}{8}\oncuenostd\\
 & Logs only & \oncuenum{1}{83}{1}\oncuestd{0.0} & \multicolumn{1}{c}{---} & \oncuenum{1}{8}{1}\oncuestd{0.2} & \oncuenum{1}{5}{7}\oncuestd{0.1} & \oncuenum{1}{1}{0}\oncuestd{0.1} & \oncuenum{1}{78}{0}\oncuestd{0.4}\\
 & \textbf{Logs + OnCue} & \oncuenum{1}{35}{8}\oncuestd{0.6} & \oncuenum{1}{47}{2}\oncuestd{0.6} & \oncuenum{1}{31}{6}\oncuestd{0.4} & \oncuenum{1}{23}{4}\oncuestd{0.3} & \oncuenum{1}{1}{7}\oncuestd{0.1} & \oncuenum{1}{63}{5}\oncuestd{0.3}\\
\bottomrule
\end{tabular*}}
\end{table*}
\begin{table*}[!t]
\caption{\textbf{Target contact and ego behavior in natural-log encounters.} All 148 natural encounters use masked observation, and columns without a count use all of them. Target contact counts contact before the test cutoff (C.3). (a) Under log replay and with following targets that brake. Encounters are classified as follower or other before evaluation, ego at fault only uses the background-contact rule (B.2), and braking followers brake within the Gipps bound and see the ego 0.1 or 1.0\,s late (F.5.4). (b) No-target distance is an 8\,s mean. Training rows are as in Table~\ref{tab:q6-costs}. Trained models give mean $\pm$ SD over three seeds, and pretrained models, which are not fine-tuned, one run.}\label{tab:q6-natural}
\begin{minipage}[t]{\textwidth}
\vspace{0pt}
{\fontsize{7.6}{8.8}\selectfont\setlength{\tabcolsep}{3pt}\renewcommand{\arraystretch}{1.0}\oncuetablealign
\begin{tabular*}{\textwidth}{@{\extracolsep{\fill}}l>{\raggedright\arraybackslash}p{80pt}rrrrrr@{}}
\toprule
\multicolumn{8}{@{}l}{\itshape (a) Target contact: log replay and braking followers}\\
& & \multicolumn{4}{c}{Log-replay target contact (\%) $\downarrow$} & \multicolumn{2}{c}{\shortstack{Target contact when followers\\brake (\%) $\downarrow$}}\\
\cmidrule(lr){3-6}\cmidrule(l){7-8}
Planner & Training & All (148) & \shortstack{Follower\\encounters (109)} & \shortstack{Other\\encounters (39)} & \shortstack{Ego at\\fault only} & \shortstack{See ego\\0.1\,s late} & \shortstack{See ego\\1.0\,s late}\\
\midrule
LatentDriver & Pretrained & \oncuenum{1}{17}{6}\oncuenostd & \oncuenum{1}{0}{9}\oncuenostd & \oncuenum{1}{64}{1}\oncuenostd & \oncuenum{1}{14}{9}\oncuenostd & \oncuenum{1}{16}{9}\oncuenostd & \oncuenum{1}{16}{9}\oncuenostd\\
 & Logs only & \oncuenum{1}{3}{4}\oncuestd{0.0} & \oncuenum{1}{0}{0}\oncuestd{0.0} & \oncuenum{1}{12}{8}\oncuestd{0.0} & \oncuenum{1}{2}{5}\oncuestd{0.4} & \oncuenum{1}{3}{4}\oncuestd{0.0} & \oncuenum{1}{3}{4}\oncuestd{0.0}\\
 & \textbf{Logs + OnCue} & \oncuenum{1}{23}{4}\oncuestd{0.8} & \oncuenum{1}{26}{3}\oncuestd{1.1} & \oncuenum{1}{15}{4}\oncuestd{0.0} & \oncuenum{1}{3}{4}\oncuestd{0.0} & \oncuenum{1}{4}{1}\oncuestd{0.0} & \oncuenum{1}{6}{1}\oncuestd{0.0}\\
 & \hspace{6pt}masked demos only & \oncuenum{1}{12}{4}\oncuestd{1.0} & \oncuenum{1}{10}{7}\oncuestd{1.1} & \oncuenum{1}{17}{1}\oncuestd{1.5} & \oncuenum{1}{4}{3}\oncuestd{0.8} & \oncuenum{1}{5}{0}\oncuestd{0.8} & \oncuenum{1}{5}{0}\oncuestd{0.8}\\
\midrule\addlinespace[3pt]
PlanT & Pretrained & \oncuenum{1}{14}{9}\oncuenostd & \oncuenum{1}{0}{0}\oncuenostd & \oncuenum{1}{56}{4}\oncuenostd & \oncuenum{1}{14}{2}\oncuenostd & \oncuenum{1}{14}{9}\oncuenostd & \oncuenum{1}{14}{9}\oncuenostd\\
 & Logs only & \oncuenum{1}{2}{5}\oncuestd{0.4} & \oncuenum{1}{1}{5}\oncuestd{0.5} & \oncuenum{1}{5}{1}\oncuestd{2.6} & \oncuenum{1}{1}{4}\oncuestd{0.7} & \oncuenum{1}{1}{4}\oncuestd{0.7} & \oncuenum{1}{1}{4}\oncuestd{0.7}\\
 & \textbf{Logs + OnCue} & \oncuenum{1}{34}{2}\oncuestd{0.4} & \oncuenum{1}{43}{1}\oncuestd{0.9} & \oncuenum{1}{9}{4}\oncuestd{1.5} & \oncuenum{1}{2}{5}\oncuestd{0.4} & \oncuenum{1}{2}{5}\oncuestd{0.4} & \oncuenum{1}{11}{3}\oncuestd{1.0}\\
\bottomrule
\end{tabular*}\par\vspace{4pt}
\begin{tabular*}{0.56\textwidth}{@{\extracolsep{\fill}}l>{\raggedright\arraybackslash}p{80pt}rr@{}}
\toprule
\multicolumn{4}{@{}l}{\itshape (b) Ego behavior: stops and progress}\\
Planner & Training & \shortstack{Runs that\\stop ($n$)} & \shortstack{8\,s no-target\\distance (m) $\uparrow$}\\
\midrule
LatentDriver & Pretrained & \oncuenum{0}{0}{}\hspace{\oncuefracone}\oncuenostd & \oncuenum{1}{76}{0}\oncuenostd\\
 & Logs only & \oncuenum{1}{4}{0}\oncuestd{0.0} & \oncuenum{1}{52}{9}\oncuestd{0.0}\\
 & \textbf{Logs + OnCue} & \oncuenum{1}{28}{7}\oncuestd{0.6} & \oncuenum{1}{41}{3}\oncuestd{0.2}\\
 & \hspace{6pt}masked demos only & \oncuenum{1}{21}{0}\oncuestd{1.0} & \oncuenum{1}{44}{3}\oncuestd{0.1}\\
\midrule\addlinespace[3pt]
PlanT & Pretrained & \oncuenum{0}{0}{}\hspace{\oncuefracone}\oncuenostd & \oncuenum{1}{94}{9}\oncuenostd\\
 & Logs only & \oncuenum{1}{1}{7}\oncuestd{1.2} & \oncuenum{1}{52}{2}\oncuestd{0.1}\\
 & \textbf{Logs + OnCue} & \oncuenum{1}{22}{0}\oncuestd{1.0} & \oncuenum{1}{38}{1}\oncuestd{0.2}\\
\bottomrule
\end{tabular*}}
\end{minipage}
\end{table*}

\subsubsection*{F.5.3 Costs and the Source of Hesitation}\label{toc:F.5.3}
To locate the costs, we compare three variants of Logs + OnCue on the same evaluation tests, shown as indented rows under Logs + OnCue in Table~\ref{tab:q6-costs} (Table~\ref{tab:q6-natural} keeps only the masked-demonstration variant). The variant without stationary windows drops training windows in which ego speed is at most 0.05\,m/s at both training steps (11.9\% of demonstration and 28.7\% of logged windows) and is compared with a logs-only control filtered the same way. The other two keep the main training settings and budget and change only how demonstrations enter training: the 25\% demos variant draws a quarter of each epoch from a fixed half of the demonstration pool (201,872 windows), and the masked-demonstration variant uses only the 176,955 masked demonstration windows (21.9\% of each epoch). Both were run after the main configuration was fixed.\looseness=-1

\par\noindent\textbf{Results.}
All three variants roughly halve other-agent contact, whereas only masked-only demonstrations reduce untriggered or late runs (Table~\ref{tab:q6-costs}). Masked-only demonstrations also raise no-target distance the most and lower excess target contact in natural-log replay more than the 25\% demos variant and main training ($9.0\pm1.0$ points, against $15.1\pm0.4$ and $20.0\pm0.8$), at a cost of $3.2\pm0.4$ points of collision reduction relative to main OnCue training. The 25\% demos variant keeps the reduction and the untriggered or late share, and removing stationary windows keeps both the avoidance gain and the reduced progress.

\par\noindent\textbf{Finding.}
The gain thus comes with two costs, hesitation (more untriggered or late runs) and extra other-agent contact, which respond differently to changes in training. The halving of contact does not come with less travel, since every variant travels at least as far as the main model without the target (Table~\ref{tab:q6-costs}). The hesitation depends partly on the unoccluded demonstrations, because with pools of similar size only masked-only demonstrations reduce it. Unoccluded demonstrations differ from masked ones in view, in starting the response at $t_v$, and in including Anticipatory tests, which this comparison cannot separate.

\begin{figure}[!t]
\centering
\includegraphics[width=\columnwidth]{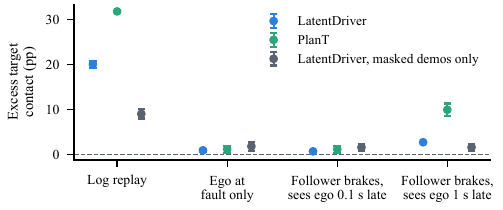}
\caption{\textbf{Excess target contact in natural-log encounters.} Target contact of each OnCue-trained model (Logs + OnCue, or its masked demos only variant) minus its same-seed logs-only model, in percentage points, on 148 masked encounters. Points give mean $\pm$ SD over three seeds. The first two conditions use the same log-replay runs and count all or only ego-at-fault contacts. In the last two, a target following the ego brakes and sees the ego 0.1 or 1\,s late (F.5.4).}
\label{fig:q6-natural}
\end{figure}

\subsubsection*{F.5.4 Natural Logged Encounters}\label{toc:F.5.4}
Generated tests insert their targets, so we also evaluate trained planners on natural logged encounters. We identify 148 evaluable encounters, in which the target can reach the ego path, on 300 WOMD training shards, excluding logs from the same driving episodes as fine-tuning data, and 79 of them are Reactive. Pretrained LatentDriver and PlanT were trained on WOMD training data, which may include these logs. Before any planner run, target position and direction at replay start classify 109 encounters as following from behind and 39 as crossing, ahead, or oncoming. Evaluation is masked, with three training seeds for every trained model.\looseness=-1

In log replay, only the ego follows the tested planner, and all other road users keep their recorded motion. We therefore also count ego-at-fault target contact with the nuPlan-derived rule used for background contact (B.2). This diagnostic applies only to natural encounters, and generated-test results still count every target contact.

Because log replay lets a following target drive into a slowing ego, two further diagnostics let the target brake like a cautious follower when the ego is ahead on its path and their headings differ by at most $45^\circ$. Otherwise the target replays its log or, after braking, returns toward its logged speed along its path until it rejoins the log. A Gipps safe-speed bound~\cite{gipps1981} assumes that both vehicles brake at 3.4\,m/s$^2$, the AASHTO stopping-sight-distance deceleration~\cite{aashto2018}, and stop with a 2\,m gap. Target deceleration is capped at 3.4\,m/s$^2$, and the target never advances beyond its recorded position. The target observes the ego 0.1 or 1\,s late, and with the longer lag it extrapolates the observed ego state from the heading and speed change over the preceding 0.5\,s. The reaction time inside the Gipps formula stays one frame (0.1\,s) in both diagnostics, so the lag only delays what the follower sees. The two lags test how sensitive the result is to the follower\textquotesingle{}s observation delay under this fixed braking cap. Runs that remain on the logged trajectory are checked field by field against replay. An ego counts as stopped when its speed stays below 0.5\,m/s for 1\,s after traveling 2\,m and reaching 1\,m/s.\looseness=-1

\par\noindent\textbf{Results.}
OnCue training raises log-replay target contact over the logs-only control by $20.0\pm0.8$ points for LatentDriver and $31.8\pm0.0$ points for PlanT, mostly in follower encounters. Counting ego-at-fault contact or letting the target brake substantially reduces these differences (Figure~\ref{fig:q6-natural}). OnCue-trained LatentDriver stops about seven times as often as its control and travels less without the target, and in the 39 other encounters both trained planners contact the target slightly more often than their controls but far less often than the pretrained models (Table~\ref{tab:q6-natural}).\looseness=-1

Braking followers with a 1\,s lag still leave PlanT $9.9\pm1.4$ points of excess contact. In seed 0, the ego\textquotesingle{}s strongest deceleration over any 0.5\,s in the preceding 3\,s exceeds the followers\textquotesingle{} 3.4\,m/s$^2$ braking cap in 10 of its 13 rear contacts, and most contacts occur near standstill. Followers brake for OnCue-trained LatentDriver in 37 to 39 runs per seed, mostly needing more braking than the cap, versus 2 to 3 runs for its control.

\par\noindent\textbf{Finding.}
The excess arises because OnCue-trained planners often slow or stop in traffic, where logged followers, which cannot brake, run into them. Since the fault rule blames the follower for such rear contacts, ego-at-fault contact changes little. Braking followers reduce this excess but can leave some of it, because some ego slowdowns are sharper than a follower braking at the 3.4\,m/s$^2$ cap can match.

\begin{figure*}[!t]
\centering
\includegraphics[width=\textwidth,height=6.75in,keepaspectratio]{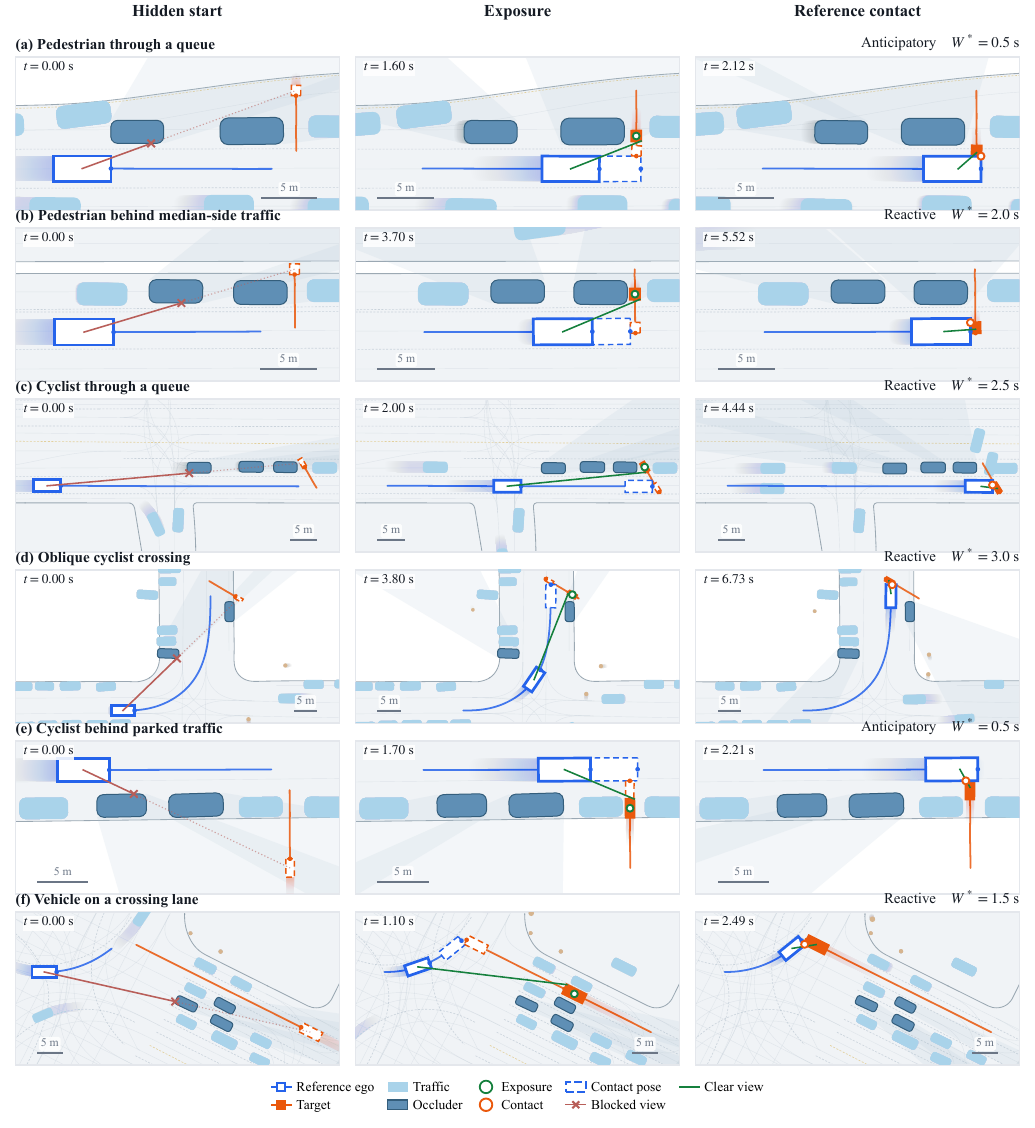}
\caption{\textbf{Occluded encounters across target classes.} Pedestrian and cyclist targets emerge near parked or stopped traffic or through queues, and a vehicle follows a crossing lane. Each row shows one saved test at the hidden start, confirmed exposure, and first reference contact. Row headers give the oracle label and requested window $W^*$. Times are measured from replay start. The three views in each row share the same metric scale. Scale bars indicate 5\,m. In the hidden-start view, the dashed orange outline marks the hidden target. In the exposure view, dashed outlines show contact poses. Colors and sightlines follow Figure~\ref{fig4}.}\label{fig16}
\end{figure*}
\section*{Appendix G. Encounter Examples}\label{toc:G}
Figure~\ref{fig16} shows six Reactive and Anticipatory encounters from different scenes, generated and labeled as in the main paper.
\clearpage
\renewcommand*{\backrefalt}[4]{}

\fi
\end{document}